%% file: main.tex
\documentclass{article}

 \usepackage[nonatbib, final]{neurips_2021}

\usepackage[utf8]{inputenc} %
\usepackage[T1]{fontenc}    %
\usepackage{hyperref}
\usepackage{url}            %
\usepackage{booktabs}       %
\usepackage{amsfonts}       %
\usepackage{nicefrac}       %
\usepackage{microtype}
\usepackage{graphicx}
\usepackage{subcaption}
\usepackage{booktabs} %
\usepackage{color}
\usepackage{lmodern}
\usepackage{float}
\usepackage{amssymb}
\usepackage{amsthm}
\usepackage{amsmath}
\usepackage{enumitem}
\usepackage[dvipsnames]{xcolor}
\usepackage{tabularx}
\usepackage{mathtools}
\usepackage{bbm}
\usepackage{times}

\usepackage{microtype}      %
\usepackage{wrapfig}
\usepackage{algorithmic}
\usepackage{algorithm}
\usepackage[xindy,acronym,toc,smallcaps,nomain]{glossaries}
\glsdisablehyper
\makeglossaries
\newacronym{PROTEINS}{proteins}{proteins}
\newacronym{IMDB-M}{imdb-m}{IMDB-M}
\newacronym{REDDIT-MULTI-5K}{reddit-multi-5k}{reddit-multi-5k}
\newacronym{COLLAB}{collab}{collab}
\newacronym{MNIST}{mnist}{mnist}
\newacronym{TU}{tu}{tu}

\glsunset{PROTEINS}
\glsunset{IMDB-M}
\glsunset{REDDIT-MULTI-5K}
\glsunset{COLLAB}
\glsunset{MNIST}
\glsunset{TU}

\newacronym{BO}{bo}{Bayesian optimisation}
\newacronym{GA}{ga}{Genetic algorithm}
\newacronym{WL}{wl}{Weisfeiler-Lehman}
\newacronym{GP}{gp}{Gaussian Process}
\newacronym{GPWL}{gpwl}{Gaussian Process with Weisfeiler-Lehman Kernel}

\newacronym{GNN}{gnn}{Graph Neural Network}
\newacronym{GCN}{gcn}{Graph Convolution Network}
\newacronym{GIN}{gin}{Graph Isomorphism Network}
\newacronym{ASR}{asr}{Attack Success Rate}
\newacronym{ARD}{ard}{Automatic Relevance Determination}
\newacronym{EI}{ei}{Expected Improvement}
\newacronym{BLR}{blr}{Bayesian Linear Regression}
\newacronym{DGL}{dgl}{Deep Graph Library}
\newacronym{RMSE}{rmse}{Root Mean Squared Error}

\newacronym{GRABNEL}{grabnel}{grabnel}
\glsunset{GRABNEL}
\newacronym{RL-S2V}{rl-s2v}{rl-s2v}
\glsunset{RL-S2V}

\usepackage{todonotes}

\usepackage{comment}
\input{math.tex}

\title{Adversarial Attacks on Graph Classification via Bayesian Optimisation}

    \author{
	Xingchen Wan%
	\quad
	Henry Kenlay%
	\quad
	Binxin Ru%
	\\
    \textbf{Arno Blaas}%
    \quad
	\textbf{Michael A. Osborne}%
	\quad
	\textbf{Xiaowen Dong}%
	\\
	\small{Machine Learning Research Group, University of Oxford, Oxford, UK}
	\\ 
	\small{\texttt{\{xwan,kenlay,robin,arno,mosb,xdong\}@robots.ox.ac.uk}}
}
\newcommand{\G}{\mathcal{G}} %
\newcommand{\V}{\mathcal{V}} %
\renewcommand{\E}{\mathcal{E}} %
\newcommand{\e}{\mathbf{e}} %
\newcommand{\X}{\mathbf{X}} %
\newcommand{\A}{\mathbf{A}} %
\DeclarePairedDelimiter\set{\lbrace}{\rbrace} %

\begin{document}
\maketitle

\begin{abstract}
Graph neural networks, a popular class of models effective in a wide range of graph-based learning tasks, have been shown to be vulnerable to adversarial attacks. While the majority of the literature focuses on such vulnerability in node-level classification tasks, little effort has been dedicated to analysing adversarial attacks on graph-level classification, an important problem with numerous real-life applications such as biochemistry and social network analysis.The few existing methods often require unrealistic setups, such as access to internal information of the victim models, or an impractically-large number of queries. We present a novel Bayesian optimisation-based attack method  for graph classification models.  Our method is \emph{black-box}, \emph{query-efficient} and \emph{parsimonious} with respect to the perturbation applied. We empirically validate the effectiveness and flexibility of the proposed method on a wide range of graph classification tasks involving varying graph properties, constraints and modes of attack. Finally, we analyse common interpretable patterns behind the adversarial samples produced, which may shed further light on the adversarial robustness of graph classification models. 
An open-source implementation is available at \url{https://github.com/xingchenwan/grabnel}.
\end{abstract}

\section{Introduction}

Graphs are a general-purpose data structure consisting of entities represented by nodes and edges which encode pairwise relationships. 
Graph-based machine learning models has been widely used in a variety of important applications such as semi-supervised learning, link prediction, community detection and graph classification \cite{cai2018comprehensive, zhou2020graph, hamilton2020graph}. Despite the growing interest in graph-based machine learning,
it has been shown that, like many other machine learning models, graph-based models are vulnerable to adversarial attacks \cite{sun2018adversarial, jin2020adversarial}. If we want to deploy such models in environments where the risk and costs associated with a model failure are high e.g. in social networks, it would be crucial to understand and assess the model stability and vulnerability by simulating adversarial attacks.

Adversarial attacks on graphs can be aimed at different learning tasks. This paper focuses on  graph-level classification, where given an input graph (potentially with node and edge attributes), we wish to learn a function that predicts a property of interest related to the graph. 
Graph classification is an important task with many real-life applications, especially in bioinformatics and chemistry \cite{Morris2020, morris2020tudataset}. For example, the task may be to accurately classify if a molecule, modelled as a graph whereby nodes represent atoms and edges model bonds, inhibits \textsc{hiv} replication or not. Although there are a few attempts on performing adversarial attacks on graph classification \cite{dai2018adversarial, ma2019attacking}, they all operate under unrealistic assumptions such as the need to query the target model a large number of times or access a portion of the test set to train the attacking agent. 
To address these limitations, we formulate the adversarial attack on graph classification as a black-box optimisation problem and solve it with \gls{BO}, a query-efficient state-of-the-art zeroth-order black-box optimiser. Unlike existing work, our method is query-efficient, parsimonious in perturbations and does not require policy training on a separate labelled dataset to effectively attack a new sample. Another benefit of our method is that it can be easily adapted to perform various modes of attacks such as deleting or rewiring edges and node injection. 
Furthermore, we investigate the topological properties of the successful adversarial examples found by our method and offer valuable insights on the connection between the graph topology change and the model robustness.

The main contributions of our paper are as follows. 
First, we introduce a novel black-box attack for graph classification, \gls{GRABNEL}\footnote{Stands for \textit{\underline{Gr}aph \underline{A}dversarial attack via \underline{B}ayesia\underline{N} \underline{E}fficient \underline{L}oss-minimisation}.}, which is both query efficient and parsimonious. We believe this is the first work on using \gls{BO} for adversarial attacks on graph data.
Second, we analyse the generated adversarial examples to link the vulnerability of graph-based machine learning models to the topological properties of the perturbed graph, an important step towards interpretable adversarial examples that has been overlooked by the majority of the literature. 
Finally, we evaluate our method on a range of real-world datasets and scenarios including detecting the spread of fake news on Twitter, which to the best of our knowledge is the first analysis of this kind in the literature.

\section{Proposed Method: GRABNEL}
\label{sec:method}

\begin{figure*}[t]
    \centering
    \vspace{-5mm}
    \includegraphics[clip, width=\linewidth]{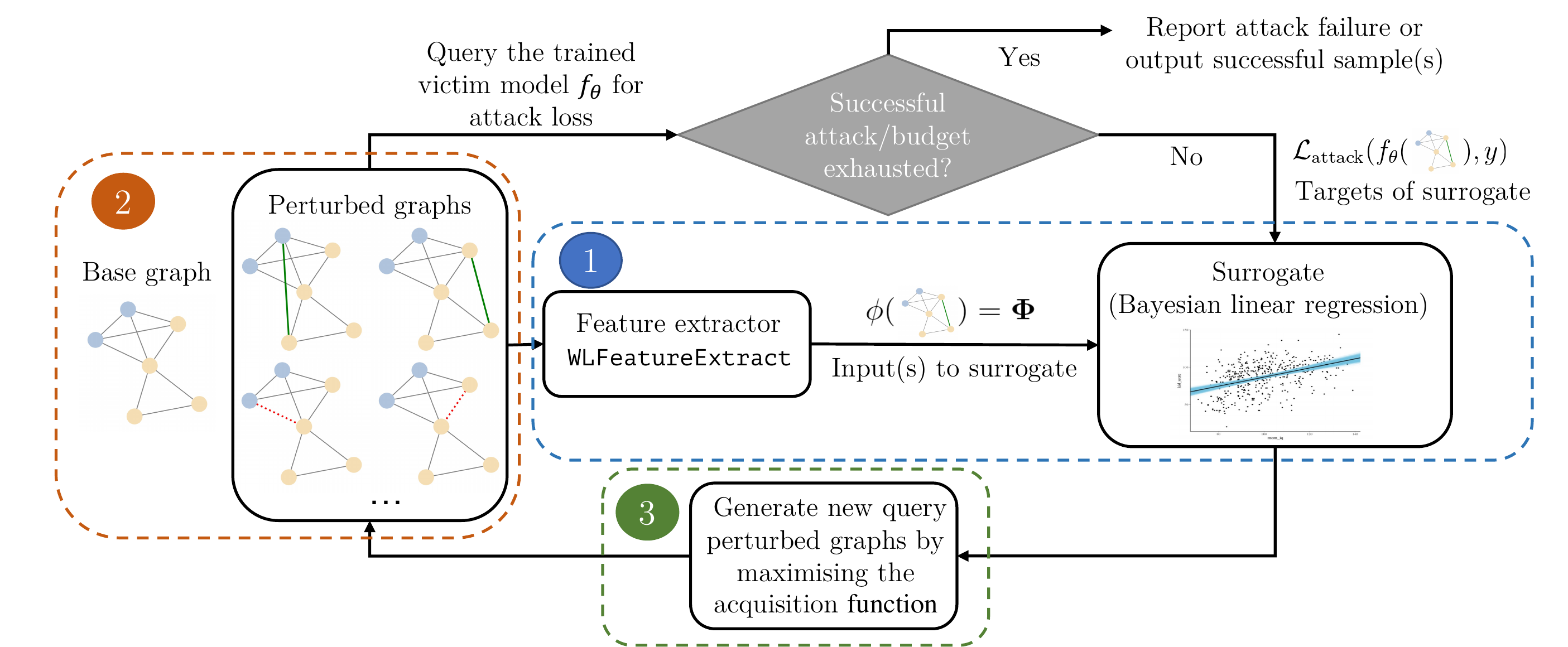}
     \vspace{-4mm}
    
    \caption{The overall pipeline of \gls{GRABNEL}. The key components are explained in different paragraphs of Sec \ref{sec:method}:  \emph{Surrogate model} describes the construction of the \gls{BO} surrogate and the feature extractor (\textcolor{blue}{Block 1}), \emph{Sequential perturbation selection} describes how base graphs and perturbed graphs as candidates of adversarial attack are selected (\textcolor{orange}{Block 2}), and \emph{Optimisation of acquisition function} describes how new query points are generated by \gls{BO} via optimising acquisition (\textcolor{green}{Block 3}). A detailed algorithmic description for \gls{GRABNEL} is also available in App. \ref{app:algorithms}.
    }  
    \label{fig:main}
    \vspace{-4mm}
\end{figure*}

\paragraph{Problem Setup} A graph $\G=(\V, \E)$ is defined by a set of nodes $\V=\set{v_i}_{i=1}^n$ and edges $\E=\set{\e_i}_{i=1}^m$ where each edge $\e_k=\set{v_i, v_j}$ connects between nodes $v_i$ and $v_j$. The overall topology can be represented by the adjacency matrix $\A \in \set{0,1}^{n \times n}$ where $A_{ij}=1$ if the edge $\set{v_i, v_j}$ is present\footnote{We discuss the unweighted graphs for simplicity; our method may also handle other graph types.}. 
The attack objective in our case is to degrade the predictive performance of the trained victim graph classifier $f_\theta$ by finding a graph $\G'$ perturbed from the original test graph $\G$ (ideally with the minimum amount of perturbation) such that $f_\theta$ produces an incorrect class label for $\G$. In this paper, we consider the \emph{black-box evasion} attack setting, where the adversary agent cannot access/modify the the victim model $f_\theta$ (i.e. network architecture, weights $\theta$ or gradients) or its training data $\{ (\G_i, y_i) \}_{i=1}^L$; the adversary can only interact with $f_\theta$ by querying it with an input graph $\G'$ and observe the model output $f_{\theta}(\G')$ as pseudo-probabilities over all classes in a $C$-dimensional standard simplex.
Additionally, we assume that \emph{sample efficiency} is highly valued: we aim to find adversarial examples with the minimum number of queries to the
victim model. We believe that this is a practical and difficult setup that accounts for the prohibitive monetary, logistic
and/or opportunity costs of repeatedly querying a (possibly huge and complicated) real-life victim model. With a high
query count, the attacker may also run a higher risk of getting detected. Formally, the objective function of our \gls{BO} attack agent can be formulated as a black-box maximisation problem:
\begin{equation}
\max_{\G'\in \Psi(\G)}\mathcal{L}_{\mathrm{attack}}\big(f_{\theta}(\G'), y\big) \text{ } \mathrm{~s.t.~} \text{ } y = \arg \max f_{\theta}(\G)
\end{equation}
where $f_\theta$ is the pretrained victim model that remains fixed in the evasion attack setup and $y$ is the correct label of the original input $\G$. Denote the output logit for the class $y$ as $f_{\theta}(\G)_y$, the \emph{attack} loss $\mathcal{L}_{\mathrm{attack}}$ can be defined as:
\begin{equation}
\mathcal{L}_{\mathrm{attack}}\Big(f_{\theta}(\G'), y\Big) = 
\begin{cases}
\max_{t \in \mathcal{Y}, t \neq y} \log f_{\theta}(\G')_t  - \log f_{\theta}(\G')_y & \text{ (untargeted attack)}\\
\log f_{\theta}(\G')_t - \log f_{\theta}(\G')_y & \text{ (targeted attack on class } t \text{)},
\end{cases} 
\end{equation}
where $f_{\theta}(\cdot)_t$ denotes the logit output for class $t$. Such an attack loss definition is commonly used both in the traditional image attack and the graph attack literature \cite{carlini2017towards, zugner2018adversarial} although our method is compatible with any choice of loss function. Furthermore, $\Psi(\G)$ refers to the set of possible $\G'$ generated from perturbing $\G$. In this work, we experiment with a diverse modes of attacks to show that our attack method can be generalised to different set-ups: 
\begin{itemize} [leftmargin=0.1in, noitemsep, topsep=0.05pt]
\item creating/removing an edge: we create perturbed graphs by flipping the connection of a small set of node pairs $\delta \A = \{\{u_i, v_i\}\}_{i=1}^{\Delta}$ of $\G$ following previous works \cite{zugner2018adversarial, dai2018adversarial}; 
\item rewiring or swapping edges: similar to \cite{ma2019attacking}, we select a triplet $(u, v, s)$ where we either rewire the edge $(u \rightarrow v)$ to $(u \rightarrow s)$ (rewire), or exchange the edge weights $w(u,v)$ and $w(u,s)$ (swap);
\item node injection: we create new nodes together with their attributes and connections in the graph.
\end{itemize}
The overall routine of our proposed \gls{GRABNEL} is presented in Fig \ref{fig:main} (and in pseudo-code form in App \ref{app:algorithms}), and we now elaborate each of its key components.

\paragraph{Surrogate model} The success of \gls{BO} hinges upon the surrogate model choice. Specifically, such a surrogate model needs to 1) be flexible and expressive enough to locally learn the latent mapping from a perturbed graph $\G'$ to its attack loss $\mathcal{L}_{\mathrm{attack}}(f_{\theta}(\G'), y)$ (note that this is different and generally easier than learning $\G' \rightarrow y$, which is the goal of the classifier $f_{\theta}$), 2) admit a probabilistic interpretation of uncertainty -- this is key for the exploration-exploitation trade-off in \gls{BO}, yet also 3) be simple enough such that the said mapping can be learned with a small number of queries to $f_\theta$ to preserve sample efficiency. Furthermore, given the combinatorial nature of the graph search space, it also needs to 4) be capable of scaling to large graphs (e.g. in the order of $10^3$ nodes or more) typical of common graph classification tasks with reasonable run-time efficiency. Additionally, given the fact that \gls{BO} has been predominantly studied in the continuous domain which is significantly different from the present setup, the design of a appropriate surrogate is highly non-trivial. To handle this set of conflicting desiderata, we propose to first use a \emph{\gls{WL} feature extractor} to extracts a vector space representation of $\G$, followed by a  \emph{sparse Bayesian linear regression} which balances performance with efficiency and gives an probabilistic output.

With reference to Fig. \ref{fig:main}, given a perturbation graph $\G'$ as a proposed adversarial sample, the \gls{WL} feature extractor first extracts a vector representation $\vphi(\G')$ in line with the \gls{WL} subtree kernel procedure (but without the final kernel computation) \cite{shervashidze2011weisfeiler}. For the case where the node features are discrete, let $x^0(v)$ be the initial node feature of node $v \in \V$ (note that the node features can be either scalars or vectors) , we iteratively aggregate and hash the features of $v$ with its neighbours, $\{u_i\}_{i=1}^{\mathrm{deg}(v)}$, using the original \gls{WL} procedure at all nodes to transform them into discrete labels:
\begin{equation}
 x^{h+1}(v) = \mathrm{hash}\Big(x^h(v), x^h(u_1), ..., x^h(u_{\mathrm{deg}(v)})\Big), ~\forall h \in \{0, 1, \ldots, H-1 \},
 \label{eq:catwl}
\end{equation}
where $H$ is the total number of \gls{WL} iterations, a hyperparameter of the procedure. At each level $h$, we compute the feature vector $\vphi_h(\G') = [c(\G', \mathcal{X}_{h1}), ..., c(\G', \mathcal{X}_{h|\mathcal{X}_h|})]^{\top}$, where $\mathcal{X}_h$ is the set of distinct node features $x^h$ that occur in all input graphs at the current level and $c(G', x^h)$ is the counting function that counts the number of times a particular node feature $x^h$ appears in $G'$. For the case with continuous node features and/or weighted edges, we instead use the modified \gls{WL} procedure proprosed in \cite{togninalli2019wasserstein}:
\begin{equation}
	x^{h+1}(v) = \frac{1}{2} \Big( x^h(v) + \frac{1}{\mathrm{deg}(v)} \sum_{i=1}^{\mathrm{deg}(v)} w(v, u_i) x^h(u_i) \Big), ~\forall h \in \{0, 1, \ldots, H-1 \},
	\label{eq:contwl}
\end{equation} where $w(v, u_i)$ denotes the (non-negative) weight of edge $e_{\{v, u_i\}}$ ($1$ if the graph is unweighted) and we simply have feature at level $h$ $\vphi_h(\G') = \mathrm{vec}(\mathbf{X}_h)$, where $\mathbf{X}_h$ is the feature matrix of graph $\G'$ at level $h$ by collecting the features at each node $\mathbf{X}_h = \Big[x^h(1), ... x^h(v)\Big]$ and $\mathrm{vec}(\cdot)$ denotes the vectorisation operator. In both cases, at the end of $H$ \gls{WL} iterations we obtain the final feature vector $\vphi(\G') = \mathrm{concat}\big(\vphi_1(\G'), ... ,\vphi_H(\G') \big)$ for each training graph in $[1, n_{\G'}]$ to form the feature matrix $\mPhi = [\vphi(\G'_1), ... \vphi(\G'_{|n_{\G'}|})]^{\top} \in \mathbb{R}^{|n_{\G'}| \times D}$ to be passed to the Bayesian regressor -- it is particularly worth noting that the training graphs here denote inputs to train the surrogate model of the attack agent and are typically perturbed versions of a \emph{test} graph $\G$ of the victim model; they are \emph{not} the graphs that are used to train the victim model itself: in an evasion attack setup, the model is considered frozen and the training inputs cannot be accessed by the attack agent any point in the pipeline. The \gls{WL} iterations capture both information related to individual nodes and topological information (via neighbourhood aggregation), and have been shown to have comparable distinguishing power to some \gls{GNN} models \cite{morris2019weisfeiler}, and hence the procedure is expressive. Alternative surrogate choices could be, for example, \gls{GNN}s with the final fully-connected layer replaced by a probabilistic linear regression layer such as the one proposed in \cite{shi2019bridging}. However, in contrast to these, our extraction process $\G' \rightarrow \vphi(\G')$ requires no learning from data (we only need to learn the Bayesian linear regression weights) and therefore should lead to better sample efficiency. Alternatively, we may also use a \gls{GP} surrogate, such as the \gls{GPWL} model proposed in \cite{ru2020neural} that directly uses a \gls{GP} model together with a \gls{WL} kernel. Nonetheless, while \gls{GP}s are theoretically more expressive (although we empirically show in App. \ref{app:surrogate_comparisons} that in most of the cases their predictive performances are comparable), they are also much more expensive with a cubic scaling w.r.t the number of training inputs. Furthermore, \gls{GPWL} is designed specifically for neural architecture search, which features small, directed graphs with discrete node features only; on the other hand, the \gls{GRABNEL} surrogate covers a much wider scope of applications

\begin{figure*}[t]
    \centering
    \vspace{-4mm}
    \includegraphics[clip, width=1\linewidth]{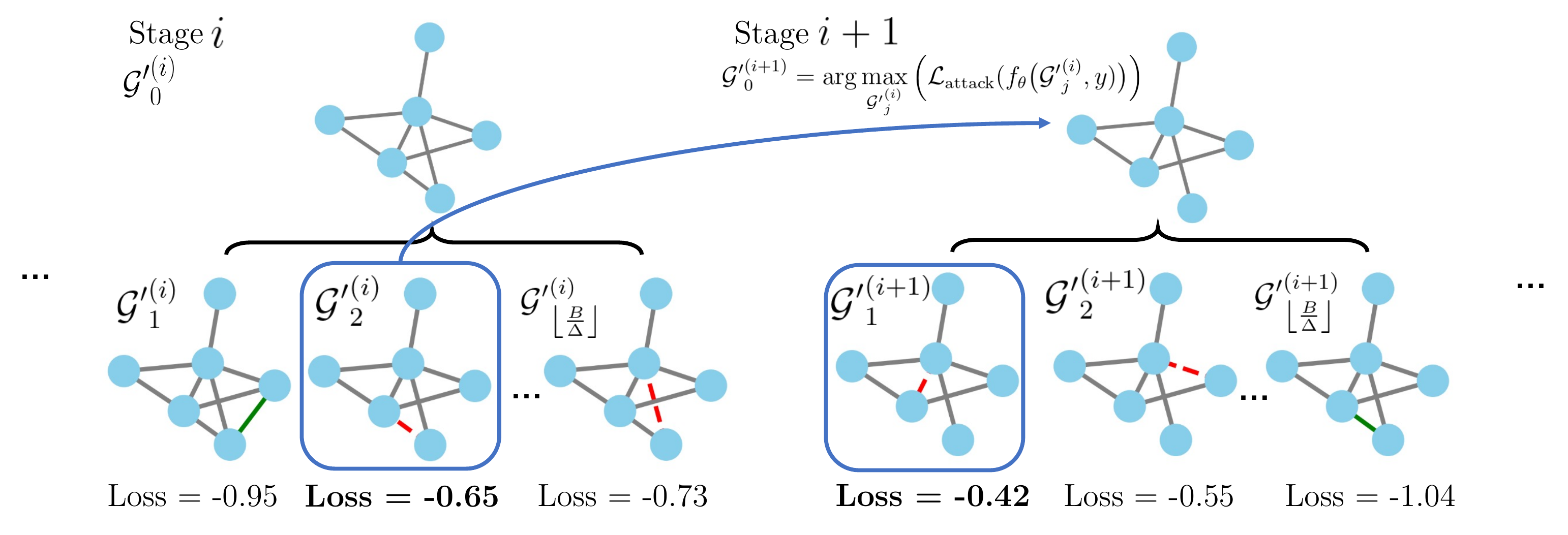}
     \vspace{-6mm}
    
    \caption{Sequential edge selection. At each stage, the \gls{BO} agent sequentially proposes candidate graphs with edge edit distance of $1$ from the base graph $\mathcal{G}'^{(i)}_0$ (which is the original unperturbed graph $\mathcal{G}$ at initialisation, or a perturbed graph that led to the largest increase in loss from the previous stage otherwise) by selecting the graph that maximises the acquisition function value amongst all candidates generated via sampling/genetic algorithm (see details at \emph{Optimisation of acquisition function}). This procedure repeats until either the attack succeeds (i.e. we find a graph $\mathcal{G}'$ with $\mathcal{L}_{\mathrm{attack}}(f_{\theta}(\mathcal{G}'), y) > 0$) or the maximum number of $B$ queries to $f_\theta$ is exhausted.
    }  
    \label{fig:seqselection}
    \vspace{-4mm}
\end{figure*}

When we select a large $H$ or if there are many training inputs and/or input graph(s) have a large number of nodes/edges, there will likely be many unique \gls{WL} features and the resulting feature matrix will be very high-dimensional, which would lead to high-variance regression coefficients $\boldsymbol{\alpha}$ being estimated if $n_{\G'}$ (number of graphs to train the surrogate of the attack agent)
is comparatively few. To attain a good predictive performance in such a case, we employ Bayesian regression surrogate with the \gls{ARD} prior to learn the mapping $\mPhi \rightarrow \mathcal{L}_{\mathrm{attack}}(f_{\theta}(\G'), y)$, which regularises weights and encourages sparsity in $\boldsymbol{\alpha}$ \cite{wipf2007new}:
\begin{align}
\mathcal{L}_{\mathrm{attack}} | \mPhi, \boldsymbol{\alpha}, \sigma_n^2 &\sim \mathcal{N}(\boldsymbol{\alpha}^{\top} \mPhi,  \sigma_n^2\mathbf{I}),\\
\boldsymbol{\alpha} | \boldsymbol{\lambda} &\sim \mathcal{N} (\boldsymbol{0}, \boldsymbol{\Lambda}),~ \mathrm{diag}(\boldsymbol{\Lambda}) = \boldsymbol{\lambda}^{-1} = \{ \lambda_1^{-1}, ..., \lambda_{D}^{-1} \}, \\
\lambda_i &\sim \mathrm{Gamma}(k, \theta) \text{ } \forall i \in [1, D],
\end{align}
where $\boldsymbol{\Lambda}$ is a diagonal covariance matrix.
To estimate $\boldsymbol{\alpha}$ and noise variance $\sigma_n^2$, we optimise the model marginal log-likelihood. Overall, the \gls{WL} routines scales as $\mathcal{O}(Hm)$ and Bayesian linear regression has a linear runtime scaling w.r.t. the number of queries; these ensure the surrogate is scalable to both larger graphs and/or a large number of graphs, both of which are commonly encountered in graph classification (See App \ref{app:runtime_analysis} for a detailed empirical runtime analysis).

\paragraph{Sequential perturbation selection} In the default structural perturbation setting, given an attack budget of $\Delta$ (i.e. we are allowed to flip up to $\Delta$ edges from $\G$), finding exactly the set of perturbations $\delta \mathbf{A}$ that leads to the largest increase in $\mathcal{L}_{\mathrm{attack}}$ entails an combinatorial optimisation over $\binom{n^2}{\Delta}$ candidates. This is a huge search space that is difficult for the surrogate to learn meaningful patterns in a sample-efficient way even for modestly-sized graphs. To tackle this challenge, we adopt the strategy illustrated in Fig. \ref{fig:seqselection}: given the query budget $B$ (i.e. the total number of times we are allowed to query $f_\theta$ for a given $\G$), we assume $B \geq \Delta$ and amortise $B$ into $\Delta$ stages and focus on selecting \emph{one} edge perturbation at each stage. While this strategy is greedy in the sense that it always commits the perturbation leading to the largest increase in loss at each stage, it is worth noting that we do \emph{not} treat the previously modified edges differently, and the agent can, and does occasionally as we observe empirically, “correct” previous modifications by flipping edges back: this is possible as the effect of edge selection is permutation invariant. Another benefit of this strategy is that it can potentially make full use of the entire attack budget $\Delta$ \emph{while} remaining parsimonious w.r.t. the amount of perturbation introduced, as it only progresses to the next stage and modifies the $\G$ further when it fails to find a successful adversarial example in the current stage.

\paragraph{Optimisation of acquisition function} At each \gls{BO} iteration, acquisition function $\alpha(\cdot)$ is optimised to select the next point(s) to query the victim model $f_\theta$. However, commonly used gradient-based optimisers cannot be used on the discrete graph search space; a na\"ive strategy would be to randomly generate many perturbed graphs, evaluate $\alpha$ on all of them, and choose the maximiser(s) to query $f_\theta$ next. While potentially effective on modestly-sized $\G$ especially with our sequential selection strategy, this strategy nevertheless discards any known information about the search space.

Inspired by recent advances in \gls{BO} in non-continuous domains \cite{cowen2020hebo, wan2021think}, we optimise $\alpha$ via an adapted version of the \gls{GA} in \cite{dai2018adversarial}, which is well-suited for our purpose but is not particularly sample efficient since many evolution cycles could be required for convergence. However, the latter is not a serious issue here as we only use \gls{GA} for acquisition optimisation where we only query the surrogate instead of the victim model, a subroutine of \gls{BO} that does not require sample efficiency. We outline its ingredients below:
\begin{itemize}[leftmargin=0.1in, noitemsep, topsep=0.02pt]
\item \emph{Initialisation:} While \gls{GA} typically starts with random sampling in the search space to fill the initial population, in our case we are not totally ignorant about the search space as we could have already queried and observed $f_\theta$ with a few different perturbed graphs $\G'$. A smoothness assumption on the search space would be that if a $\G'$ with an edge $(u, v)$ flipped from $G$ led to a large $\mathcal{L}_{\mathrm{attack}}$, then another $\G'$ with $(u, s), s \not\in \{u,  v\}$ flipped is more likely to do so too. To reflect this, we fill the initial population by \emph{mutating} the top-$k$ queried $\G'$s leading to the largest $\mathcal{L}_{\mathrm{attack}}$ seen so far in the current stage, where for $\G'$ with $(u, v)$ flipped from the base graph we 1) randomly choose an end node ($u$ or $v$) and 2) change that node to another node in the graph except $u$ or $v$ such that the perturbed edges in all children shares one common end node with the parent.
\item \emph{Evolution}: After the initial population is built, we follow the standard evolution routine by evaluating the acquisition function value for each member as its \emph{fitness}, selecting the top-$k$ performing members as the breeding population and repeating the mutation procedure in initialisation for a fixed number of rounds. At termination, we simply query $f_\theta$ with the graph(s) seen so far (i.e. computing the loss in Fig. \ref{fig:seqselection}) with the largest acquisition function value(s) seen during \gls{GA}.
\end{itemize}

\section{Related Works}
\label{sec:related}

\textbf{Adversarial attack on graph-based models} 
There has been an increasing attention in the study of adversarial attacks in the context of \gls{GNN}s \cite{sun2018adversarial, jin2020adversarial}. One of the earliest models, Nettack, attacks a \gls{GCN} node classifier by optimising the attack loss of a surrogate model using a greedy algorithm \cite{zugner2018adversarial}. Using a simple heuristic, \textsc{dice} attacks node classifiers by adding edges between nodes of different classes and deleting edges connecting nodes of the same class \cite{waniek2018hiding}. However, they cannot be straightforwardly transferred graph classification: for Nettack, unlike node classification tasks, we have no access to training input graphs or labels for the victim model during test time to train a similar surrogate in graph classification; for \textsc{dice} (and also more recent works like \cite{wang2020evasion}), node labels do not exist in graph classification (we only have a single label for the entire graph). We nonetheless acknowledge the other contributions in these works, such as the introduction of constraints to improve imperceptibility, in our experiments in Sec \ref{sec:experiments}.

First methods that do extend to graph classification include \cite{dai2018adversarial, ma2019attacking}: 
\cite{dai2018adversarial} propose a number of techniques, including \gls{RL-S2V}, which uses reinforcement learning to attack both node and graph classifiers in a black-box manner, and the \gls{GA}-based attack, which we adapt into our \gls{BO} acquisition optimisation. However, \cite{dai2018adversarial} primarily focus on the \textsc{s2v} victim model, do not emphasise on sample efficiency, and to train a policy that attacks in an one-shot manner on the test graphs, \gls{RL-S2V} has to query repeatedly on a separate validation set. We empirically compare against it in App. \ref{sec:rls2v}. Another related work is ReWatt \cite{ma2019attacking}, which similarly uses reinforcement learning but through rewiring. Compared to both these methods, \gls{GRABNEL} does not require an additional validation set and is much more query efficient. Other black-box methods without surrogate models have also been proposed that could be \emph{potentially} be applied to graph classification: \cite{ma2020towards} exploit common \gls{GNN} structural bias to attack node features, while \cite{chang2020restricted} relate graph embedding to graph signal processing and construct tailored attack objectives in different \gls{GNN}s. In comparison to these works that exploit the characteristics of existing architectures to varying degrees, we argue that the optimisation-based method proposed in the our work is more flexible and agnostic to architecture choices, and should be more generalisable to new architectures. Nonetheless, in cases where some architectural information is available, we believe there could be \emph{combinable} benefits: for example, the importance scores proposed in \cite{ma2020towards} could be used as \emph{sampling weights} as priors to bias \gls{GRABNEL} towards selecting more vulnerable nodes. We defer detailed investigations of such possibilities to a future work. Finally, there have also been various previous works that focus on a different setup than ours: A white-box optimisation strategy (alternating direction method of multipliers) is proposed in \cite{jin2020certified}. \cite{zhang2020backdoor, xu2021explainability, bojchevski2019adversarial} propose back-door attacks that involve poisoning of the training data before training and/or the test data at inference. \cite{tang2020adversarial} attack hierarchical graph pooling networks, but similar to \cite{zugner2018adversarial} the method requires access to training input/targets. Ultimately, a number of factors, including but not limited to 1) existence/strictness of the query budget, 2) strictness of the perturbation budget, 3) attacker capabilities and 4) sizes of the graphs, would decide which algorithm/setup is more appropriate and should be adopted in a problem-specific way. Nonetheless, we argue that our
setup is both challenging and highly significant as it resembles the
capabilities a real-life attacker might have (no access to training data; no access to model
parameter/gradients and limited query/perturbation budgets).

\textbf{Adversarial attacks using \gls{BO}}
\gls{BO} as a means to find adversarial examples in the black-box evasion setting has been successfully proposed for classification models on tabular \cite{suya2017query} and image data \cite{ru2019bayesopt,  zhao2019design, shukla2019black, munoz2017bayesian}. However, we address the problem for graph classification models, which work on structurally and topologically fundamentally different inputs. This implies several nontrivial challenges that require our method to go beyond the vanilla usage of \gls{BO}: for example, 
the inputs cannot be readily represented as vectors like for tabular or image data and the perturbations that we consider for such inputs are not defined on a continuous, but on a discrete domain.

\section{Experiments}
\label{sec:experiments}
We validate the performance of the proposed method in a wide range of graph classification tasks with varying graph properties, including but not limited to the typical \gls{TU} datasets considered in previous works \cite{dai2018adversarial, ma2019attacking}. As a demonstration of the versatility of the proposed method, instead of considering a single mode of attack which is often impossible in real-life, we also select the attack mode specific to each task. All additional details, including the statistics of the datasets used and implementation details of the victim models and attack methods, are presented in App. \ref{app:implementation_details}. 

\paragraph{TU Datasets}

\begin{figure}[t]
    \centering
    \begin{subfigure}{0.16\linewidth}
    \includegraphics[trim=0cm 0cm 0cm  0cm, clip, width=1.0\linewidth]{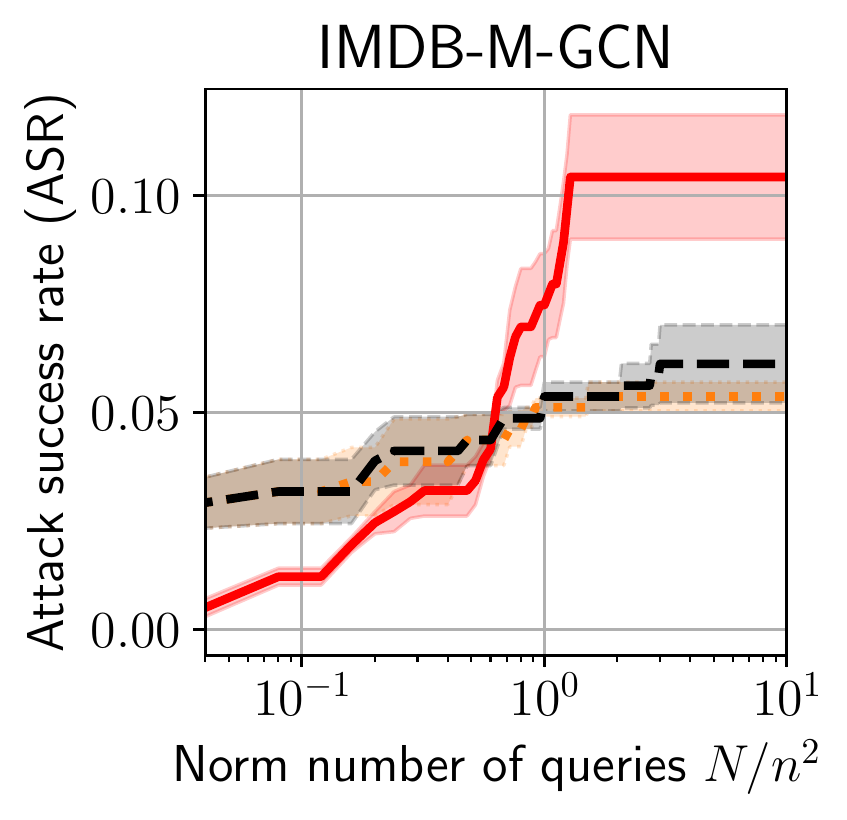}
    \end{subfigure}
         \begin{subfigure}{0.16\linewidth}
    \includegraphics[trim=0cm 0cm 0cm  0cm, clip, width=1.0\linewidth]{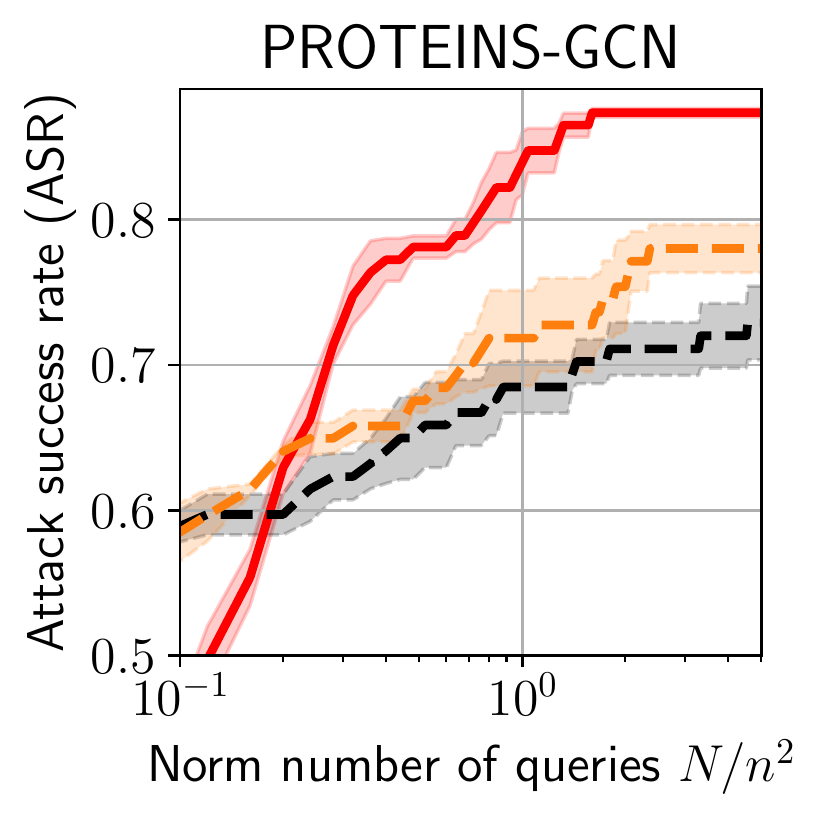}
    \end{subfigure}
    \begin{subfigure}{0.16\linewidth}
    \includegraphics[trim=0cm 0cm 0cm  0cm, clip, width=1.0\linewidth]{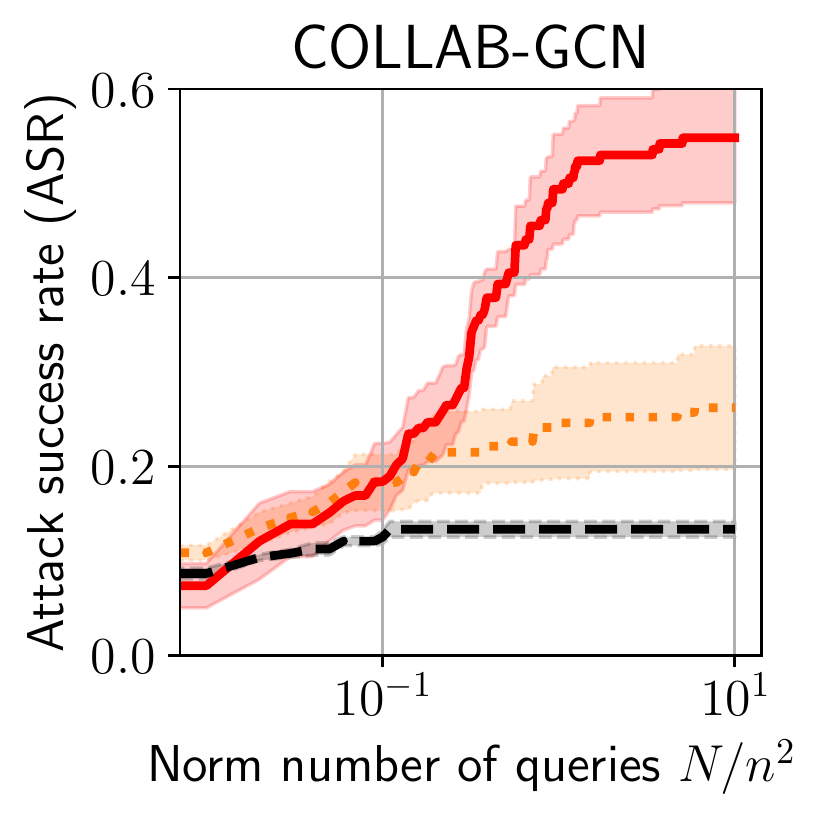}
    \end{subfigure}
       \begin{subfigure}{0.16\linewidth}
    \includegraphics[trim=0cm 0cm 0cm  0cm, clip, width=1.0\linewidth]{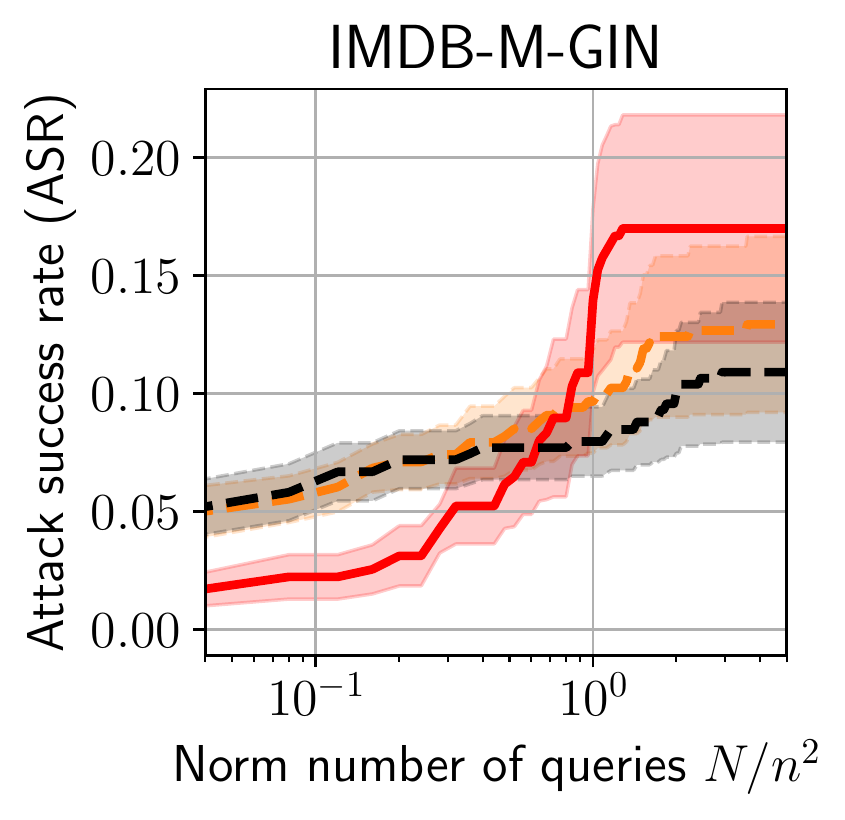}
    \end{subfigure}
       \begin{subfigure}{0.16\linewidth}
    \includegraphics[trim=0cm 0cm 0cm  0cm, clip, width=1.0\linewidth]{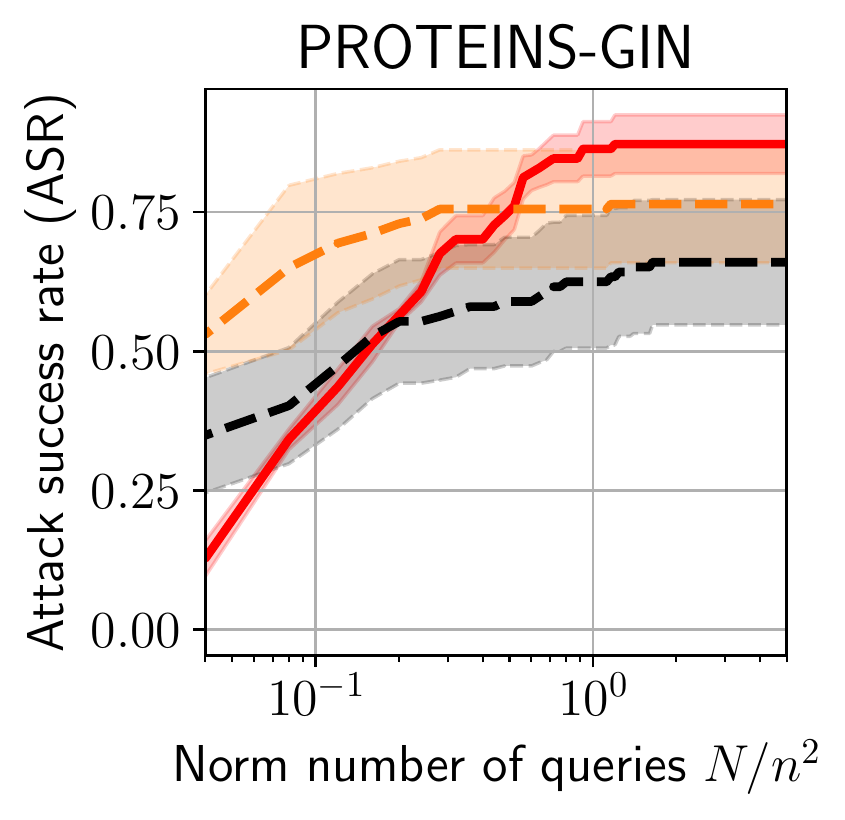}
    \end{subfigure}
         \begin{subfigure}{0.16\linewidth}
    \includegraphics[trim=0cm 0cm 0cm  0cm, clip, width=1.0\linewidth]{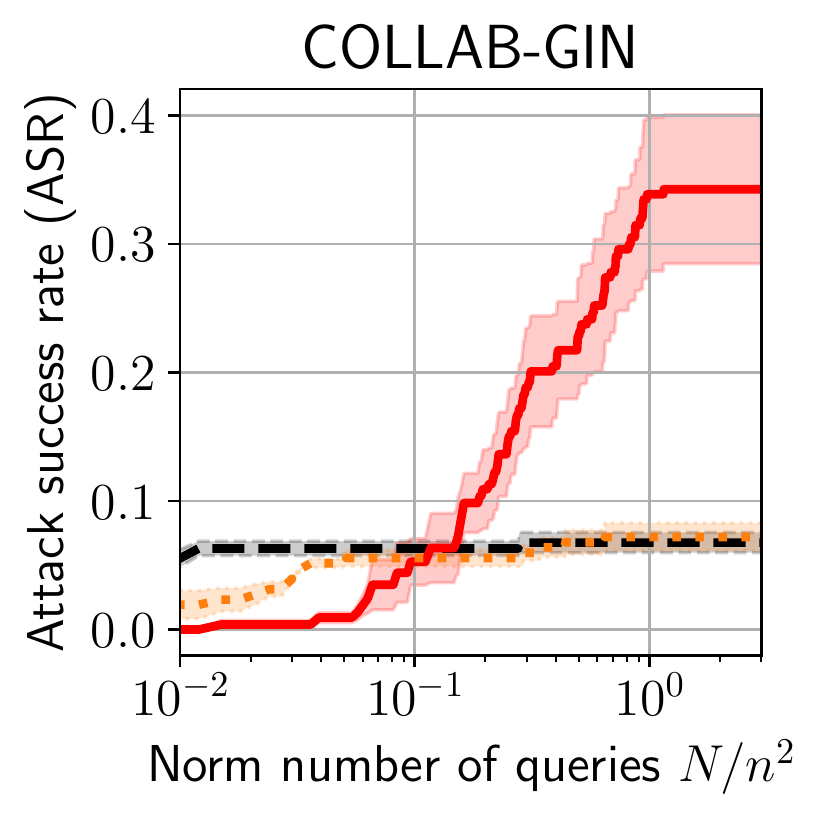}
    \end{subfigure}
    
    \begin{subfigure}{0.8\linewidth}
    \includegraphics[trim=0cm 0.5cm 0cm  0.3cm, clip, width=1.0\linewidth]{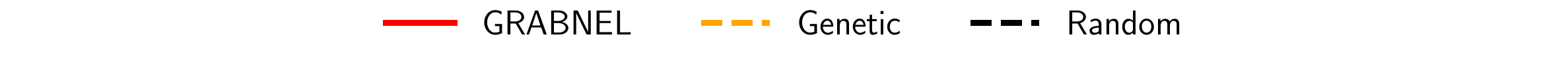}
    \end{subfigure}
    \vspace{-2mm}
    \caption{\gls{ASR} against the number of queries to \gls{GCN} and \gls{GIN} (normalised by the square of the number of nodes of each graph) of \gls{TU} datasets. Note the x-axis is on log-scale. Lines and shades denote mean $\pm 1$ sd across 3 random initialisations. \textcolor{red}{\texttt{\gls{GRABNEL}}} outperforms other attack methods considerably. \texttt{Random} and \textcolor{orange}{\texttt{Genetic}} appear to converge faster initially as they always use exploit full perturbation budget allocated, while \textcolor{red}{\texttt{\gls{GRABNEL}}} is parsimonious and only attempts a higher perturbation budget when attacks perturbing fewer edges fail. It is also possible to derive a variant of random search with such parsimony, and the readers are referred to the detailed ablation studies in App. \ref{app:ablation_studies}.
    }    
    \label{fig:asrtu}
  \vspace{-3mm}
\end{figure}

We first conduct experiments on four common \gls{TU} datasets \cite{morris2020tudataset}, namely (in ascending order of average graph sizes in the dataset) \gls{IMDB-M}, \gls{PROTEINS}, \gls{COLLAB} and \gls{REDDIT-MULTI-5K}. In all cases, unless specified otherwise, we define the attack budget $\Delta$ in terms of the maximum \emph{structural perturbation ratio} $r$ defined in \cite{chen2021graphattacker} where $\Delta \leq rn^2$. We similarly link the maximum numbers of queries $B$ allowed for individual graphs to their sizes as $B = 40\Delta$, thereby giving larger graphs and thus potentially more difficult instances higher attack\footnote{Due to computational constraints, we cap the maximum number of queries to be $2\times10^4$ on each graph.} and query budgets similar to the conventional image adversarial attack literature \cite{ru2019bayesopt}. In this work, unless otherwise specified we set $r=0.03$ for all experiments, and for comparison we consider a number of baselines, including random search, \gls{GA} introduced in \cite{dai2018adversarial}\footnote{The original implementation of \gls{RL-S2V}, the primary algorithm in \cite{dai2018adversarial}, primarily focus on a S2V-based victim model \cite{dai2016discriminative}. We compare \gls{GRABNEL} against it in the same dataset considered in \cite{dai2018adversarial} in App. \ref{sec:rls2v}.}. On some task/victim model combinations, we also consider an additional simple gradient-based method which greedily adds or delete edges based on the magnitude computed input gradient similar to the gradient based method described in \cite{dai2018adversarial} (note that this method is \emph{white-box} as access to parameter weights and gradients is required), which is also similar in spirit to methods like Nettack \cite{zugner2018adversarial}. To verify whether the proposed attack method can be used for a variety of classifier architectures we also consider various victim models: we first use \gls{GCN} \cite{kipf17semi} and \gls{GIN} \cite{xu2018how}, which are most commonly used in related works \cite{sun2018adversarial}. Considering the strong performance of hierarchical models in graph classification \cite{gao2019graph, ying2018hierarchical}, we also conduct some experiments on the Graph-U-Net \cite{gao2019graph} as a representative of such architectures. We show the classification performance of both victim models before and after attacks using various methods in Table \ref{tab:tu}, and we show the \gls{ASR} against the (normalised) number of queries in Fig. \ref{fig:asrtu}. It is worth noting that in consistency with the image attack literature, we launch and consider attacks on the \textit{graphs that were originally classified correctly}, and statistics, such as the \gls{ASR}, are also computed on that basis. We report additional statistics, such as the evolution of the attack losses as a function of number of queries of selected individual data points in App. \ref{app:more_stats_adv}.

The results generally show that the attack method is effective against both \gls{GCN} and \gls{GIN} models with \gls{GRABNEL} typically leading to the largest degradation in victim predictions in all tasks, often performing on par or better than \emph{Gradient-based}, a white-box method. It is worth noting that although \emph{Gradient-based} often performs strongly, there is no guarantee that it always does so: first, for general edge flipping problems, \emph{Gradient-based} computes gradients w.r.t. all possible edges (including those that do not currently exist) and an accurate estimation of such high
dimensional gradients can be highly difficult. Second, gradients only capture local information and they are not necessarily accurate when used to extrapolate
function value beyond that neighbourhood. However, relying on gradients to select edge perturbations
constitutes such an extrapolation, as edge addition/deletion is binary and discrete. Lastly, on the tasks with larger graphs (e.g. \gls{COLLAB} on \gls{GCN} and \gls{GIN}), due to the huge search spaces, we find neither random search nor \gls{GA} could flip predictions effectively except for some ``easy'' samples already lying close to the decision boundary; \gls{GRABNEL} nonetheless performs well thanks to the effective constraint of the search space from the sequential selection of edge perturbation, which is typically more significant on the larger graphs.

\begin{table}[t]
\centering
\caption{Test accuracy of \gls{GCN} and \gls{GIN} victim models on the \gls{TU} datasets before (\emph{clean}) and after various attack methods. Results shown in mean $\pm$ 1 standard deviation across 3 trials. The results for \gls{REDDIT-MULTI-5K} are shown in App. \ref{app:reddit_multi_5k}.}
\resizebox{0.95\linewidth}{!}{
\begin{tabular}{@{}lcccccc@{}}
\toprule
& \multicolumn{3}{c}{\gls{GCN} \cite{kipf17semi}} & \multicolumn{3}{c}{\gls{GIN}\cite{xu2018how}} \\
 & \gls{IMDB-M} & \gls{PROTEINS} & \gls{COLLAB} & \gls{IMDB-M} & \gls{PROTEINS} & \gls{COLLAB} \\ 
 \midrule
\emph{Clean} & $50.53_{\pm 1.4}$ & $71.73_{\pm 2.6}$ & $79.73_{\pm 2.1}$ & $48.85_{\pm 0.4}$ & $70.53_{\pm 2.3}$ &$80.80_{\pm 0.9}$ \\
Random & $47.43_{\pm 1.2}$ & $19.46_{\pm 1.7}$ & $67.00_{\pm 3.7}$& $40.44_{\pm 2.5}$ & $23.21_{\pm 14}$ & $73.01_{\pm 5.0}$\\
Genetic \cite{dai2018adversarial} & $47.82_{\pm 1.5}$& $14.88_{\pm 1.7}$ & $ 58.61_{\pm 7.9}$& $39.68_{\pm 3.1}$ & $15.47_{\pm 10}$ &  $72.34_{\pm 2.5}$ \\
Gradient-based$^{\dagger}$ & $\mathbf{39.31_{\pm 2.2}}$& $50.60_{\pm 4.5}$ &
$36.67_{\pm 1.2}$ & $\mathbf{37.56_{\pm 2.2}}$ & $11.90_{\pm 4.4}$ & $\mathbf{54.00_{\pm 2.9}}$ \\
\midrule
\gls{GRABNEL} (ours) & $45.23_{\pm 0.2}$& $\mathbf{10.82_{\pm 2.5}}$ &  $\mathbf{35.38_{\pm 9.3}}$& $38.22_{\pm 3.9}$ & $\mathbf{10.72_{\pm 6.3}}$ & $57.33_{\pm 4.7}$\\
\bottomrule
\multicolumn{5}{l}{$^{\dagger}$: White-box method}
\label{tab:tu}
\end{tabular}
}
\vspace{-6mm}
\end{table}

\begin{figure}[t]
\centering
\hfill
\begin{minipage}{0.28\textwidth}
\includegraphics[width=\textwidth]{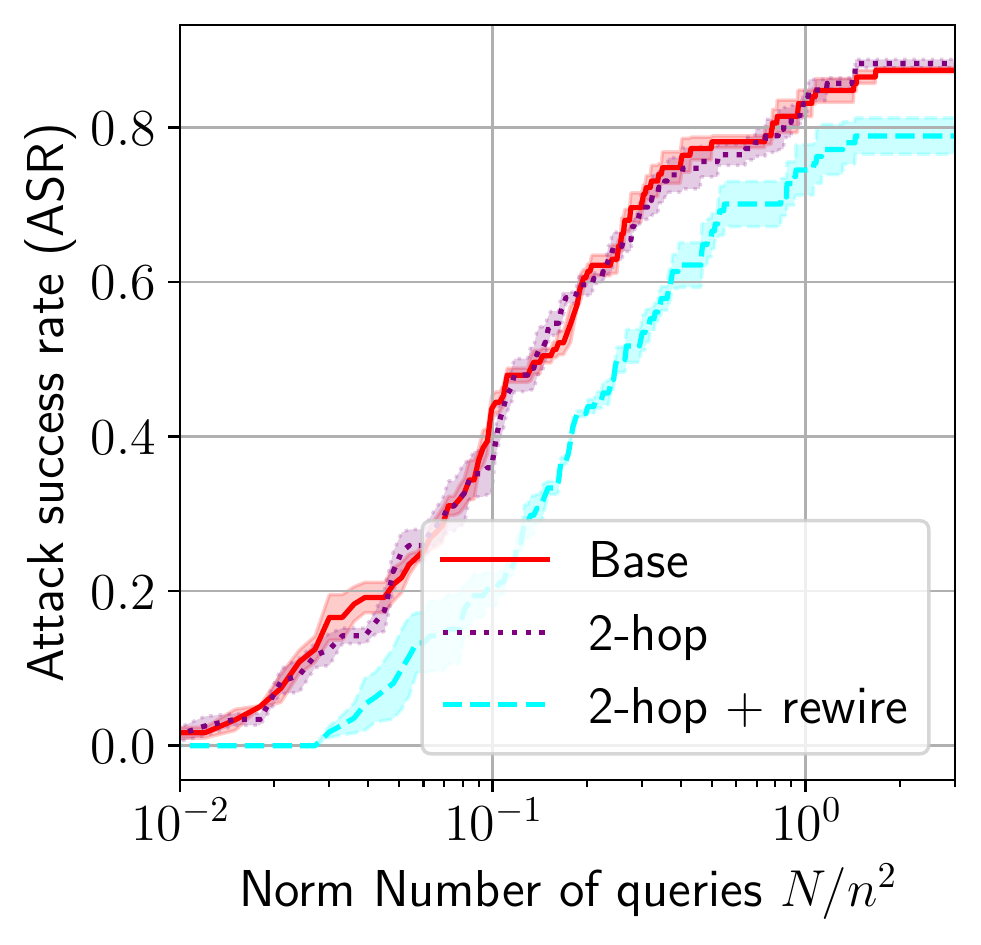}
\vspace{-3mm}
\caption{\gls{ASR} vs normalised \# queries with constraints on a \gls{GCN} model trained on \gls{PROTEINS}.}
\label{fig:constraints}
\end{minipage} \hfill
\begin{minipage}{0.3\textwidth}
\includegraphics[width=\textwidth]{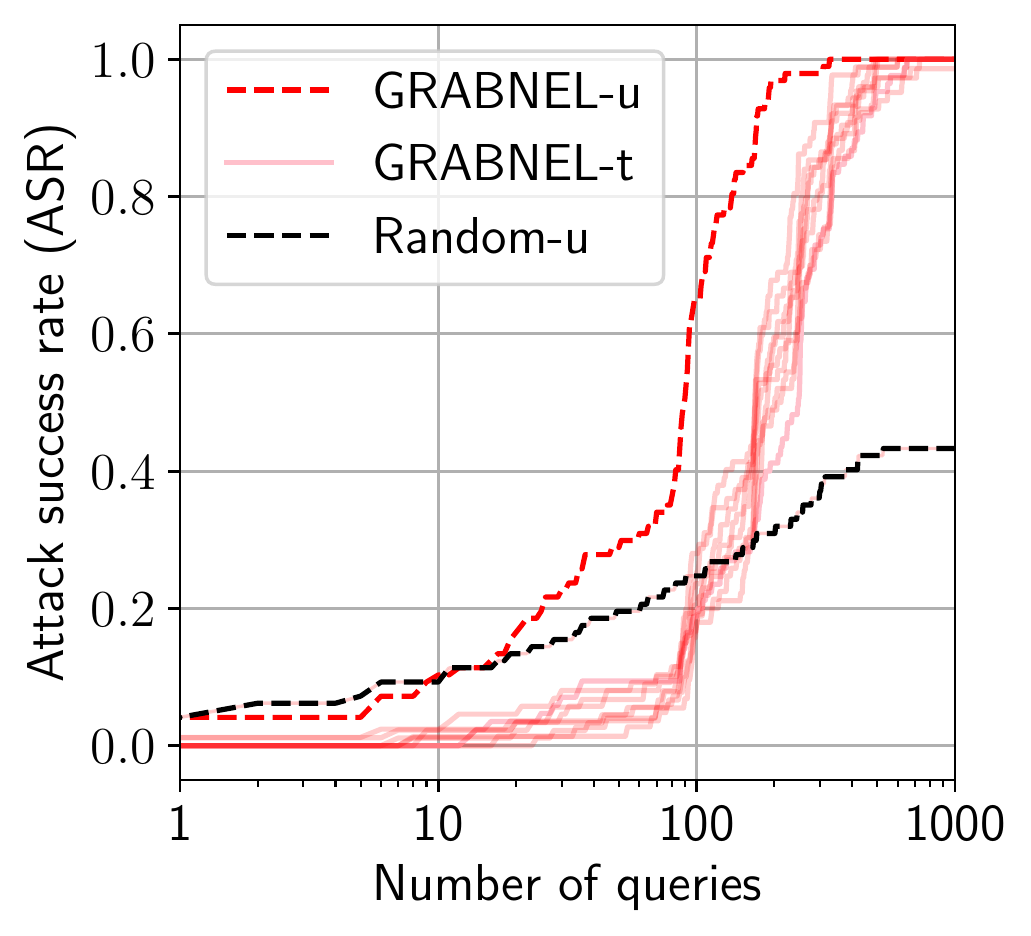}
\vspace{-3mm}
\caption{\gls{ASR} vs \# queries on a Cheby\gls{GIN} trained on \gls{MNIST}-75sp on targeted and untargeted attack setups.}  
\label{fig:mnist}
\end{minipage} \hfill
\begin{minipage}{0.28\textwidth}
\includegraphics[width=\textwidth]{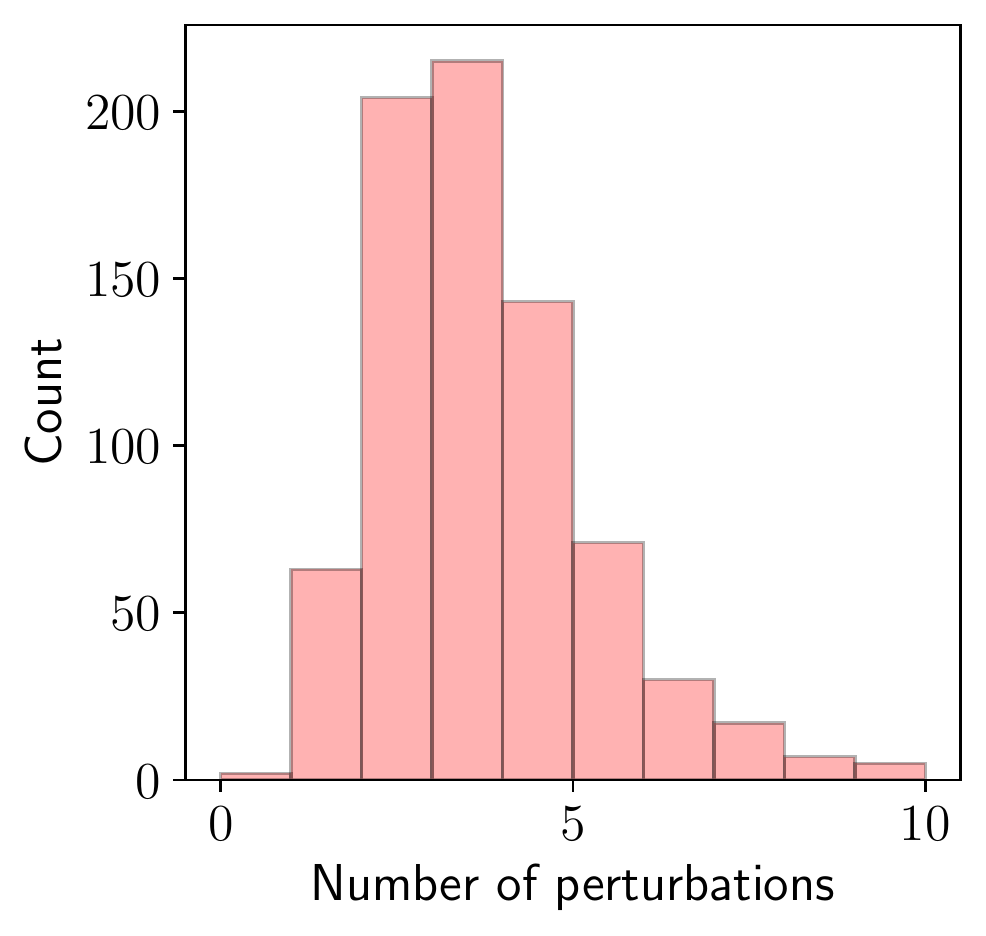}
\vspace{-3mm}
\caption{Histogram of \#edges swapped in successfully attacked \gls{MNIST}-75sp instances by \textcolor{pink}{\texttt{\gls{GRABNEL}-t}}.}   
\label{fig:mnist_histogram}
\end{minipage} \hfill
\vspace{-6mm}
\end{figure}

\begin{wraptable}{r}{0.35\textwidth}
\vspace{-4mm}
\centering
\caption{Test accuracy of Graph U-net \cite{gao2019graph} on \gls{IMDB-M} and \gls{PROTEINS} before and after attack.}
\resizebox{\linewidth}{!}{
\begin{tabular}{@{}lcccccc@{}}
\toprule
 & \gls{IMDB-M} & \gls{PROTEINS} \\ 
 \midrule
\emph{Clean} & $55.33$ & $79.46$  \\
Random & $45.33$ & $75.00$ \\
Genetic \cite{dai2018adversarial} & $44.00$ & $75.00$ \\
\midrule
\gls{GRABNEL} (ours) & $\mathbf{41.33}$& $\mathbf{58.80}$ \\
\bottomrule
\label{tab:tu_gunet}
\end{tabular}
}
\vspace{-10mm}
\end{wraptable}

We report the results on the Graph U-net victim model in Table \ref{tab:tu_gunet}: as expected, Graph U-net performs better in terms of clean classification accuracy compared to the \gls{GCN} and \gls{GIN} models considered above, and it also seem more robust to all types adversarial attacks on the \gls{PROTEINS} dataset. Nonetheless, in terms of relative performance margin, \gls{GRABNEL} still outperforms both baselines considerably, demonstrating the flexibility and capability for it to conduct effective attacks even on the more complicated and realistic victim models.

As discussed, in real life, adversarial agents might encounter additional constraints other than the number of queries to the victim model or the amount of perturbation introduced. To demonstrate that our framework can handle such constraints, %
we further carry out attacks on victim models using identical protocols as above but with a variety of additional constraints considered in several previous works. Specifically, the scenarios considered, in the ascending order of restrictiveness, are:
\begin{itemize}[leftmargin=0.15in, noitemsep, topsep=0.05pt]
\item \texttt{Base}: The base scenario is identical to the setup in Table \ref{tab:tu} and Fig \ref{fig:asrtu};
\item \texttt{2-hop}: Edge addition between nodes $(u, v)$ is only permitted if $v$ is within 2-hop distance of $u$;
\item \texttt{2-hop+rewire} \cite{ma2019attacking}: Instead of flipping edges, the adversarial agent is only allowed to rewire from nodes $(u, v)$ (where an edge exists) to $(u, w)$ (where no edge currently exists). Node $w$ must be within 2-hop distance of $u$;
\end{itemize}
We test on the \gls{PROTEINS} dataset, and show the results in Fig. \ref{fig:constraints}: interestingly, the imposition of the 2-hop constraint itself leads to no worsening of performance -- in fact, as we elaborate in Sec. \ref{sec:attackanalysis}, we find the phenomenon of adversarial edges remaining relatively clustered within a relatively small neighbourhood is a general pattern in many tasks. This implies that the 2-hop condition, which constrains the spatial relations of the adversarial edges, might already hold even without explicit specification, thereby explaining the marginal difference between the base and the 2-hop constrained cases in Fig. \ref{fig:constraints}. While the additional rewiring constraint leads to (slightly) lower attack success rates, the performance of \gls{GRABNEL} remains relatively robust in all scenarios considered. 

\paragraph{Image Classification} Beyond the typical ``edge flipping'' setup on which existing research has been mainly focused, we now consider a different setup involving attacks on the \gls{MNIST}-75sp dataset \cite{lecun1989backpropagation, Knyazev2019} with weighted graphs with continuous attributes -- note that . The dataset is generated by first partitioning \gls{MNIST} image into around 75 superpixels with \textsc{slic} \cite{achanta2012slic, Dwivedi2020} as the graph nodes (with average superpixel intensity as node attributes). The pairwise distances between the superpixels form the edge weights. We use the pre-trained Cheby\gls{GIN} with attention model released by the original authors \cite{Knyazev2019} (with an average validation classification accuracy of around $95\%$) as the victim model. 
Given that the edge values are no longer binary, simply flipping the edges (equivalent to setting edge weights to $0$ and $1$) is no longer appropriate. To generalise the sparse perturbation setup and inspired by edge rewiring studied by previous literature, we instead adopt an attack mode via \emph{swapping edges}: each perturbation can be defined by 3 end nodes $(u, v, s)$ where edge weights $w_{uv}$ is swapped with edge weight $w(u, s)$. We show the results in Fig. \ref{fig:mnist}:  \textcolor{red}{\texttt{\gls{GRABNEL}-u}} and \texttt{Random-u} denotes the \gls{GRABNEL} and random search under the \textit{untargeted} attack, respectively, whereas \textcolor{pink}{\texttt{\gls{GRABNEL}-t}} denotes \gls{GRABNEL} under the \textit{targeted attack} with each line denoting 1 of the 9 possible target classes in \gls{MNIST}. We find that \gls{GRABNEL} is surprisingly effective in attacking this victim model, almost completely degrading the victim (Fig. \ref{fig:mnist}) with very few swapping operations (Fig. \ref{fig:mnist_histogram}) even in the more challenging targeted setup. This seems to suggest that, at least for the data considered, the victim model is very brittle towards carefully crafted edge swapping, with its predictive power seemingly hinged upon a very small number of key edges. We believe a thorough analysis of this phenomenon is of an independent interest, which we defer to a future work.

\begin{figure}[t]
    \centering
    \begin{subfigure}{0.3\linewidth}
    \includegraphics[trim=0cm 0cm 0cm  0cm, clip, width=1.0\linewidth]{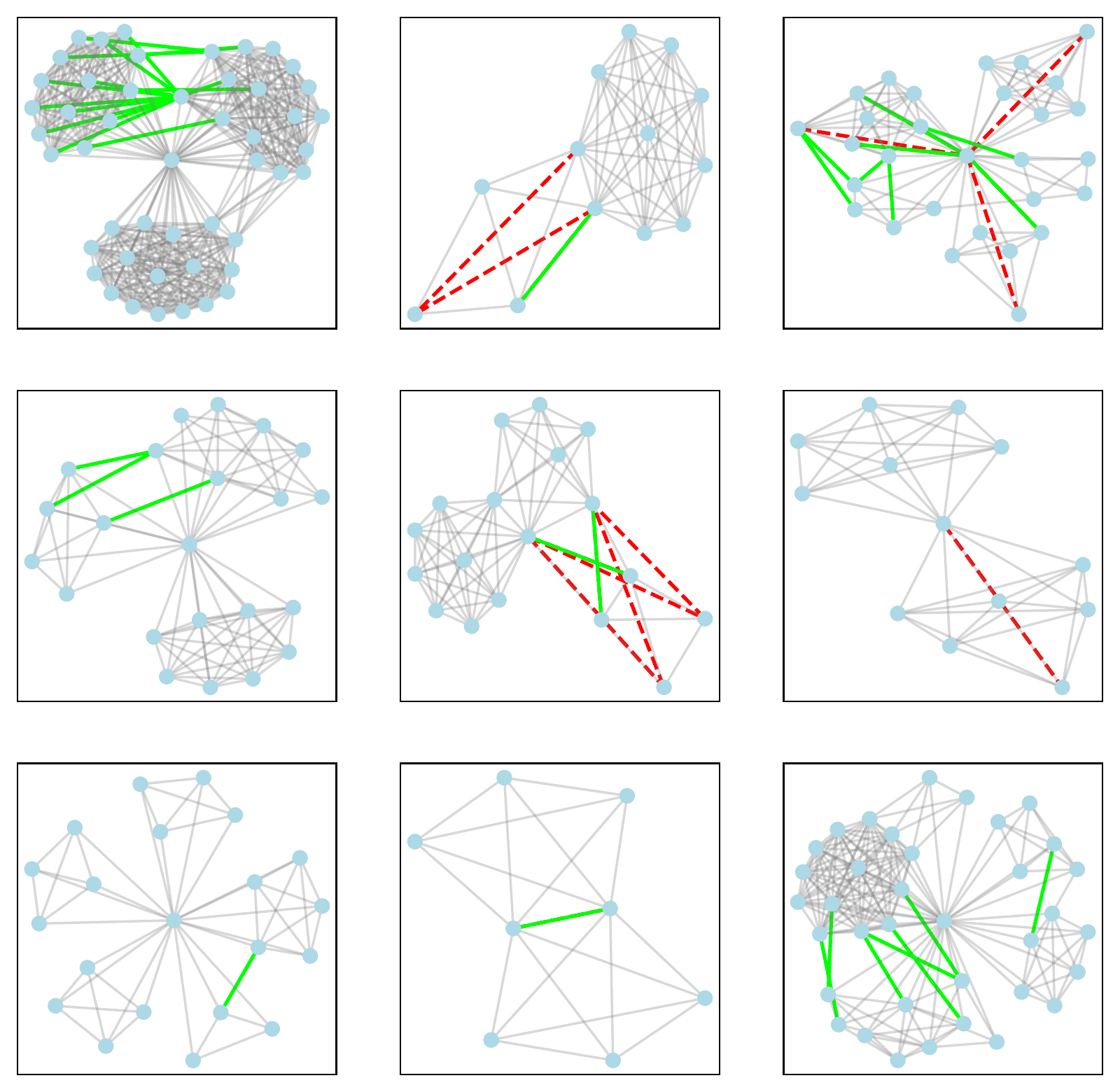}
        \caption{\gls{IMDB-M} (\gls{GCN})}
    \end{subfigure}
    \quad
         \begin{subfigure}{0.3\linewidth}
    \includegraphics[trim=0cm 0cm 0cm  0cm, clip, width=1.0\linewidth]{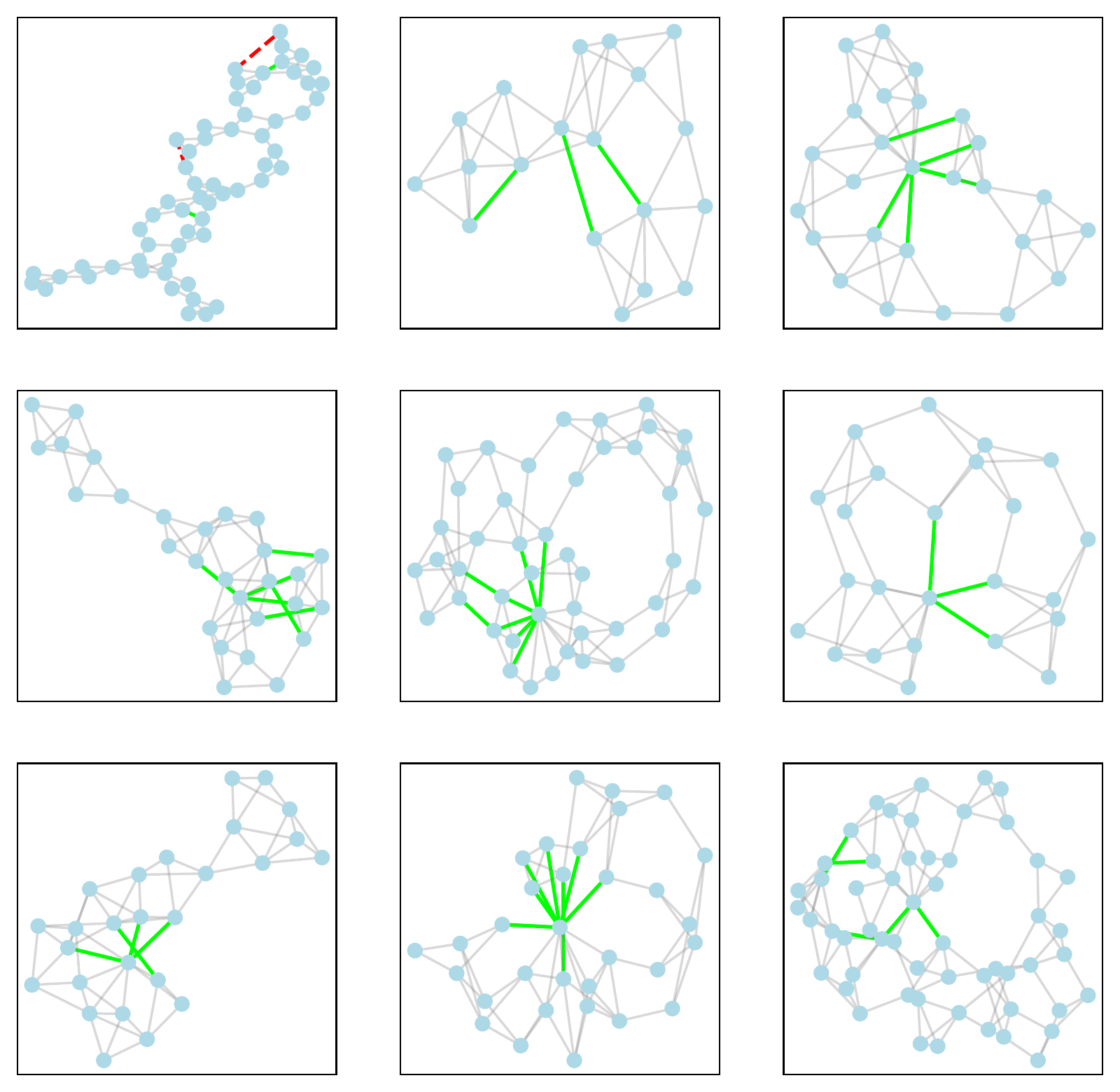}
    \caption{\gls{PROTEINS} (\gls{GCN})}
    \end{subfigure}
        \quad
    \begin{subfigure}{0.3\linewidth}
    \includegraphics[trim=0cm 0cm 0cm  0cm, clip, width=1.0\linewidth]{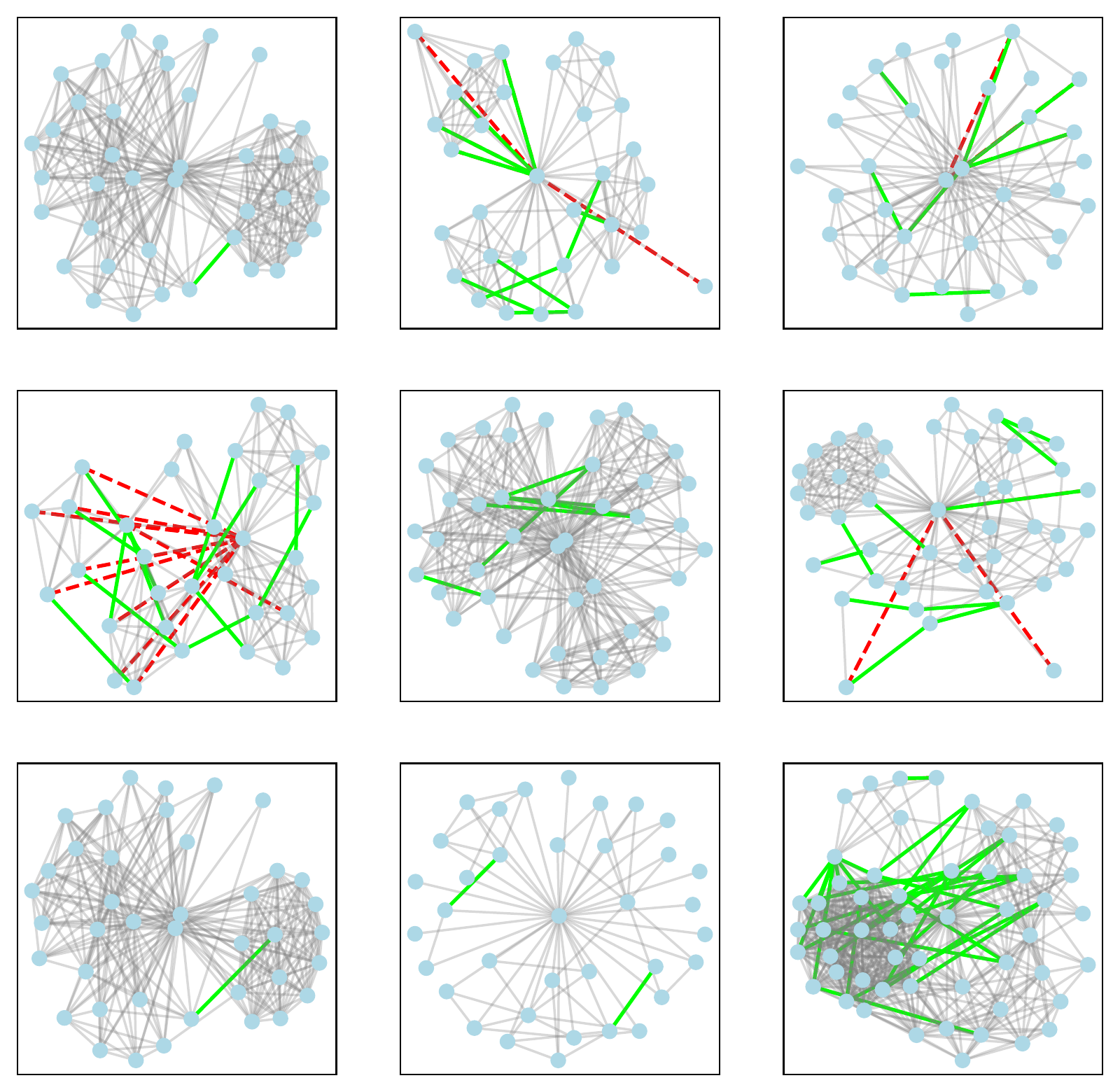}
    \caption{\gls{COLLAB} (\gls{GCN})}
    \end{subfigure}
    \centering
       \begin{subfigure}{0.3\linewidth}
    \includegraphics[trim=0cm 0cm 0cm  0cm, clip, width=1.0\linewidth]{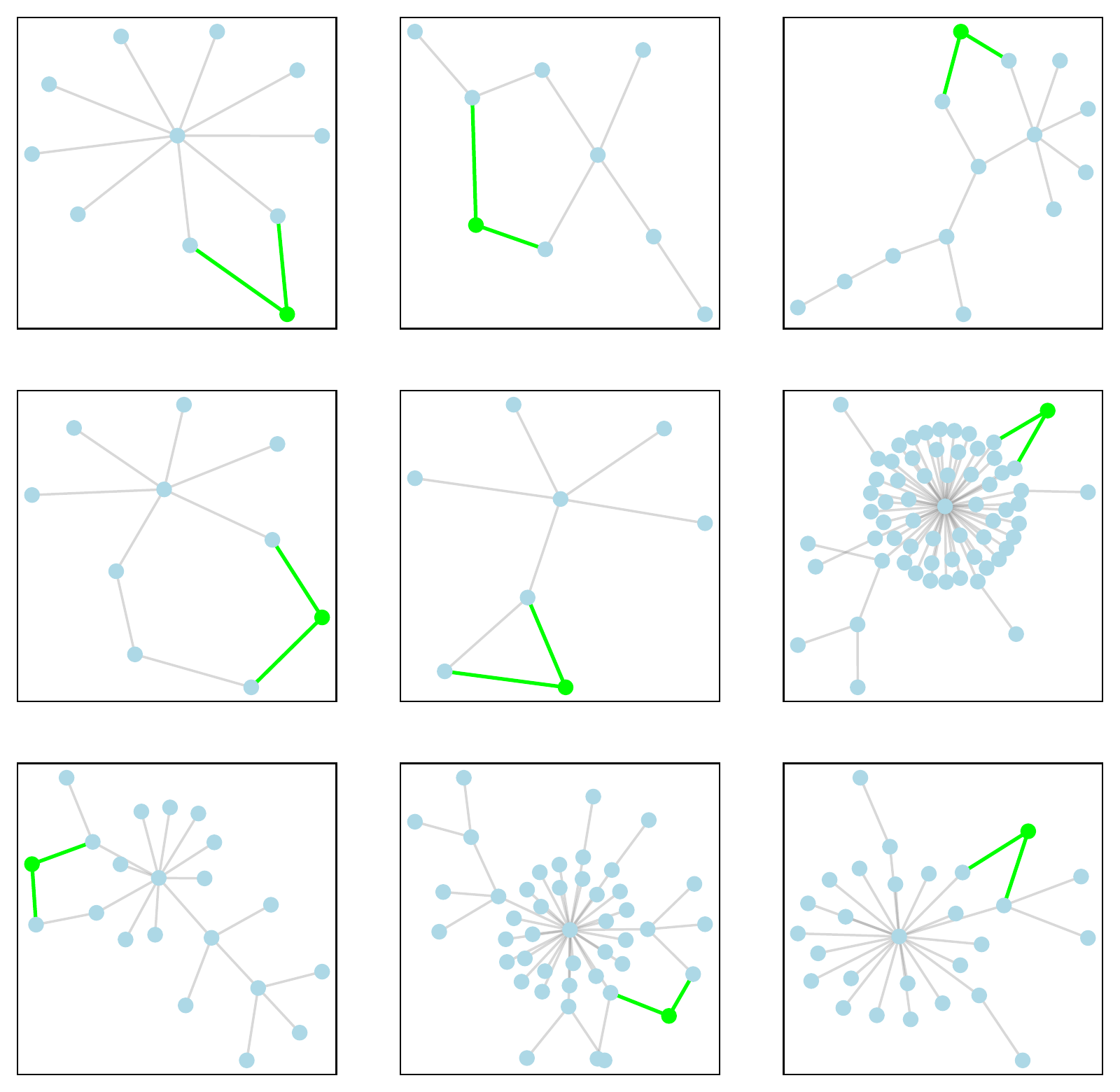}
    \caption{Twitter fake news}
    \end{subfigure}
        \quad
       \begin{subfigure}{0.3\linewidth}
    \includegraphics[trim=0cm 0cm 0cm  0cm, clip, width=1.0\linewidth]{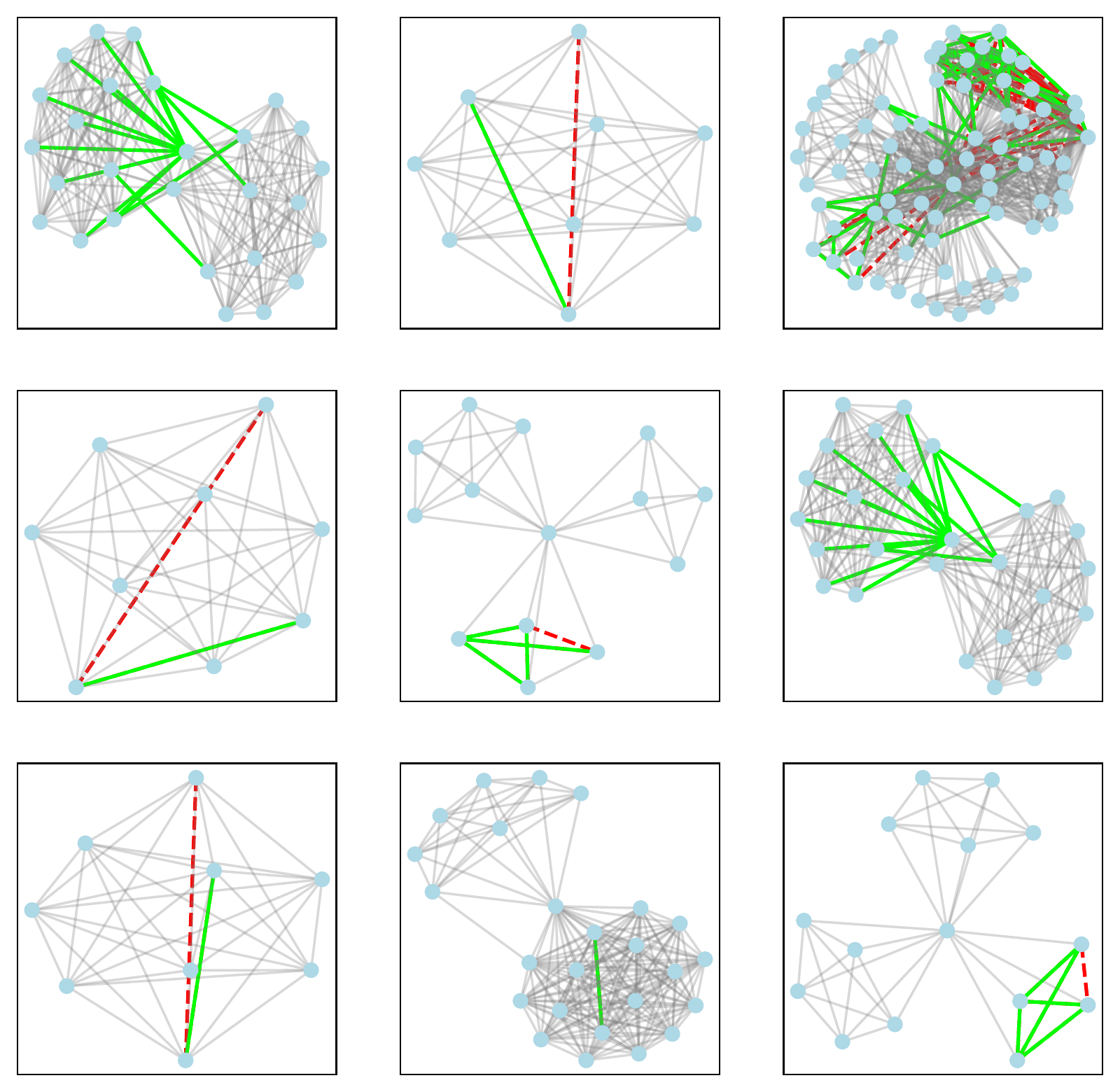}
        \caption{\gls{IMDB-M} (\gls{GIN})}
    \end{subfigure}
        \quad
         \begin{subfigure}{0.3\linewidth}
    \includegraphics[trim=0cm 0cm 0cm  0cm, clip, width=1.0\linewidth]{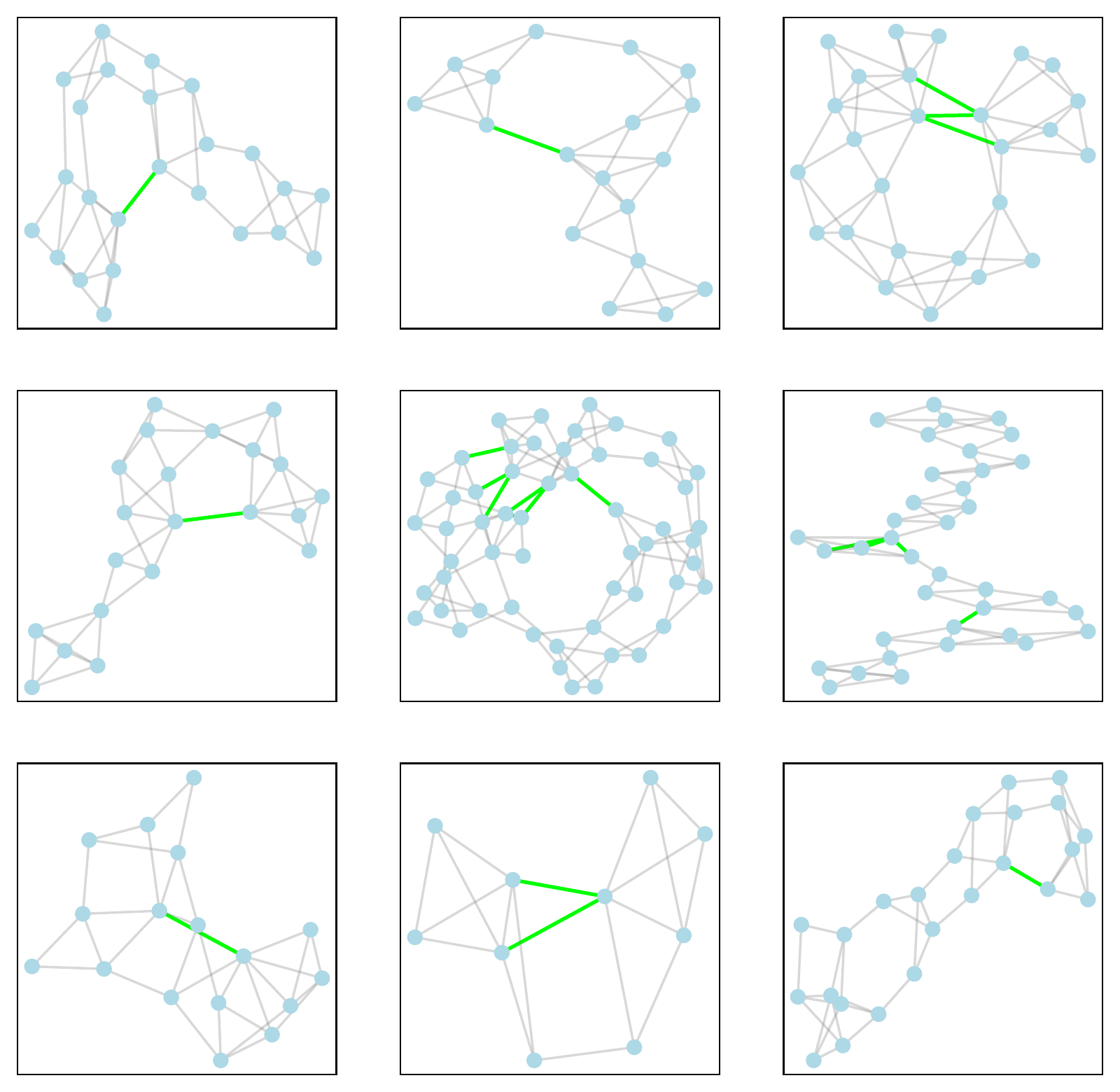}
    \caption{\gls{PROTEINS} (\gls{GIN})}
    \end{subfigure}
    \vspace{-2mm}
    \caption{Adversarial examples found by \gls{GRABNEL}. \textcolor{red}{Red edges} denote deleted edges and \textcolor{green}{green edges} indicate added ones. In Twitter fake news detection task, \textcolor{green}{green nodes/edges} denote the injected nodes and their connections to the existing graphs. Refer to App \ref{app:more_stats_adv} for more examples.
    }    
    \label{fig:advexamples}
  \vspace{-5mm}
\end{figure}

\paragraph{Fake news detection} As a final experiment, we consider a real-life task of attacking a \gls{GCN}-based fake news detector trained on a labelled dataset in \cite{vosoughi2018spread}. Each discussion cascade (i.e. a chain of tweets, replies and retweets) is represented as an undirected graph, where each node represents a Twitter account (with node features being the key properties of the account such as age and number of followers/followees; see App. \ref{app:implementation_details} for details) and each edge represents a reply/retweet. As a reflection of what a real-life adversary may and may not do, we note that modifying the connections or properties of the existing nodes, which correspond to modifying existing accounts and tweets, is considered impractical and prohibited. 
\begin{wrapfigure}{r}{0.3\textwidth}
\vspace{-5mm}
  \begin{center}
    \includegraphics[width=0.3\textwidth]{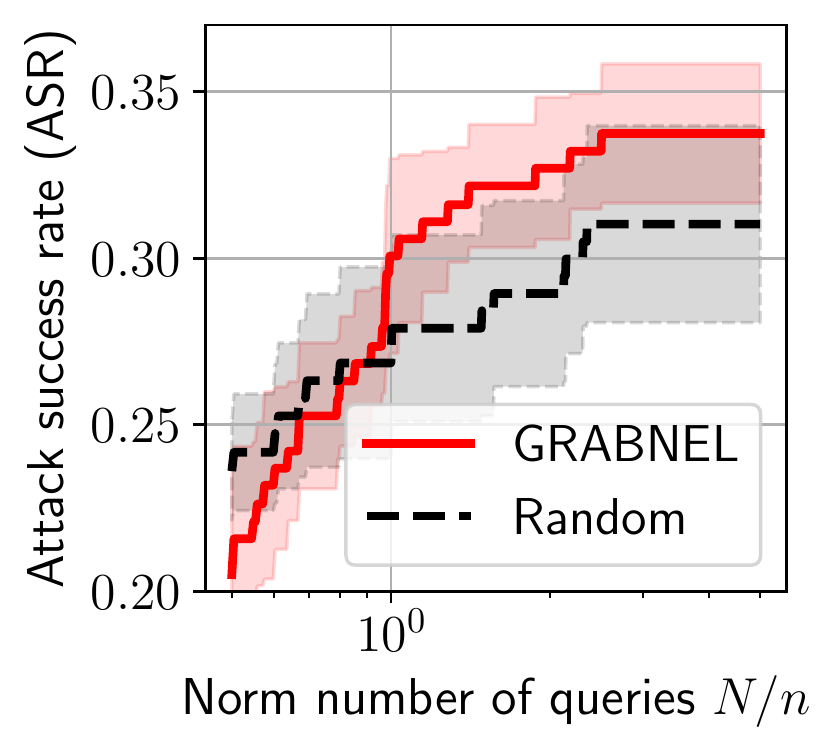}
  \end{center}
  \vspace{-5mm}
  \caption{\gls{ASR} vs. \#queries (normalised by the number of nodes, since the attack involves node injection) on the Twitter dataset.}
  \label{fig:twitter}
\end{wrapfigure}
Instead, we consider a \emph{node injection} attack mode (i.e. creating new malicious nodes and connect them to existing ones): injecting nodes is equivalent to creating new Twitter accounts and connecting them to the rest of the graph is equivalent to retweeting/replying existing accounts. We limit the maximum number of injected nodes to be $0.05N$ and the maximum number of new edges that may be created per each new node is set to the average number of edges an existing node has -- in this context, this limits the number of re-tweets and replies the new accounts may have to avoid easy detection. For the injected node, we initialise its node features in a way that reflects the characteristics of a new Twitter user (we outline the detailed way to do so in App. \ref{app:implementation_details}). We show the result in Fig. \ref{fig:twitter}, where \gls{GRABNEL} is capable of reducing the effectiveness of a \gls{GCN}-based fake news classifier by a third. In this case \emph{Random} also performs reasonably well, as the discussion cascade is typically small, allowing any adversarial examples to be exhaustively found eventually.

\paragraph{Ablation Studies} \gls{GRABNEL} benefits from a number of design choices and it is important to understand the relative contribution of each to the performance. We find that in \emph{some} tasks \gls{GRABNEL} without surrogate (i.e. random search with sequential perturbation selection. We term this variant \emph{SequentialRandom}) is a very strong baseline in terms of final \gls{ASR}, although the full \gls{GRABNEL} is much better in terms of overall performance, sample efficiency and the ability to produce successful examples with few perturbations. The readers are referred to our ablation studies in App. \ref{app:ablation_studies}.

\paragraph{Runtime Analysis} Given the setup we consider (sample-
efficient black-box attack with minimal amount of perturbation), the cost of the algorithm should not only be considered from the viewpoint of the computational
runtime of the attack algorithm itself alone, and this is a primary reason why we use the (normalised) number of queries as the main cost criterion. Nonetheless, a runtime analysis is still informative which we provide in App. \ref{app:runtime_analysis}. We find that \gls{GRABNEL} maintains a reasonable overhead even on, e.g., graphs with $\sim 10^3$ nodes/edges that are larger than most graphs in typical graph classification tasks.

\section{Attack Analysis}
\label{sec:attackanalysis}
Having established the effectiveness of our method, in this section we provide a qualitative analysis on the common interpretable patterns behind the adversarial samples found, which provides further insights into the robustness of graph classification models against structural attacks. We believe such analysis is especially valuable, as it may facilitate the development of even more effective attack methods, and may provide insights that could be useful for identification of real-life vulnerabilities for more effective defence. We show examples of the adversarial samples in Fig. \ref{fig:advexamples} (and Fig. \ref{fig:more_advexamples} in App. \ref{app:more_stats_adv}). We summarise some key findings below.

\begin{itemize}[leftmargin=0.1in, noitemsep, topsep=0.02pt]
 \item \emph{Adversarial edges tend to cluster closely together}: We find the distribution of the adversarial edges (either removal or addition) in a graph to be highly uneven, with many adversarial edges often sharing common end-nodes or having small spatial distance to each other. This is empirically consistent with recent theoretical findings on the stability of spectral graph filters in \cite{kenlay2021interpretable}. %
 From an attacker point of view, this may provide a ``prior'' on the attack to constrain the search space, as the regions around existing perturbations should be exploited more; we leave a practical investigation of the possibility of leveraging this to enhance attack performance to a future work.
 
 \item \emph{Adversarial edges often attempt to destroy or modify community structures}: for example, the original graphs in the \gls{IMDB-M} dataset can be seen to have community structure, a graph-level topological property that is distinct from the existing works analysing attack patterns on node-level tasks \cite{wu2019adversarial, zugner2020adversarial}. When the \gls{GCN} model is attacked, the attack tends to flip the edges \emph{between} the communities, and thereby destroying the structure by either merging communities or deleting edges within a cluster. On the other hand, the \gls{GIN} examples tend to strengthen the community structures by adding edges within clusters and deleting edges between them. With similar observations also present in, for example, \gls{PROTEINS} dataset, this may suggest that the models may be fragile to modification of the community structure. 
 
 \item \emph{Beware the low-degree nodes!} While low-degree nodes are important in terms of degree centrality, we find some victim models are vulnerable to manipulations on such nodes. Most prominently, in the Twitter fake news example, the malicious nodes almost never connect directly to the central node (original tweet) but instead to a peripheral node. This finding corroborates the theoretical argument in \cite{kenlay2021interpretable} which shows that spectral graph filters are more robust towards edge flipping involving \emph{high-degree} nodes than otherwise, and is also consistent with observations on node-level tasks \cite{zugner2020adversarial} with the explanation being lower-degree nodes having larger influence in the neighbourhood aggregation in \gls{GCN}. Nonetheless, we note that changes in a higher-degree node are likely to cascade to more nodes in the graph than low degree nodes, and since graph classifiers aggregate across all nodes in the readout layer the indirect change of node representations also matter. Therefore, we argue that this phenomenon in graph classification is still non-trivial.

\end{itemize}

\section{Conclusion}
\label{sec:conclusion}
\paragraph{Summary} This work proposes a novel and flexible black-box method to attack graph classifiers using Bayesian optimisation. We demonstrate the effectiveness and query efficiency of the method empirically. Unlike many existing works, we qualitatively analyse the adversarial examples generated. We believe such analysis is important to the understanding of adversarial robustness of graph-based learning models. Finally, we would like to point out that a potential negative social impact of our work is that bad actors might use our method to attack real-world systems such as a fake new detection system on social media platforms. Nevertheless, we believe that the experiment in our paper only serves as a proof-of-concept and the benefit of raising awareness of vulnerabilities of graph classification systems largely outweighs the risk.
\paragraph{Limitations and Future Work} Firstly, the current work only considers topological attack, although the surrogate used is also compatible with attack on node/edge features or hybrid attacks. Secondly, while we have evaluated several mainstream victim models, it would also be interesting to explore defences against adversarial attacks and to test \gls{GRABNEL} in robust \gls{GNN} setups such as those with advanced graph augmentations \cite{you2020graph}, randomised smoothing \cite{zhang2020backdoor, gao2020certified} and adversarial detection \cite{chen2020adversarial}. Lastly, the current work is specific to graph classification; we believe it is possible to adapt it to attack other graph tasks by suitably modifying the loss function. We leave these for future works. 

\section*{Acknowledgement and Funding Disclosure}
The authors would like to acknowledge the following sources of funding in direct support of this work: XW and BR are supported by the Clarendon Scholarship at University of Oxford; HK is supported by the EPSRC Centre for Doctoral Training in Autonomous Intelligent Machines and Systems EP/L015897/1; AB thanks the Konrad-Adenauer-Stiftung and the Oxford-Man Institute of Quantitative Finance for their support. The authors would also like to thank the Oxford-Man Institute of Quantitative Finance for providing the computing resources necessary for this project. The authors declare no conflict in interests.

\bibliographystyle{plain}

\bibliography{references}

\clearpage
\appendix

\section*{{\Large APPENDICES of \textit{Adversarial Attacks on Graph Classification via Bayesian Optimisation}}}

\section{Algorithms}
\label{app:algorithms}

The overall algorithm of \gls{GRABNEL} is shown in Algorithm \ref{alg:grabnel}.

\begin{algorithm}[H]
\begin{scriptsize}
	    \caption{Overall pseudocode of the \gls{GRABNEL} routine.}
	    \label{alg:grabnel}
	\begin{algorithmic}[1]
		\STATE {\bfseries Input:}  Original graph $\mathcal{G}_0$, victim model $f_{\theta}$, $n_{\mathrm{init}}$ (the number of random initialising points), Query budget $B$, Perturbation budget $\Delta$.
		\STATE {\bfseries Output:} An adversarial graph $\mathcal{G}^*$
		\STATE Set base graph $\mathcal{G}_{\mathrm{base}} \leftarrow \mathcal{G}_0$; initialise stage count $\texttt{stage} \leftarrow 0$.
		\STATE Randomly sample $n_{\mathrm{init}}$ perturbed graphs $\{\mathcal{G}'\}_{i=1}^{n_{\mathrm{init}}}$ that are 1 edit distance different from $\mathcal{G}$ and query each perturbed graph to obtain their attack losses $\mathcal{L}_{\mathrm{attack}}(f_{\theta}, \mathcal{G}'_i)$ (these random samples are counted towards the query budget of the current stage).
		\STATE Compute the WL feature encoding for all graphs: $(\Phi(\mathcal{G}'_1), \ldots, \Phi(\mathcal{G}'_{n_{\textnormal{init}}})) = \texttt{WLFeatureExtract}(\G_0, (\G'_1, \ldots \G'_{n_{\textnormal{init}}}))$. \hfill \emph{\textcolor{gray}{// See App. \ref{app:wl_extractor} for details of \texttt{WLFeatureExtract}.}}
		\STATE Fit the sparse Bayesian linear regression surrogate with the data 
		$\{ \Phi(\mathcal{G}'_i), \mathcal{L}_{\mathrm{attack}}(f_{\theta}, \mathcal{G}'_i)\}_{i=1}^{n_{\mathrm{init}}}$
		\STATE Divide total budget of $B$ into $\Delta$ stages  \hfill \emph{\textcolor{gray}{// See ``Sequential perturbation selection''}}
		\WHILE{query budget is not exhausted and attack has not succeeded}
		\IF{query budget of the \emph{current stage} is exhausted}
		\STATE Increment the stage count $\texttt{stage} \leftarrow \texttt{stage} + 1$ and update the base graph $\mathcal{G}_{\mathrm{base}}$ with the graph leading to largest increase in attack loss in the previous stage.  \hfill \emph{\textcolor{gray}{// Refer to Fig. 1}}
		\ENDIF
		\STATE Propose graph to be queried next $\mathcal{G}'_\textnormal{proposal}$ via acquisition optimisation. \hfill \emph{\textcolor{gray}{// See ``Optimisation of acquisition function''}}
		\STATE Query $f_{\theta}$ for the graph proposed in the previous step to calculate its attack loss.
		\IF{attack succeeded}	
		\STATE Set $\mathcal{G^*} \leftarrow \mathcal{G}'_\textnormal{proposal}$ and \textbf{return} it.
		\ENDIF
		\STATE Augment the observed data: $\mathcal{D} \leftarrow \mathcal{D} \cup \{\mathcal{G}'_\textnormal{proposal}, \mathcal{L}_{\mathrm{attack}}(f_{\theta}, \mathcal{G}'_\textnormal{proposal}) \}$, update the \gls{WL} feature encodings of \emph{all} observed graphs $(\Phi(\mathcal{G}'_1), \ldots, \Phi(\mathcal{G}'_{|\mathcal{D}|})) = \texttt{WLFeatureExtract}(\G_0, (\G'_1, \ldots \G'_{|\mathcal{D}|}))$ and re-fit the surrogate.
		\ENDWHILE 
	\STATE \textbf{return} \texttt{None} \hfill \emph{\textcolor{gray}{// Failed attack within the query budget}}
	\end{algorithmic}
	
\end{scriptsize}
\end{algorithm}

\section{WL feature extractor}
\label{app:wl_extractor}
\begin{figure}[h]
    \centering
    \includegraphics[trim=0cm 0cm 0cm  0cm, clip, width=0.8\linewidth]{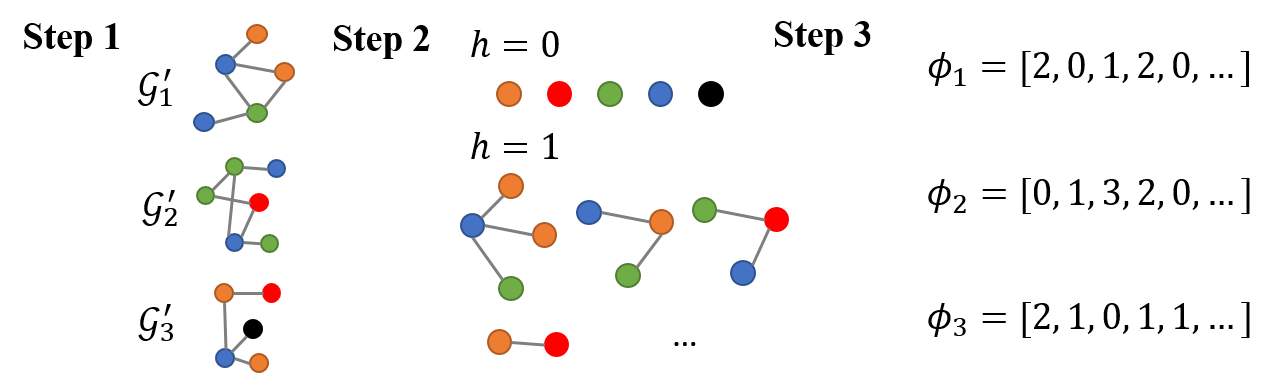}
  
    \caption{Illustration of the \gls{WL} extractor. Consider an example of an input of three graphs to the extractor $\{\mathcal{G}'_{1}, \mathcal{G}'_{2}, \mathcal{G}'_{3}\}$ with colors representing the different (discrete) node labels and we would like to compute \texttt{WLFeatureExtract}$(\{\mathcal{G}'_{i}, \{\mathcal{G}'_{1}, \mathcal{G}'_{2}, \mathcal{G}'_{3}\}) \text{ } \forall i$ (\textbf{Step 1}). The extractor module takes in 2 arguments, as the second argument consisting of the set of all input graphs is used to generate a collection of all possible Weisfeiler-Lehman features seen in \emph{all input graphs} (\textbf{Step 2}), up to $H \in 
    \mathbb{N}^+$ where $H$ is the number of \gls{WL} iterations specified. This step involves computing the Weisfeiler-Lehman embedding on all of the input graphs using the routine introduced in \cite{shervashidze2011weisfeiler}. The extractor finally counts the number of each features present from Step 2 and outputs the feature vector (\textbf{Step 3}; only $h=0$ part of the feature vector is shown in the figure -- note that $\mathcal{G}'_1$ has 2 orange nodes, 2 blue nodes and 1 green nodes which yields the corresponding feature vector $\phi_1$). Note that if a particular feature present in the entire set of input graphs is not present in a particular graph, the entry is filled with zero.  \emph{The graphs here are for illustration only; in our task each input graph is only one edit distance different from the base graph $\mathcal{G}_0$}.
    }    
    \label{fig:wlextractor}
  \vspace{-3mm}
\end{figure}

In this section we describe \texttt{WLFeatureExtract} in Algorithm \ref{alg:grabnel} in greater detail. The module takes in both the input graph itself and the set of all input graphs (including itself), as the second argument is to construct a collection of all \gls{WL} features seen in any of the input graphs and controls the dimensionality of the output feature vector so that the entries in feature vectors of different input graphs represent the same \gls{WL} feature. For an illustrated example of the procedure, the readers are referred to Fig. \ref{fig:wlextractor}.

\section{Implementation Details}
\label{app:implementation_details}

\paragraph{Datasets} We provide some key descriptive statistics of the \gls{TU} datasets \cite{morris2020tudataset} in Table \ref{tab:tu_stats}. All \gls{TU} datasets may be downloaded at \url{https://chrsmrrs.github.io/datasets/docs/datasets/}. The \gls{MNIST}-75sp dataset is generated from scripts available at \url{https://github.com/bknyaz/graph_attention_pool}, and the ER-graphs dataset used in App. \ref{sec:rls2v} is available at \url{https://github.com/Hanjun-Dai/graph_adversarial_attack}. The details on the Twitter fake news data are described in the following section.

\paragraph{Twitter dataset} We used the Twitter dataset described in \cite{vosoughi2018spread}. In this dataset, each graph represents a rumour cascade. A cascade is made up of nodes which represents both a tweet and the corresponding user who posted the tweet. An edge exists between nodes $u$ and $v$ if $u$ is a retweet of node $v$. The graphs are directed but for simplicity we drop the direction of edges. We use node features which are described in table \ref{tab:twitter}. We apply a log transform to features with a high level of skewness. All data was normalised by subtracting the mean and divided by the standard deviation (estimated using node features from the training set). Further information of these features are give in the supplementary material of \cite{vosoughi2018spread}. The graph is labelled as true, false or mixed, corresponding to the judged veracity of the rumour, and the learning task is to correctly predict the label of the graph. Many of the graphs in the original dataset are small (and therefore many were topologically similar), we discard samples where the graph has less than $5$ nodes. Furthermore, we use downsampling to balance the dataset so each label appears an equal number of times. After all preprocessing steps are complete the dataset is made up of $4746$ labelled graphs. 

A challenge of node injection is deciding how to choose node features. We reasoned about how to do this for each feature based on the feature semantics. We choose values based on the assumption that the inserted node (tweet) is being inserted from a bot account trying to emulate other users in the cascade. The rumour category is the same for every node in the graph, so inserted nodes used the same as other nodes in the graph. We set the tweet date to the largest of other nodes, to represent an attack in the evasion setting. We set the user account age, followers, followees, retweet status and verified status to the minimum of all other nodes in the graph (using the convention that $\textit{False} \leq \textit{True}$). The user engagement was set to the median among other nodes in the graph. In practice, a user may have control over some of these variables such as the number of followees (by following other accounts) or user engagement (by posting tweets).  

\begin{table}[]
\centering
\caption{Node features used in the Twitter dataset.}
\begin{tabular}{llll}
\toprule
Feature          & Datatype    & Transform applied & Description                                 \\
\midrule
Rumour category  & categorical & One hot encoding  & Topic of the rumour \\
Tweet date       & float       &               & Date and time the tweet was posted                   \\
User account age & float       &               & How old the account is in days \\
User verified    & bool        &               & If the user is verified by Twitter          \\
User followers   & int         & Log transform     & How many followers the user has             \\
User followees   & int         & Log transform     & How many other users the user follows            \\
User engagement  & float       & Log transform     & How active the user has been since joining  \\
Was retweeted    & bool        &               & If the tweet was retweeted    \\
\bottomrule
\label{tab:twitter}
\end{tabular}
\end{table}

\paragraph{Computing Environment} We conduct all experiments, unless otherwise specified, on a shared server with an Intel Xeon CPU and 256GB of RAM.

\begin{table}[t]
\centering
\caption{Key statistics of the \gls{TU} datasets used.}
\begin{tabular}{lllll}
\toprule
Dataset          & \#graphs    & \#labels & Avg \#nodes & Avg \#edges                                 \\
\midrule
\gls{IMDB-M} & $1500$ & $3$ & $13.0$ & $65.9$\\
\gls{PROTEINS} & $1113$ & $2$ & $39.1$ & $72.8$\\
\gls{COLLAB} & $5000$ & $3$ & $74.5$ & $2457.8$ \\
\gls{REDDIT-MULTI-5K} & $4999$ & $5$ & $ 508.8$ & $594.9$ \\
\bottomrule
\label{tab:tu_stats}
\end{tabular}
\end{table}

\paragraph{Victim models}
We focus our attack on two widely used graph neural networks, namely graph convolutional network (\gls{GCN}) \cite{kipf17semi} and graph isomorphism network (\gls{GIN}) \cite{xu2018how}. We also consider an attack on Cheby\gls{GIN} \cite{Knyazev2019} and Graph U-Net \cite{gao2019graph}. The graph convolution layers in these models work by aggregating information across the graph edges and then updating combined node features to output new node features. Multiple layers of graph convolution are used. A readout layer transforms the final node embeddings into a fixed-sized graph embedding which can then be fed through a linear layer and a softmax activation function to provide predicted probabilities for each class. 

The \gls{GCN} graph convolutions take the form 
\begin{equation*}
    \mathbf{X}^{(h)} = \sigma(\tilde{\mathbf{D}}^{-1/2} \tilde{\A} \tilde{\mathbf{D}}^{-1/2} \X^{(h-1)} \mathbf{\Theta}^{(h)})
\end{equation*}
where $\tilde{\A}=\mathbf{A}+\mathbf{I}_n$ is the adjacency matrix with self loops, $\tilde{\mathbf{D}} = \textnormal{diag}(d_1+1, d_2+1,\ldots d_n+1)=\textnormal{diag}(\mathbf{1}\tilde{\A}),$ is a diagonal matrix where $d_u$ is the degree of node $u$. $\mathbf{\Theta}^{(h)}$ and $\X^{(h)}$ are the weight matrix and node features in layer $h$, respectively. For the first layer $\X^{(0)}$ is the original node features. We use three \gls{GCN} convolutions where the dimension of the hidden node representations are 16. A max pooling across feature maps is applied to the final layer to give a fixed length graph representation which is then used as input to a linear layer.  

The graph isomorphism architecture (\gls{GIN}) is provably more expressive (in terms of distinguishing graph topologies) than the \gls{GCN} architecture \cite{xu2018how}. The graph convolution takes the form 
\begin{equation*}
    \mathbf{X}^{(h)} = \textnormal{MLP}^{(h)} \big((1+\epsilon^{(h)}) \X^{(h-1)} + \A \X^{(h-1)} \big),
\end{equation*}
where $\eps^{(h)}$ is a learnable scalar parameter and $\textnormal{MLP}^{(h)}$ is a multilayer perceptron. In our experiments the MLP consists of a single hidden layer of dimension $64$ using ReLU activation functions and batch norm applied before applying the activation to the hidden units. We use 5 convolutional layers, applying batchnorm and ReLU activation functions in-between. For the readout function we utilise the representation after each of the \gls{GIN} convolutions. For each representation, a sum pooling is applied followed by a linear layer. During training dropout is applied to the output of each of the linear layers with probability $p=0.5$. The outputs of the linear layer are summed to give a final logit score for each class.

The Cheby\gls{GIN} architecture is similar to the \gls{GIN} architecture but aggregates information from nodes across multi-hop neighbourhoods. This is achieved by using higher-order Chebyshev polynomials as the aggregation matrix. The polynomial filter is evaluated using the (shifted) normalised Laplacian matrix $\mathbf{L}=-\mathbf{D}^{-1/2}\A \mathbf{D}^{-1/2}$. Chebyshev polynomials can be defined recursively:
\begin{equation}
    T_k(\mathbf{L}) = 
    \begin{cases}
    \mathbf{I}_n , & \text{for } k = 0 \\ 
    \mathbf{L} , & \text{for } k = 1 \\
    2\mathbf{L}T_{k-1}(\mathbf{L}) - T_{k-2}(\mathbf{L}), & \text{for } k >2 
    \end{cases}.
\end{equation}
The Cheby\gls{GIN} convolution is then defined to be
\begin{equation*}
    \mathbf{X}^{(h)} = \textnormal{MLP} \big((1+\epsilon) \X^{(h-1)} + T_k(\mathbf{L}) \X^{(h-1)} \big).
\end{equation*}
We used a pretrained model used in \cite{Knyazev2019} available at \url{https://github.com/bknyaz/graph_attention_pool}. We use the model with supervised attention, the best-performing pre-trained model available. 

Graph U-Nets are an autoencoder like architecture with skip connections \cite{{gao2019graph}}. The architecture consists of a differential graph pooling layer \textit{gPool} and a differential graph unpooling layer \textit{gUnpool} which we briefly describe. To do differential graph pooling a $d \times p$ projection matrix $\mathbf{p}$ is used to compute a length $n$ vector $\mathbf{y}=\mathbf{X} \mathbf{p}$. A ranking operation is used to select the indices of the largest $k$ entries $\textit{idx} = \operatorname{rank}(\mathbf{y}, k)$ which represent nodes to be included in the sub-graph after pooling. Using this we can select the subgraph adjacency $\mathbf{A}^{(l+1)} = \mathbf{A}^{(l)}[\textit{idx}, \textit{idx}]$. The corresponding rows of the features matrix are selected $\tilde{\mathbf{X}}^{(l)} = \mathbf{X}^{(l)}[\textit{idx}, :]$ and then a re-normalisation is applied to give features for the next layer $\mathbf{X}^{(l+1)} = \tilde{\mathbf{X}}^{(l)} \odot (\operatorname{sigmoid}(\mathbf{y}) \mathbf{1}_d^T)$. The \textit{gUnpool} operator does a reverse operation by using the indices computed in the pooling layer and using zero vectors for indices that were not selected during pooling. The architecture uses rounds of pooling and unpooling, as well as skip connections between representations of the same size. For a detailed description of the architecture we refer the reader to \cite[Section 3]{gao2019graph}. We used the open source implementation used in \cite{gao2019graph}p and provided by the authors at \url{https://github.com/HongyangGao/Graph-U-Nets}.  

\paragraph{\gls{GRABNEL}} \gls{GRABNEL}, which uses the \gls{WL} feature extractor, involves a number of hyperparameters: the \gls{WL} procedure is parameterised by a single hyperparameter $H$, which specifies the number of Weisfiler-Lehman iterations to perform. While it is possible for $H$ to be selected automatically via, for example, maximising the log-marginal likelihood of the surrogate model (e.g. \cite{ru2020neural}), in our case we find fixing $H = 1$ to be performing well. For the sparse Bayesian linear regression model used, we always normalise the input data into hypercubes $[0, 1]^d$ and standardise the target by deducting its mean and dividing by standard deviation. We optimise the marginal log-likelihood via a simple gradient optimiser and we set the maximum number of iterations to be $300$. As described in Sec. \ref{sec:method}, we need to specify a Gamma prior over over $\{\lambda_i\}$ and we use shape parameter and inverse scale parameters of $1 \times 10^{-6}$. For the acquisition optimisation, we set the maximum number of evaluation of the acquisition function to be $500$: we initialise with $50$ randomly sampled perturbed graphs, each of which is generated from flipping one pair of randomly selected end nodes from the base graph. To generate the initial population, we fill generate $50$ candidates by mutating from the top-3 queried graphs that previously led to the largest attack loss (if we have not yet queried any graphs, we simply sample 100 randomly perturbed graphs). We then evolve the population $10$ times, with each evolution cycle involving mutating the current population to generate offspring and popping the oldest members in the population. Finally, we select the top $5$ unique candidates seen during the evolution process that have the highest acquisition function value (we use the \gls{EI} acquisition function) to query the victim model.

\section{Additional Experiments}
\label{app:more_experiments}

\subsection{Comparison with Alternative Surrogate Models}
\label{app:surrogate_comparisons}

\begin{figure}[t]
    \centering
    \begin{subfigure}{0.32\linewidth}
    \includegraphics[trim=0cm 0cm 0cm  0cm, clip, width=1.0\linewidth]{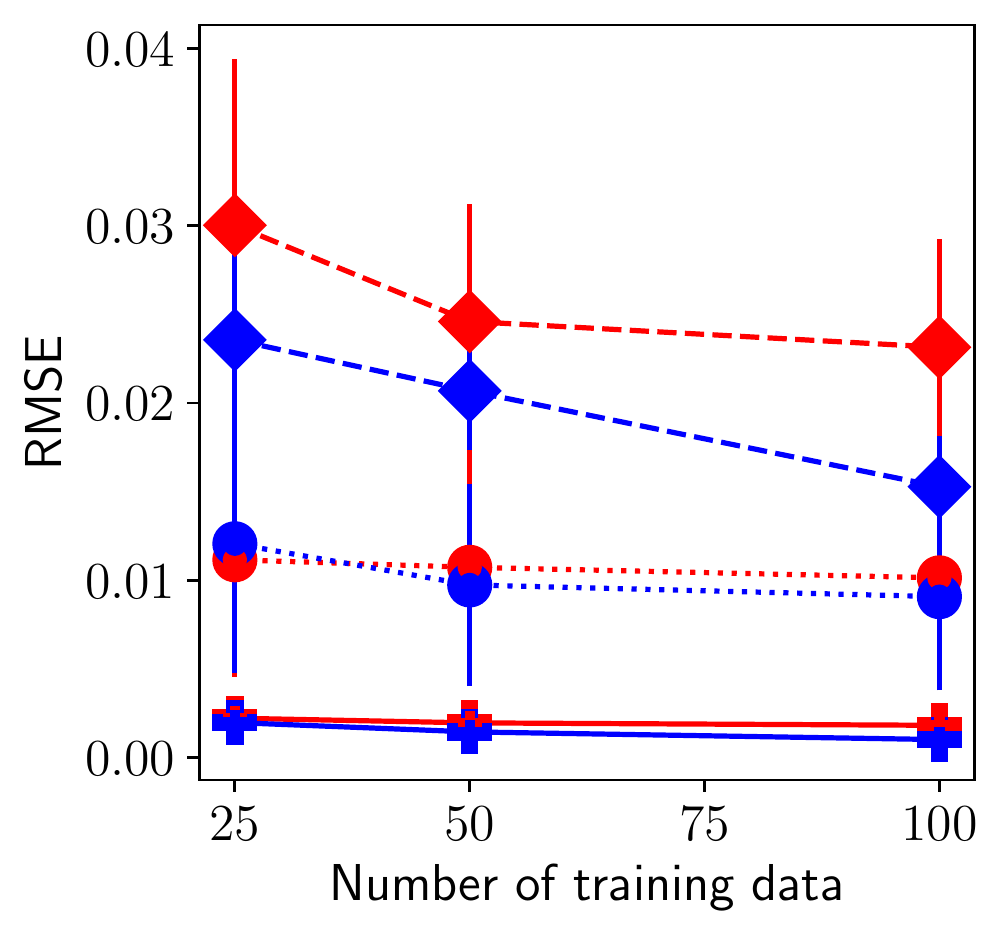}
    \end{subfigure}
         \begin{subfigure}{0.32\linewidth}
    \includegraphics[trim=0cm 0cm 0cm  0cm, clip, width=1.0\linewidth]{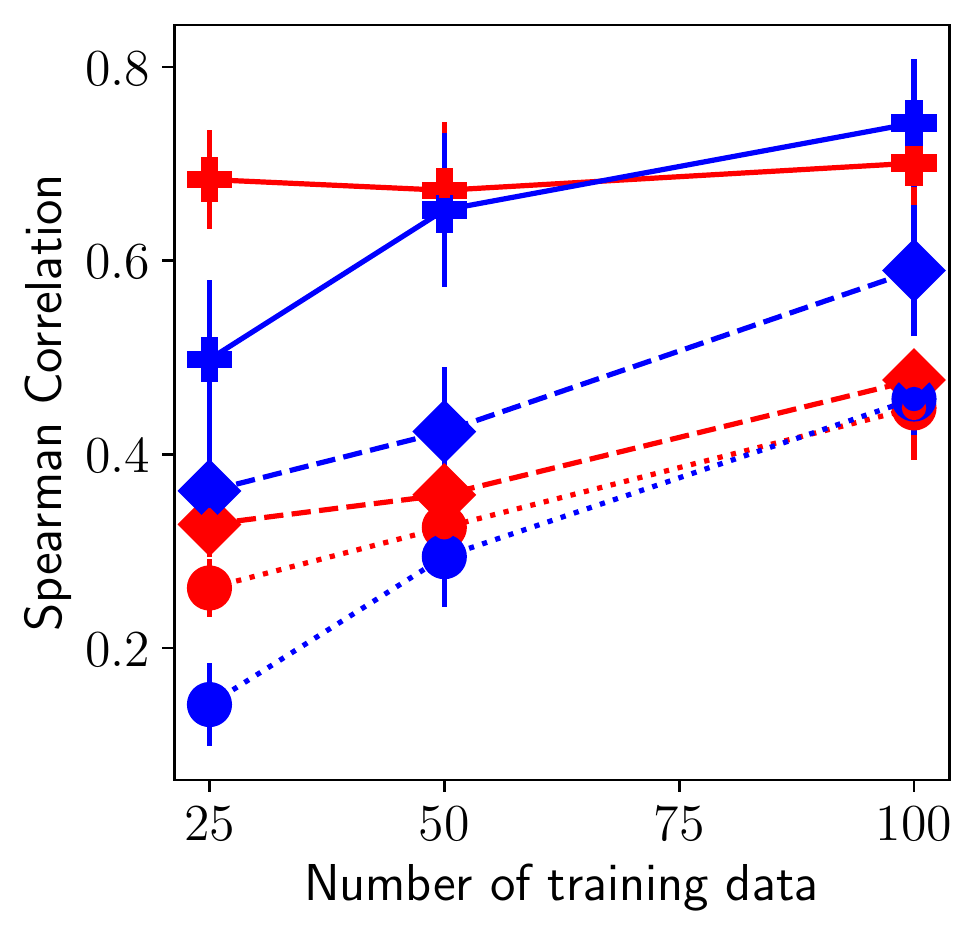}
    \end{subfigure}
    \begin{subfigure}{0.32\linewidth}
    \includegraphics[trim=0cm 0cm 0cm  0cm, clip, width=1.0\linewidth]{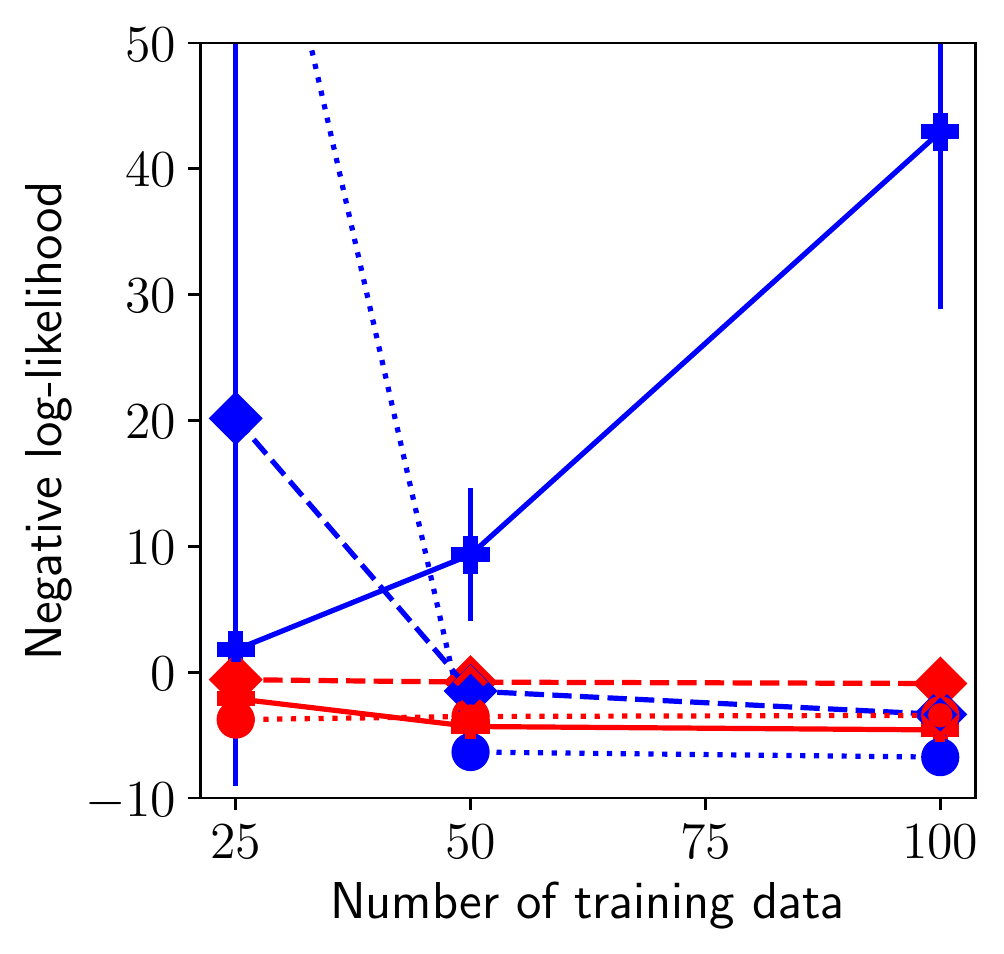}
    \end{subfigure}
    
    \begin{subfigure}{0.4\linewidth}
    \includegraphics[trim=0cm 0cm 0cm  0cm, clip, width=1.0\linewidth]{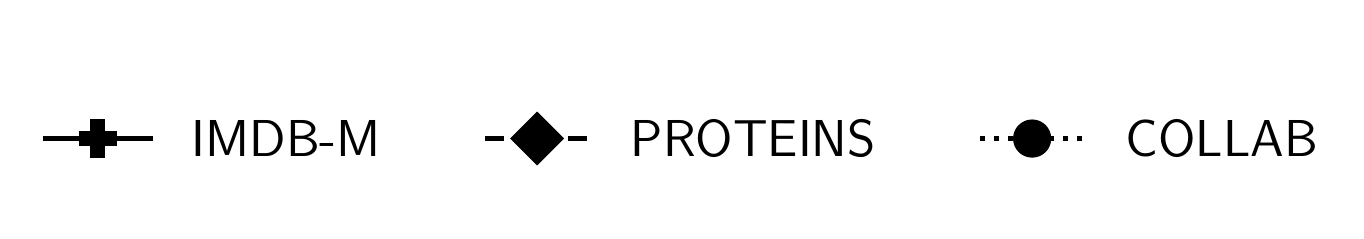}
    \end{subfigure}
    \caption{Comparison of the \textcolor{red}{Bayesian linear regression} and \textcolor{blue}{\gls{GP}} surrogate models on 3 \gls{TU} datasets in terms of \gls{RMSE} (lower is better), Spearman correlation (higher is better) and the negative log-likelihood (lower is better) on validation set. Error bars denote 1 standard error.
    }    
    \label{fig:surrogates}
  \vspace{-3mm}
\end{figure}

We compare the surrogate performance between a \gls{GP} model with RBF kernel and a Bayesian linear regression model (which is equivalent to \gls{GP} with linear kernel) to justify the usage of the latter in this section. To make a fair comparison, in each of the 3 \gls{TU} datasets, we randomly select 20 graph samples that the victim model (we use the trained \gls{GCN} models identical to those used in Section \ref{sec:experiments}) originally classify correctly. For each of the sample, we generate $\{25, 50, 100\}$ perturbed samples that are 1 edit distance different from the original graphs and query the victim model to obtain their respective attack losses. We then train the surrogate models with the perturbed graphs and their attack losses as the training inputs and targets, and validate their performance on a further validation set of $200$ perturbed graphs with the objective of predicting their attack losses. We use 3 metrics to evaluate the quality of the surrogate model: \gls{RMSE} between predicted attack losses and the ground-truth atack losses, Spearman correlation between them and the negative log-likelihood on the validation set (which assess the quality of the prediction mean as long as its predictive uncertainty, as a principled uncertainty estimation is crucial in Bayesian optimisation). The results are shown in Fig. \ref{fig:surrogates}: it can be seen that the difference between Bayesian linear regression and \gls{GP} models are insignificant in most cases in terms of \gls{RMSE} and Spearman correlation (except in the \gls{PROTEINS} dataset where \gls{GP} model is arguably better), whereas the uncertainty estimation seems to be more stable throughout for Bayesian linear regression. It therefore justifies our usage of Bayesian linear regression, as its performance is often comparable with \gls{GP}, but it is much more cheaper in terms of computation, making computations much more tractable especially when the number of queries is modestly large.

\subsection{Comparison with RL-S2V}
\label{sec:rls2v}
We compare \gls{GRABNEL} with \gls{RL-S2V} on the graph classification dataset described in \cite{dai2018adversarial}. Each input graph is made of $1$, $2$ or $3$ connected components. Each connected component is generated using the Erdős–Rényi random graph model (additional edges are added if the generated graph is disconnected). The label node features are set to a scalar value of 1 and the corresponding graph label is the number of connected components. The authors consider three variants of this dataset using different graph sizes, we consider the variant with the smallest graphs ($15-20$ nodes. The victim model, as well as the surrogate model used to compute Q-values in \gls{RL-S2V} is structure2vec \cite{dai2016discriminative}. This embedding has a hyper-parameter determining the depth of the computational graph. We fix both to be the the smallest model considered in \cite{dai2018adversarial}. These choices were made to keep the computational budget to a minimum. 

\begin{figure}[ht]
\centering

\begin{minipage}{0.35\textwidth}
    \includegraphics[trim=0cm 0cm 0cm  0cm, clip, width=1.0\linewidth]{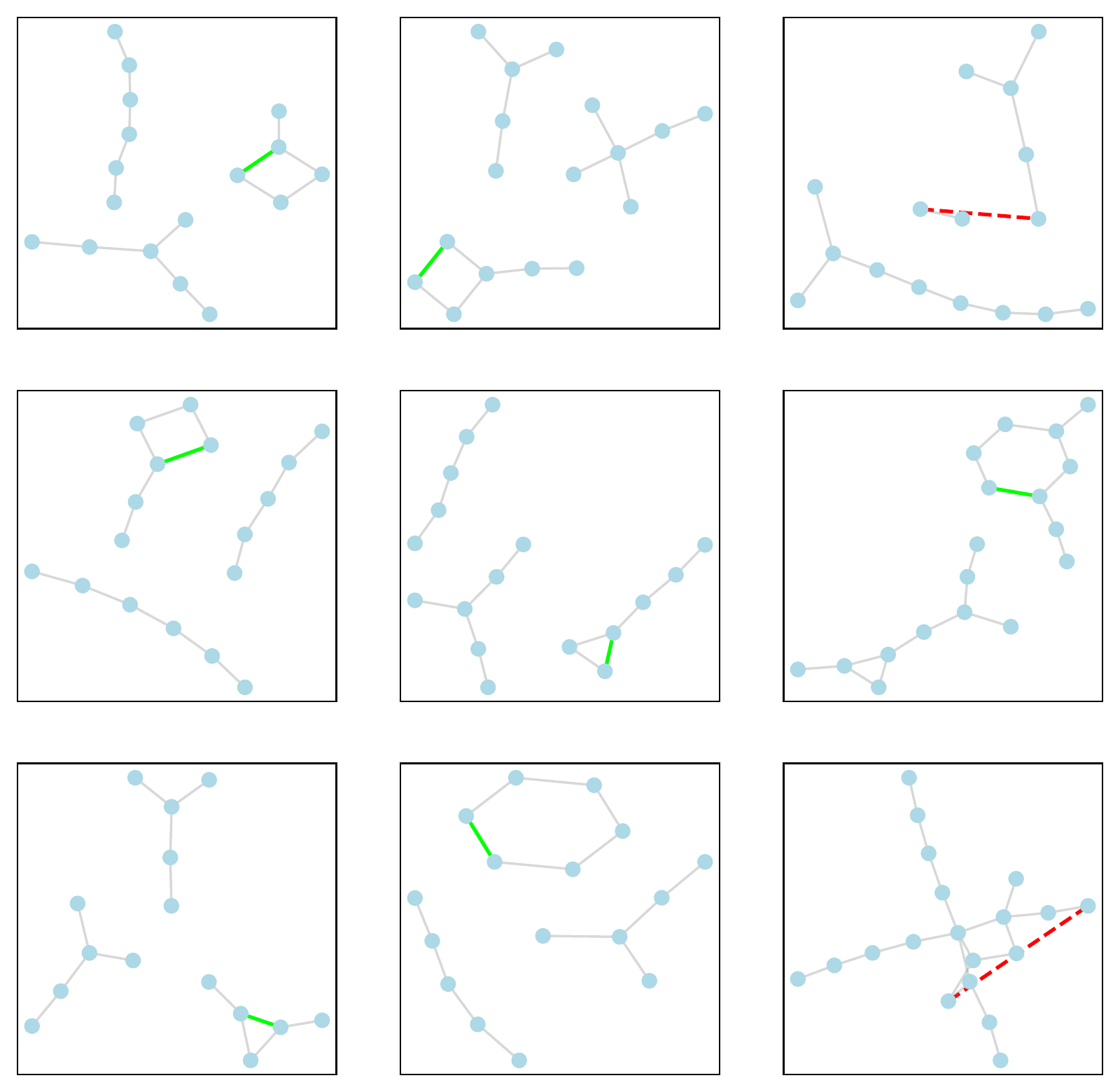}
\caption{Adversarial examples found by the proposed method on the ER graphs with S2V being the victim model. Similar to Fig. \ref{fig:advexamples}, \textcolor{red}{Red edges} denote deleted edges from the original samples and \textcolor{green}{green edges} indicate those added.}
\label{fig:advexampless2v}
\end{minipage}
\qquad
\begin{minipage}{0.4\textwidth}
\includegraphics[width=.9\textwidth]{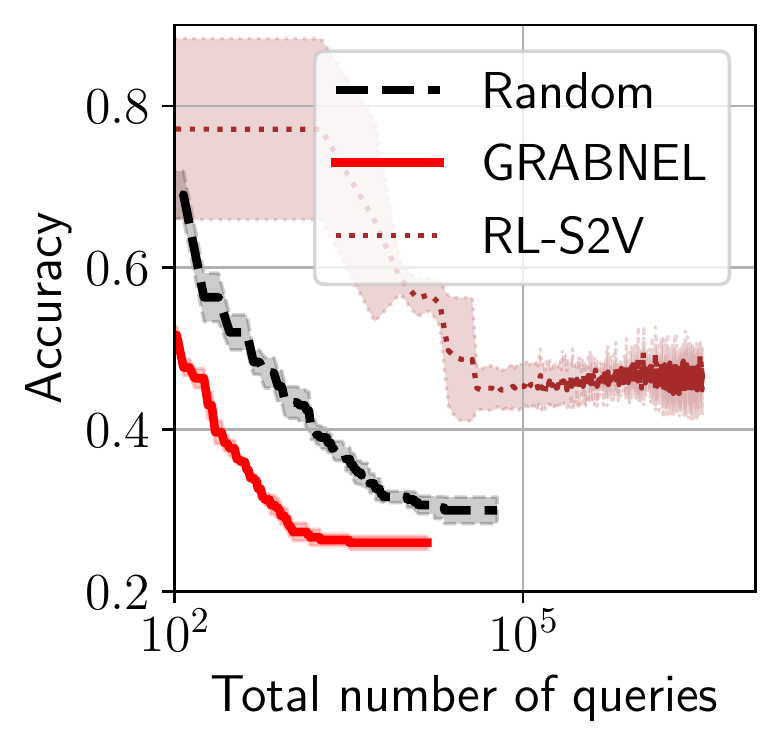}
\caption{Validation accuracy vs \emph{total} number of queries to the victim model. \gls{RL-S2V} requires significantly more victim model queries, as it attempts to learn an attack policy by repeatedly querying a subset of the validation set which is used for policy training.}  
\label{fig:acc_vs_queries_s2v}
\end{minipage}
\end{figure}

To adapt to the settings in \cite{dai2018adversarial}, we only allow one edge edit (addition/deletion), and for \gls{GRABNEL} we allow up to 100 queries to the victim model per sample in the validation set. For Random baseline, we instead allow up to 400 queries. Similar to \cite{dai2018adversarial}, we enforce the constraint such that any edge edit must not result in a change of the number of disconnected components (i.e. the label) and any such edit proposed is rejected before querying the victim model. We show the results in Fig. \ref{fig:acc_vs_queries_s2v}, and we similarly visualise some of the adversarial samples found by \gls{GRABNEL} in Fig. \ref{fig:advexampless2v}. The final performance of \gls{RL-S2V} is similar to that reported in the \cite{dai2018adversarial}, whereas we find that random perturbation is actually a very strong baseline if we give it sufficient query budget\footnote{The random baseline reported in \cite{dai2018adversarial} is obtained by only querying victim model with a randomly perturbed graph \emph{once}.}. Again, we find that \gls{GRABNEL} outperforms the baselines, offering orders-of-magnitude speedup compared to \gls{RL-S2V}, with the main reasons being 1) \gls{GRABNEL} is designed to be sample-efficient, and 2) \gls{GRABNEL} does not require a separate training set \emph{within} the validation to train a policy like what \gls{RL-S2V} does. Fig. \ref{fig:advexampless2v} shows that the edge addition is more common than deletion in the adversarial examples in this particular case, and often the attack agent forms \emph{ring structures}. Such structures are rather uncommon in the original graphs generated from the Erdos-Renyi generator, and are thus might not be familiar to the classifier during training. This might explain why the victim model seems particularly vulnerable to such attacks.

\subsection{Additional Examples of Adversarial Samples Discovered}
\label{app:more_stats_adv}

\begin{figure}[H]
    \centering
    \begin{subfigure}{0.45\linewidth}
    \includegraphics[trim=0cm 0cm 0cm  0cm, clip, width=1.0\linewidth]{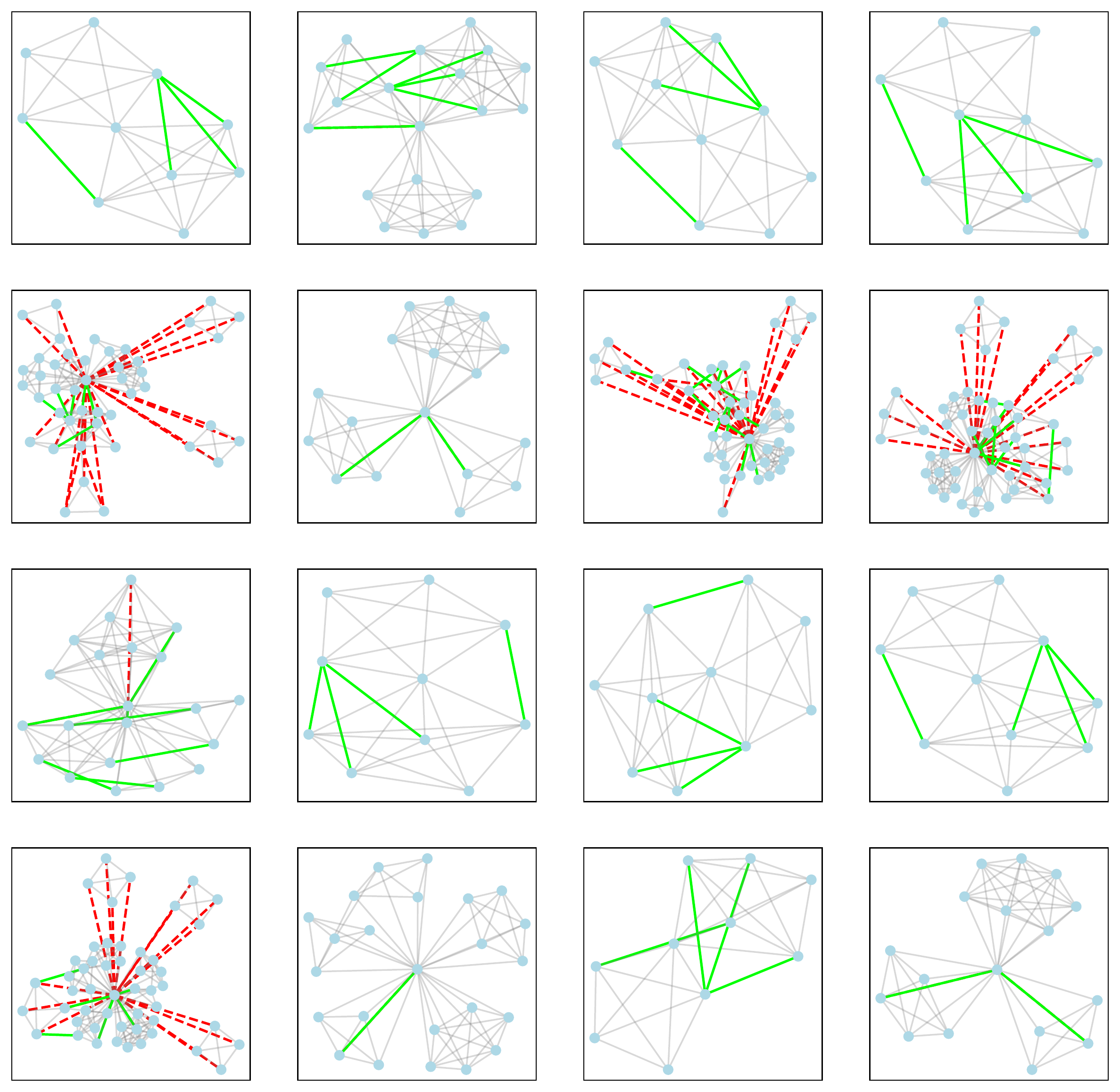}
        \caption{\gls{IMDB-M} (\gls{GCN})}
    \end{subfigure}
    \quad
         \begin{subfigure}{0.45\linewidth}
    \includegraphics[trim=0cm 0cm 0cm  0cm, clip, width=1.0\linewidth]{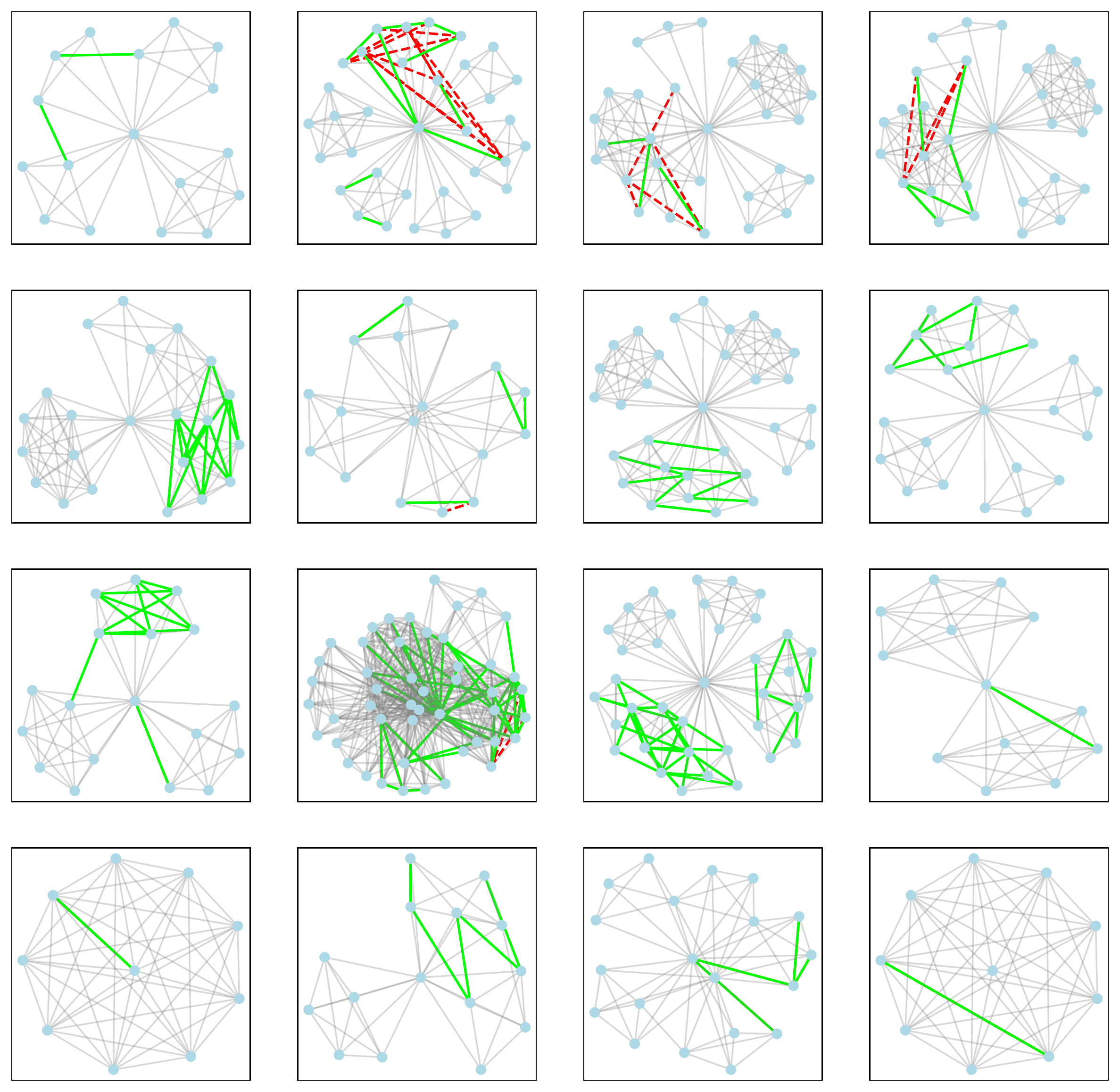}
        \caption{\gls{IMDB-M} (\gls{GIN})}
    \end{subfigure}
    \begin{subfigure}{0.45\linewidth}
    \includegraphics[trim=0cm 0cm 0cm  0cm, clip, width=1.0\linewidth]{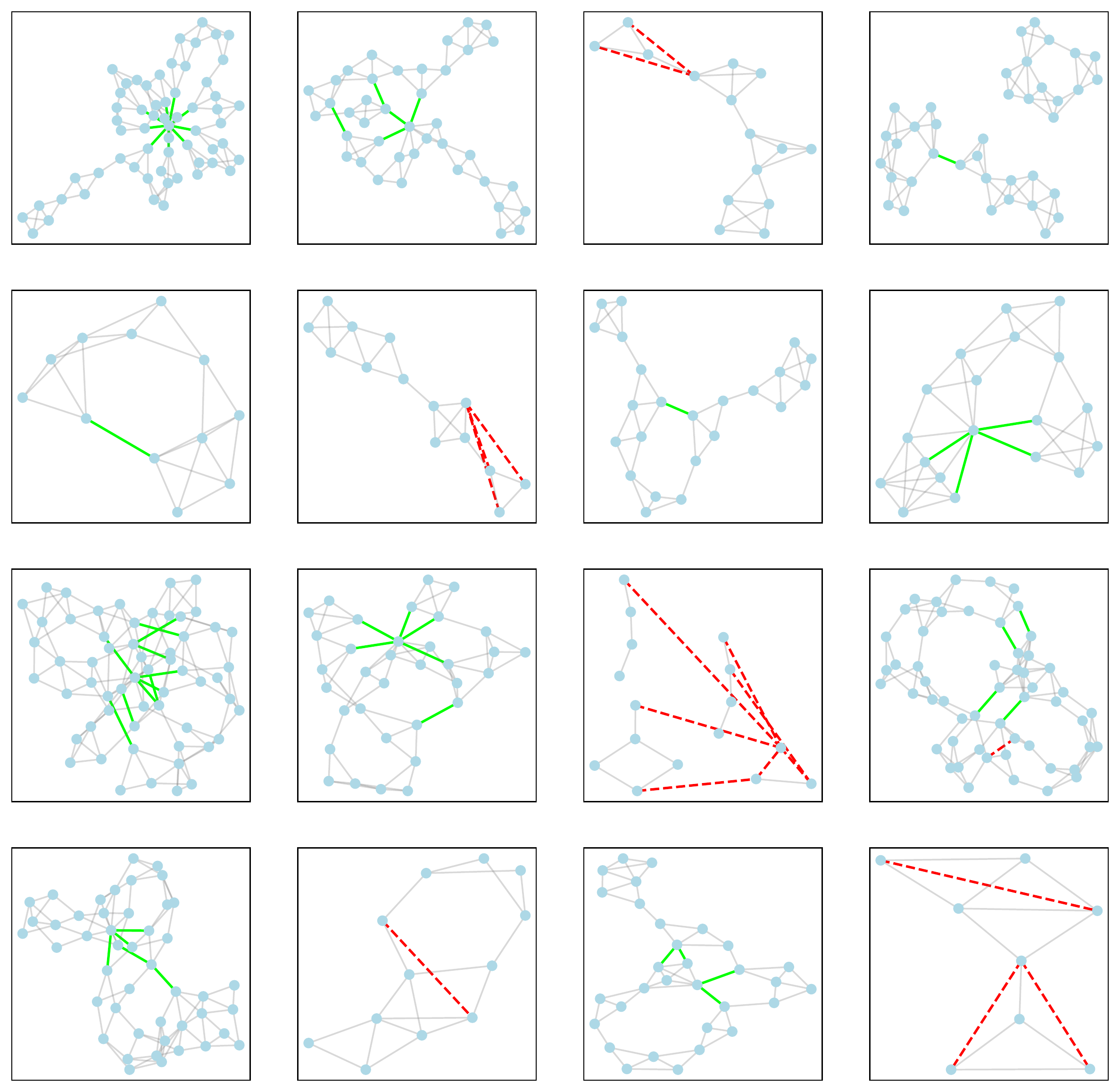}
    \caption{\gls{PROTEINS} (\gls{GCN})}
    \end{subfigure}
        \quad
    \centering
       \begin{subfigure}{0.45\linewidth}
    \includegraphics[trim=0cm 0cm 0cm  0cm, clip, width=1.0\linewidth]{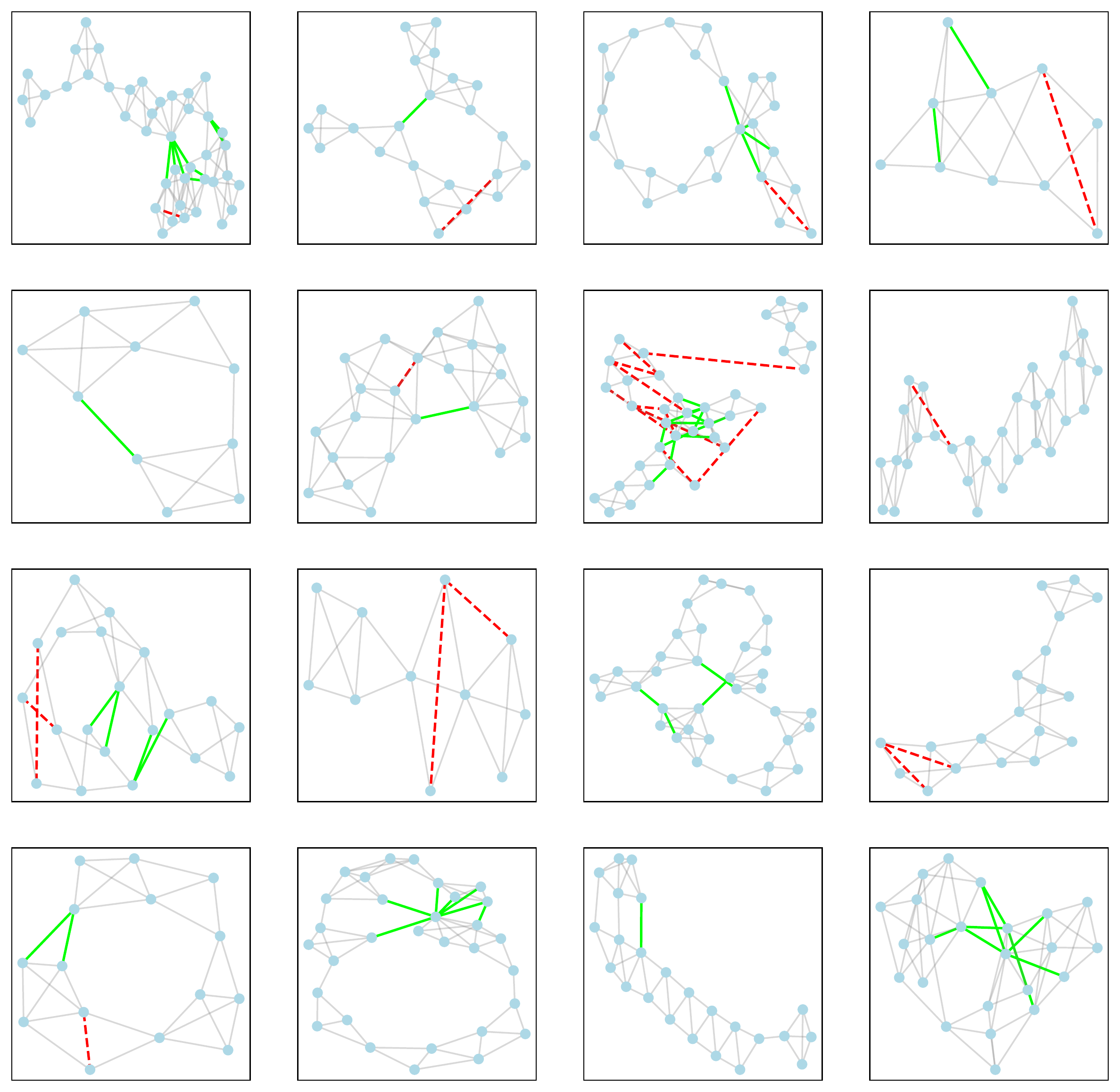}
    \caption{\gls{PROTEINS} (\gls{GIN})}
    \end{subfigure}
       \begin{subfigure}{0.45\linewidth}
    \includegraphics[trim=0cm 0cm 0cm  0cm, clip, width=1.0\linewidth]{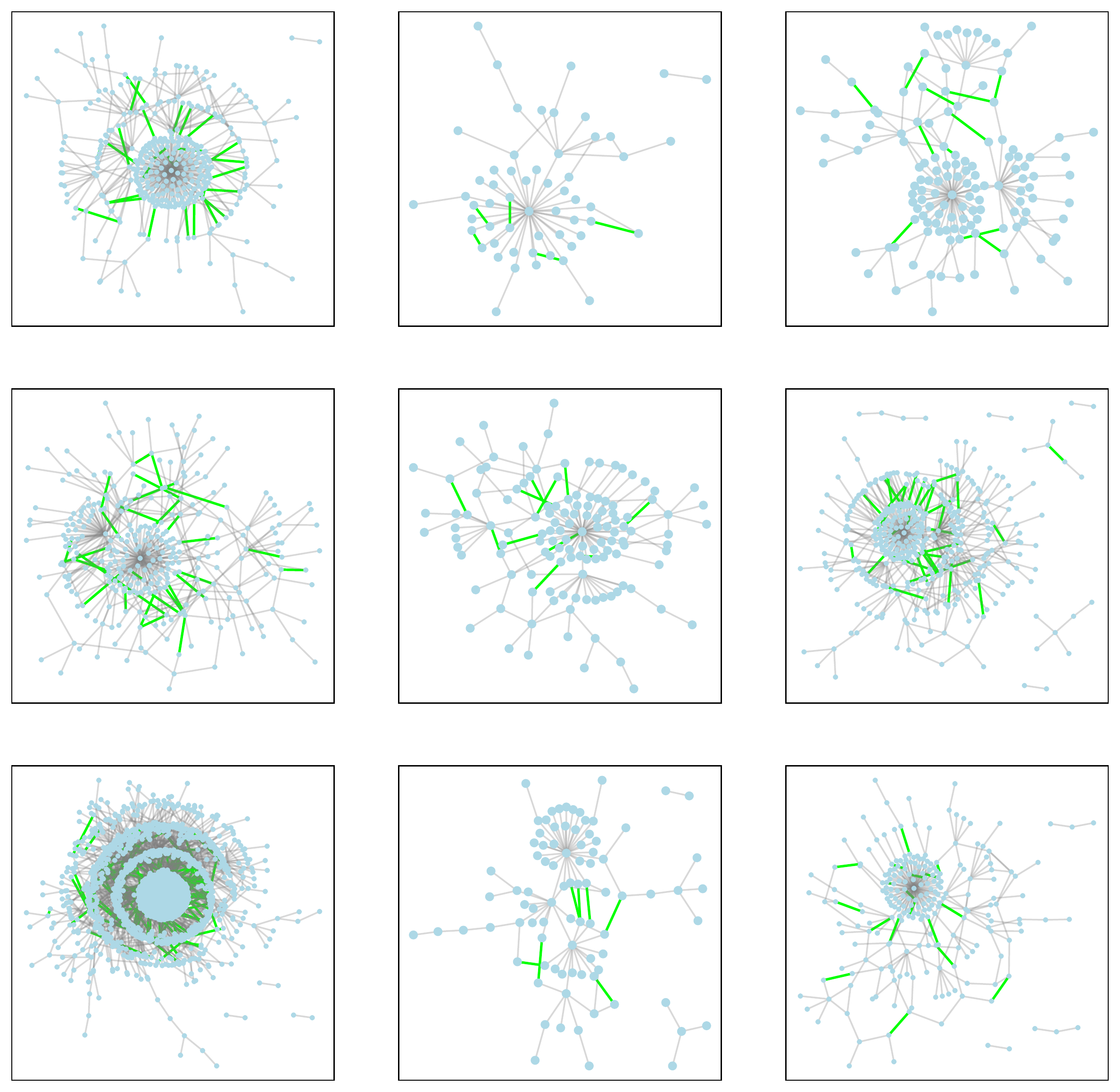}
        \caption{\gls{REDDIT-MULTI-5K} (\gls{GCN})}
    \end{subfigure}
        \quad
         \begin{subfigure}{0.45\linewidth}
    \includegraphics[trim=0cm 0cm 0cm  0cm, clip, width=1.0\linewidth]{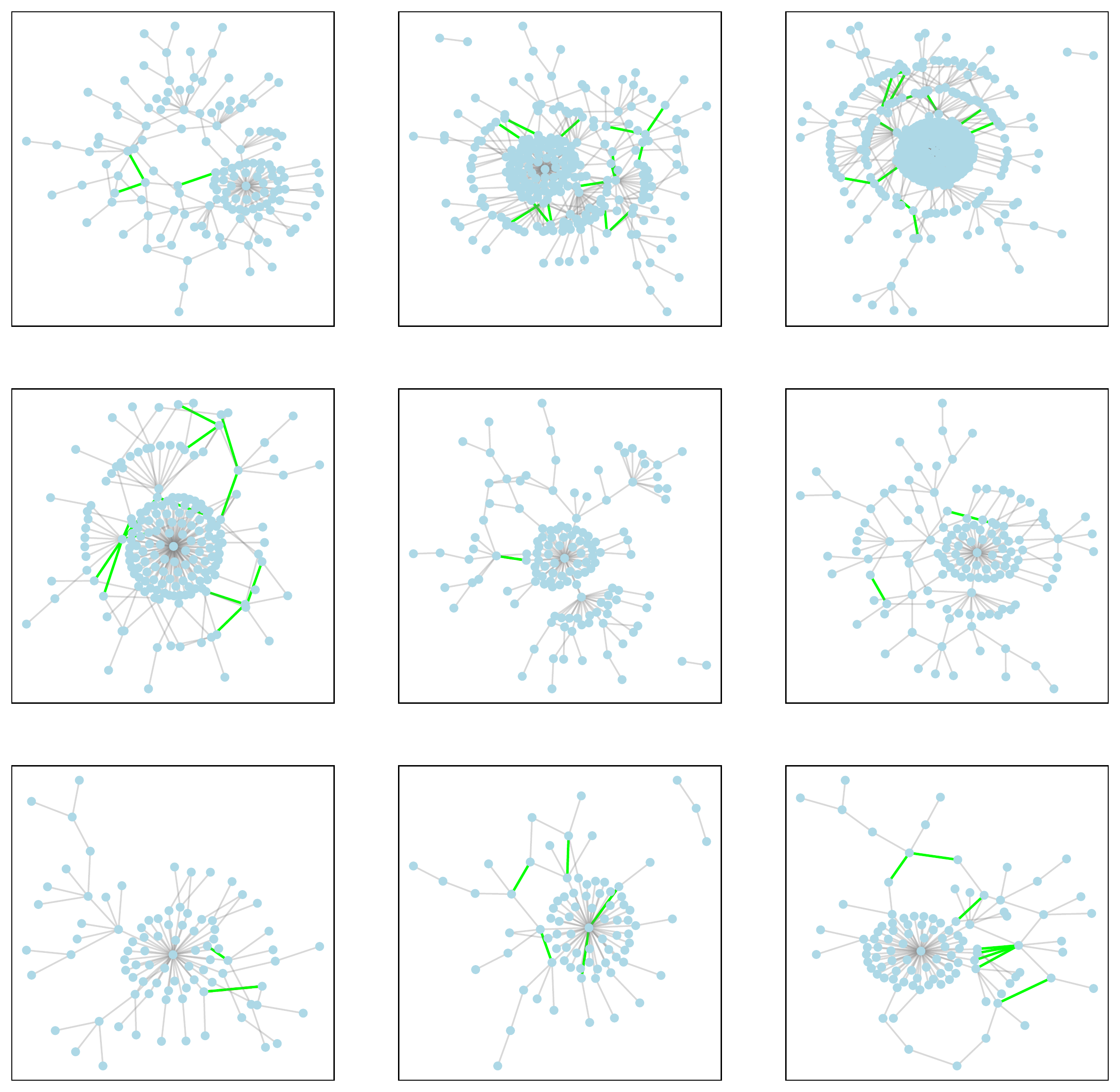}
    \caption{\gls{REDDIT-MULTI-5K} (\gls{GIN})}
    \end{subfigure}
    \caption{More examples of adversarial examples found by \gls{GRABNEL} using \gls{GCN}/\gls{GIN} victim models. The colors have the same meaning as Fig. \ref{fig:advexamples} in the main text.
    }    
    \label{fig:more_advexamples}
\end{figure}

\begin{figure}[H]
    \centering
    \begin{subfigure}{0.45\linewidth}
    \includegraphics[trim=0cm 0cm 0cm  0cm, clip, width=1.0\linewidth]{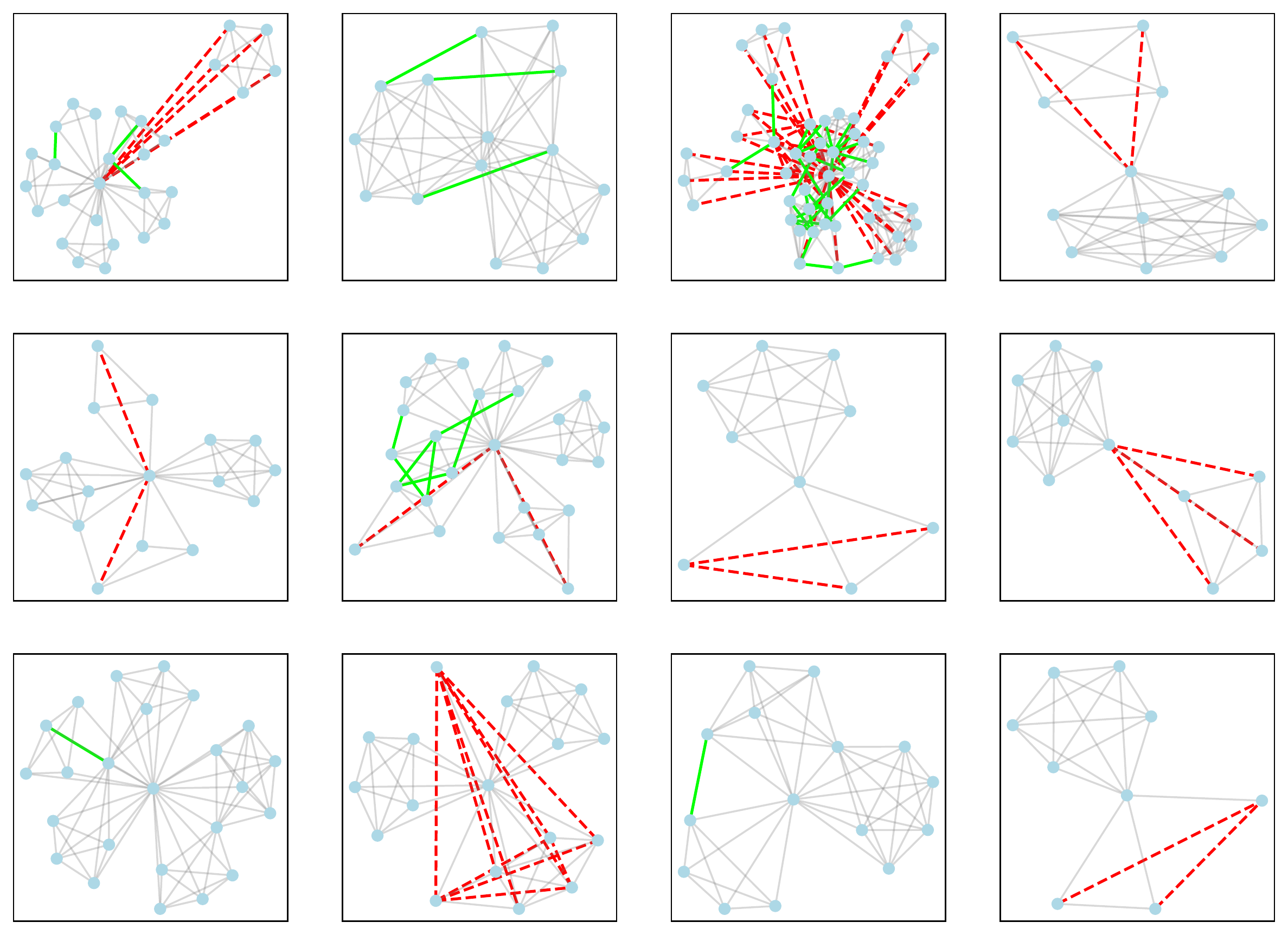}
        \caption{\gls{IMDB-M}}
    \end{subfigure}
    \quad
         \begin{subfigure}{0.45\linewidth}
    \includegraphics[trim=0cm 0cm 0cm  0cm, clip, width=1.0\linewidth]{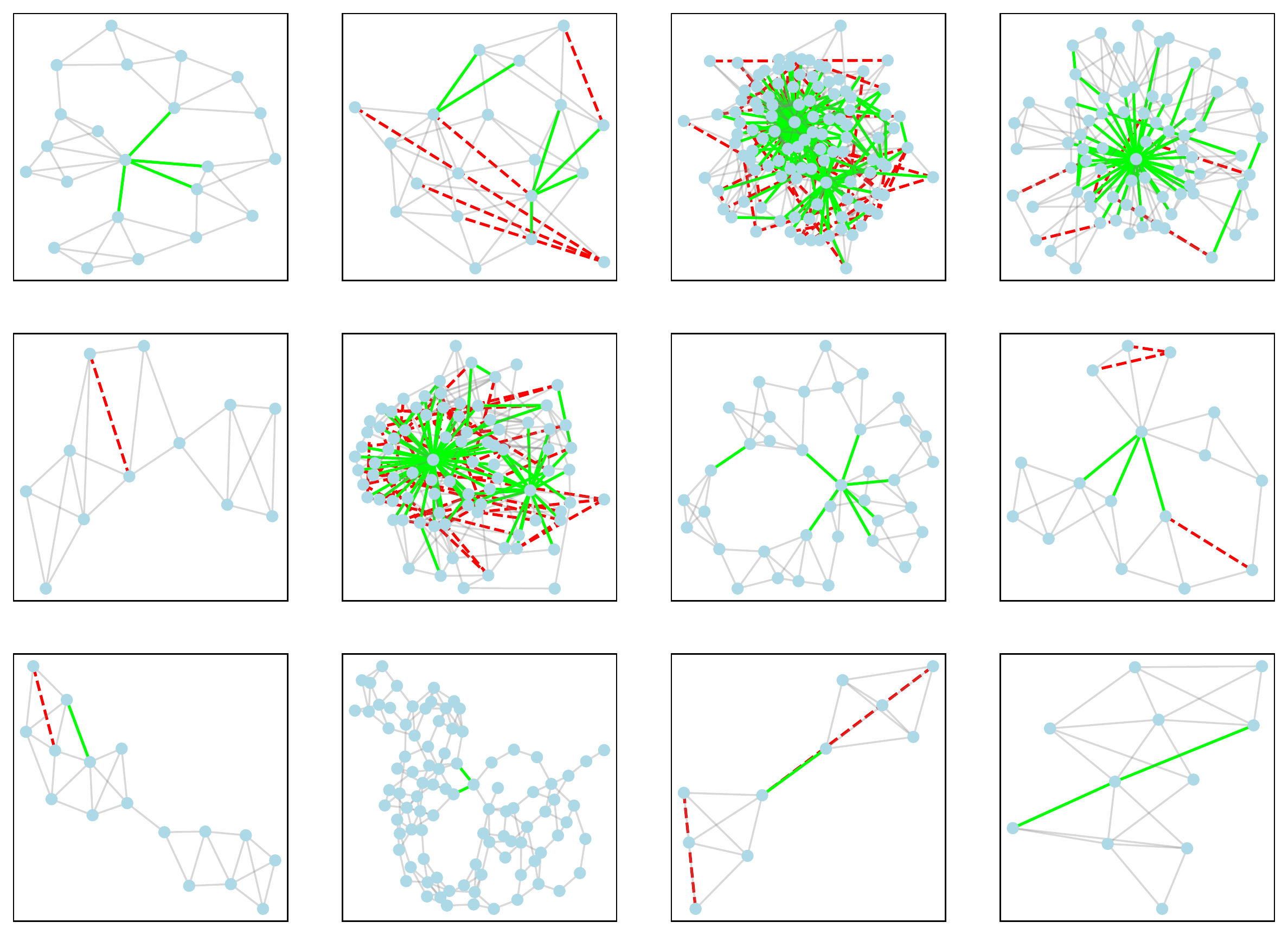}
        \caption{\gls{PROTEINS}}
    \end{subfigure}
   
    \caption{Examples of adversarial examples found by \gls{GRABNEL} using Graph U-net victim models. The colors have the same meaning as Fig. \ref{fig:advexamples} in the main text.
    }    
    \label{fig:more_advexamples_gunet}
\end{figure}

We show more examples of the adversarial examples found by \gls{GRABNEL} on various datasets and victim models in Figs \ref{fig:more_advexamples} and \ref{fig:more_advexamples_gunet}.

\subsection{REDDIT-MULTI-5K Results}
\label{app:reddit_multi_5k}

For the \gls{REDDIT-MULTI-5K} experiments, the graphs are in general typically much sparser and denoting the perturbation budget $\Delta$ with structural perturbation budget in terms of $n^2$ may be too lenient. Instead, we limit the perturbation budget in terms of the fraction of the \emph{number of edges} of the individual graphs, and we set $\Delta \leq 0.03m$ for all experiments. We report the results in Table \ref{tab:tu_r5k} and some examples of the adversarial examples discovered can be found in Fig. \ref{fig:more_advexamples}.

\begin{table}[h]
\centering
\caption{Test accuracy of \gls{GCN} and \gls{GIN} victim models \gls{REDDIT-MULTI-5K} before (\emph{clean}) and after various attack methods.}
\resizebox{0.4\linewidth}{!}{
\begin{tabular}{@{}lcc@{}}
\toprule
& \gls{GCN} \cite{kipf17semi} & \gls{GIN}\cite{xu2018how} \\
 \midrule
\emph{Clean} & $45.20$ & $48.40$ \\
Random & $32.77$ & $42.77$ \\
Genetic \cite{dai2018adversarial} & $28.25$ & $42.35$ \\
\midrule
\gls{GRABNEL} (ours) & $\mathbf{23.73}$ & $\mathbf{29.27}$\\
\bottomrule
\label{tab:tu_r5k}
\end{tabular}
}
\end{table}

\subsection{Ablation Studies}
\label{app:ablation_studies}
In this section, we conduct ablation studies on \gls{GRABNEL} on two datasets previously considered in Section \ref{sec:experiments} in the main text, namely the \gls{PROTEINS} dataset and the \gls{MNIST}-75sp image classification task. We conduct the following variants of \gls{GRABNEL} to understand how different components affect the performance:
\begin{itemize} [leftmargin=0.1in, itemsep=0.05pt, topsep=0.05pt]
	\item \texttt{Random} and \texttt{GA}: Identical to those in Sec. \ref{sec:experiments}.
	
	\item \texttt{SequentialRandom}: instead of perturbing all edges simultaneously we use the sequential perturbation generation: we divide the total query budget into stages according to the description in Sec. \ref{sec:method}, and at each stage we only modify one edge from the base graph \emph{via random sampling} and commit to the perturbation that leads to the largest attack loss in the previous stage. This setup is otherwise identical to \gls{GRABNEL} but the candidates are generated via random sampling instead of surrogate-guided \gls{BO}.
	
	\item \texttt{GrabnelNoSequential}:  \gls{GRABNEL} but without the sequential perturbation selection. At each \gls{BO} iteration, the \gls{BO} needs to search and propose \emph{all} (instead of one) edges to perturb up to the attack budget
	
	\item \texttt{Grabnel}: Full \gls{GRABNEL} with both surrogate-guided \gls{BO} and sequential perturbation selection.
	
\end{itemize}

\begin{figure}[t]
    \centering
    \begin{subfigure}{0.31\linewidth}
    \includegraphics[trim=0cm 0cm 0cm  0cm, clip, width=1.0\linewidth]{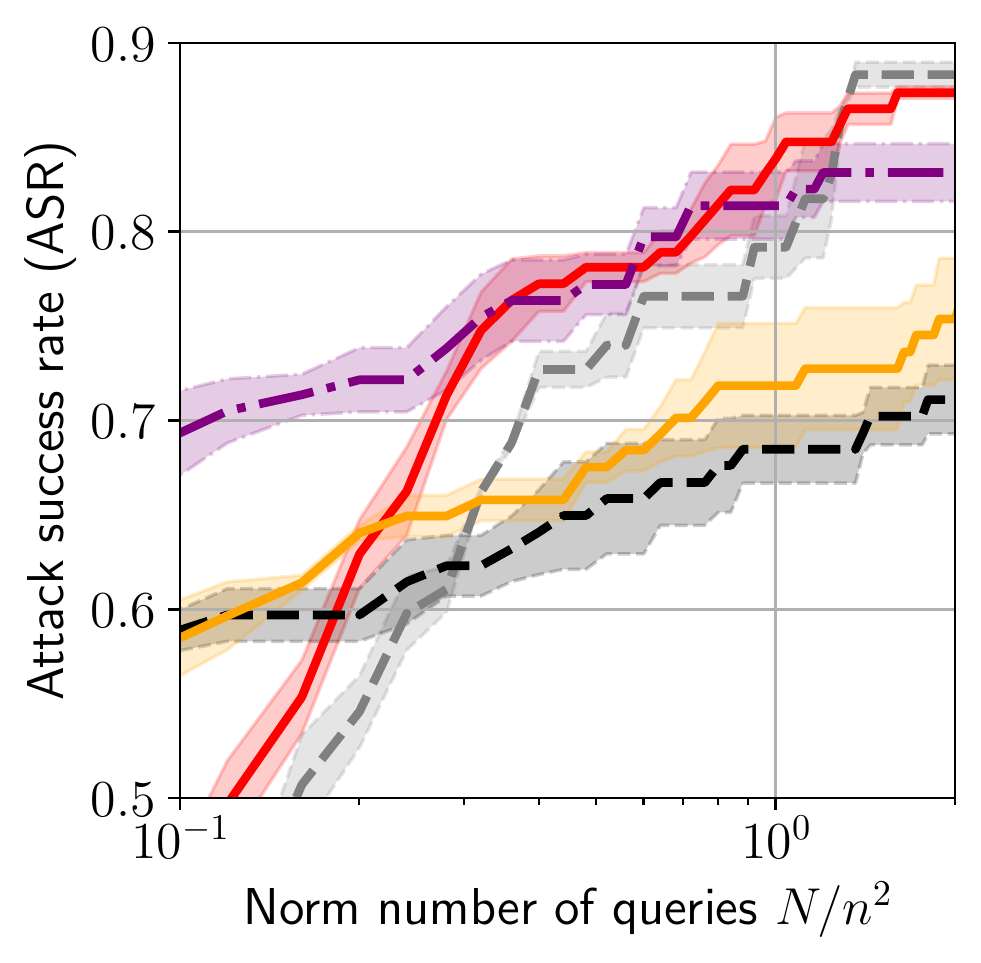}
    \end{subfigure}
         \begin{subfigure}{0.32\linewidth}
    \includegraphics[trim=0cm 0cm 0cm  0cm, clip, width=1.0\linewidth]{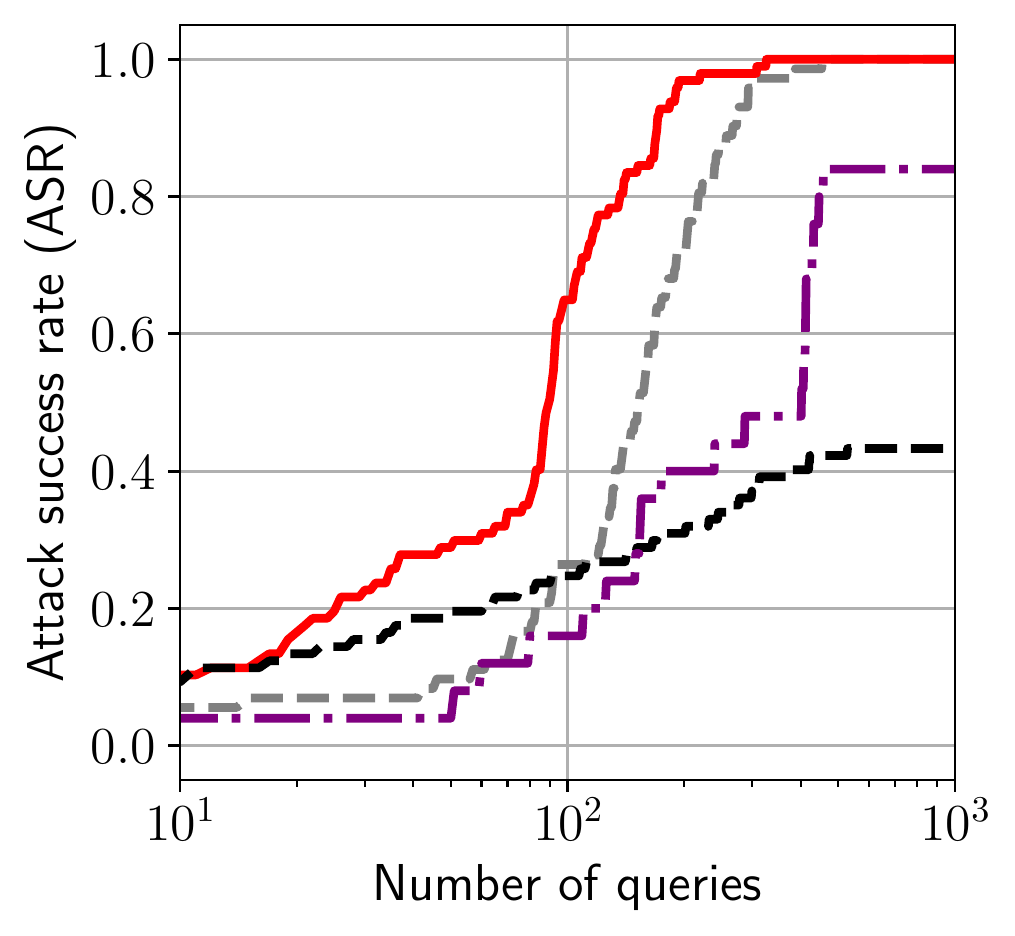}
    \end{subfigure}
    \qquad
    \vspace{-3mm}
    \begin{subfigure}{0.8\linewidth}
    \includegraphics[trim=0cm 0cm 0cm  0cm, clip, width=1.0\linewidth]{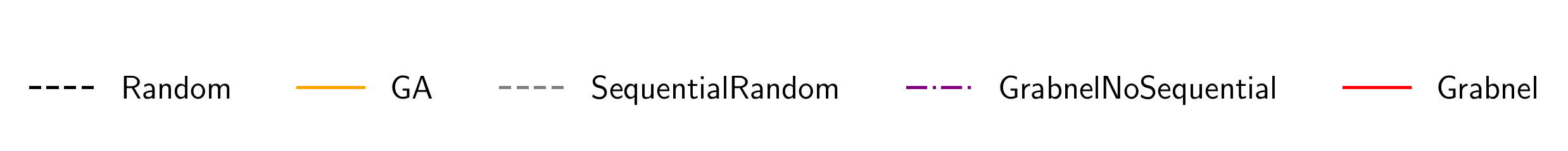}
    \end{subfigure}
    \caption{Ablation studies on \gls{PROTEINS} and \gls{MNIST}-75sp datasets.
    }    
    \label{fig:ablation}
  \vspace{-3mm}
\end{figure}

We show the results in Fig. \ref{fig:ablation}: it is evident that in both cases the use of surrogates and the use of sequential perturbation selection has led to improvements over baselines. In particular, \texttt{SequentialRandom} seems also to be a simple and strong baseline, with their final performance on par with \gls{GRABNEL} but is less sample efficient. In \gls{PROTEINS}, \gls{GRABNEL} converges much faster initially (noting the log scale of the x-axis), while the final performance is comparable between \gls{GRABNEL} and \texttt{SequentialRandom}. In the \gls{MNIST}-75sp task, \gls{GRABNEL} is roughly 1-2$\times$ faster throughout (it is worth noting that since we explicitly link the number of queries to the amount of perturbation applied, \gls{GRABNEL} being 1-2$\times$ faster also suggests that it requires 1-2$\times$ less perturbation compared to \texttt{SequantialRandom} to reach the same \gls{ASR}). The strong performance on \gls{PROTEINS} of \texttt{SequentialRandom} is because 1) the perturbation budget $\Delta \leq 0.03n^2$ we set is relatively lenient: with many possible adversarial examples present in the search space, the importance of the different search algorithms may diminish (as we will show later, when we set a much stricter perturbation budget, the out-performance of \gls{GRABNEL} is markedly larger. This could also be seen by the margin of out-performance in Fig. \ref{fig:ablation} when the number of queries is small), and 2) the dataset mainly features modestly-sized graphs on which simple a relatively small number random search might already be sufficient in finding vulnerable edges. On the other hand, \texttt{GrabnelNoSequential} improves significantly over \texttt{Random}, showing the effectiveness of the surrogate in \gls{BO} and in \gls{PROTEINS} it starts off with a much higher \gls{ASR} because it utilises the entire attack budget throughout instead of the approaches using sequential perturbation selection, which is parsimonious with respect the amount of perturbation applied. Nonetheless, in both tasks its final performance is worse than the full \gls{GRABNEL}, presumably due to the fact that the unconstrained search space, which scales exponentially with the attack budget (i.e. the number of edges to edit), is too large for the surrogate model to explore effectively even for modestly-sized graphs.

In view of the strong performance of \texttt{SequentialRandom}, we conduct more detailed experiments to compare it against the full \gls{GRABNEL} and we show the results in Table \ref{tab:comp_against_seqrandom}: with exceptions of \gls{GCN} on \gls{PROTEINS} and Graph U-net on \gls{IMDB-M} where the
two perform similarly, possibly within margin of error, \gls{GRABNEL} outperforms in all other cases: it is easy to see that the margin of \gls{GRABNEL} over SequentialRandom increases significantly as the task difficulty
increases. This is because SequentialRandom takes more queries to reach the similar level of \gls{ASR} of \gls{GRABNEL} and is less
efficient in increasing the attack loss.

\begin{table}[h]
\centering
\caption{Test accuracy of a \gls{GCN} victim model after attacks by \gls{GRABNEL} and \texttt{SequentialRandom} under perturbation budget $\Delta \leq 0.03n^2$. Mean (and $\pm$ standard deviation, if available) shown.}
\begin{tabular}{@{}lccccc@{}}
\toprule
Victim model & \multicolumn{3}{c}{\gls{GCN}} & \multicolumn{2}{c}{Graph U-Net} \\
Dataset & \gls{IMDB-M} & \gls{PROTEINS} & \gls{COLLAB} & \gls{IMDB-M} & \gls{PROTEINS} \\
 \midrule
 \emph{Clean} & $50.53_{\pm 1.4}$ & $71.23_{\pm 2.6}$ & $79.93_{\pm 2.1}$ & $55.33$ & $79.46$\\
\gls{GRABNEL}$^*$ & $\mathbf{45.23}_{\pm 0.2}$ & ${10.82}_{\pm 2.5}$ & $\textbf{35.38}_{\pm 9.3}$ & $\mathbf{41.33}$ & $\mathbf{58.80}$\\
SequentialRandom & $46.22_{\pm 0.2}$ & $\mathbf{10.12}_{\pm 1.5}$ & $49.80_{\pm 6.8}$ & $42.05$ & $66.62$ \\
\bottomrule
\multicolumn{5}{l}{$^*$: Taken from results in the main text.} \\
\label{tab:comp_against_seqrandom}
\end{tabular}
\end{table}

\begin{figure}[t]
    \centering
    \begin{subfigure}{0.49\linewidth}
    \includegraphics[trim=0cm 0cm 0cm  0cm, clip, width=1.0\linewidth]{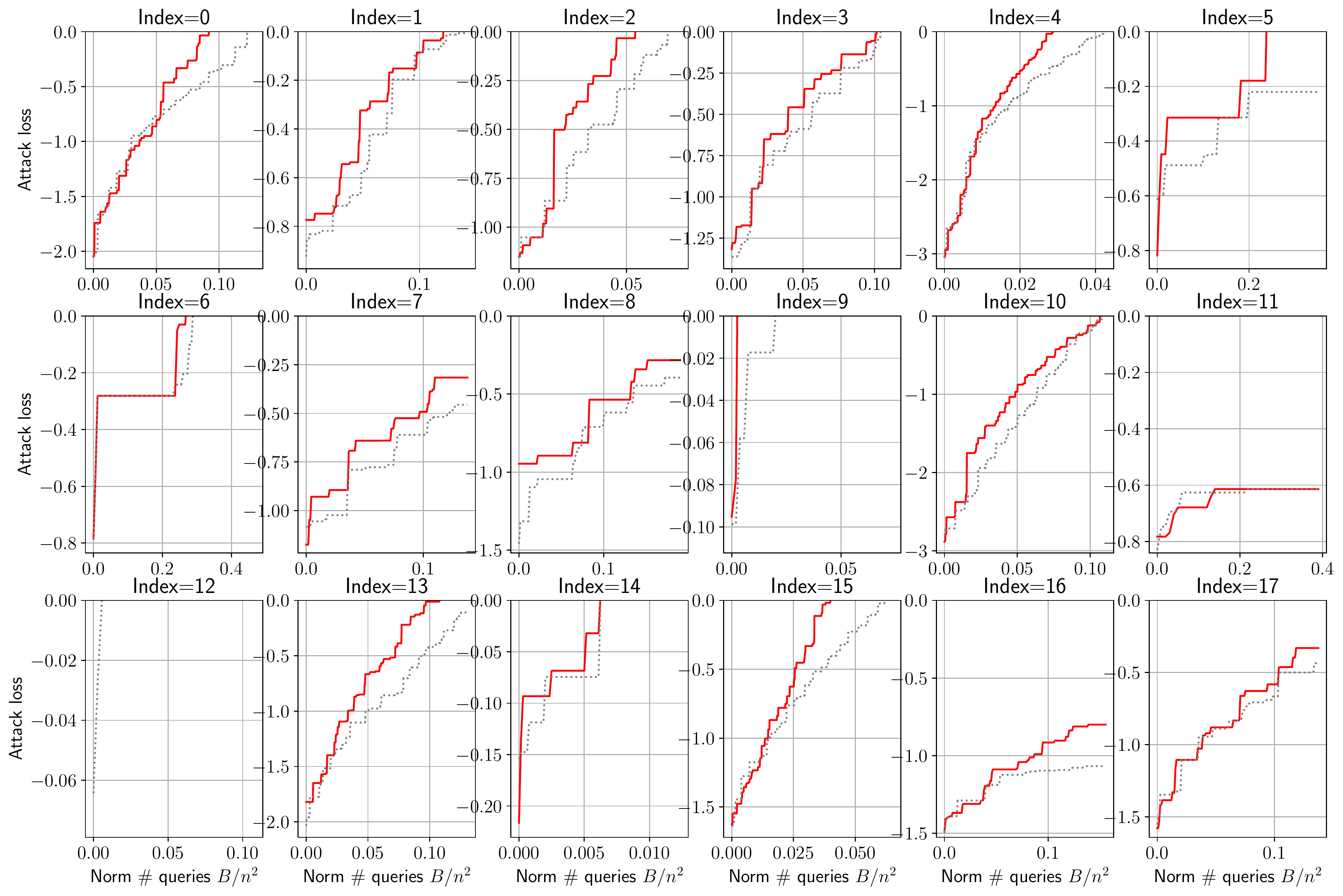}
    \caption{\gls{GCN}}
    \end{subfigure}
         \begin{subfigure}{0.49\linewidth}
    \includegraphics[trim=0cm 0cm 0cm  0cm, clip, width=1.0\linewidth]{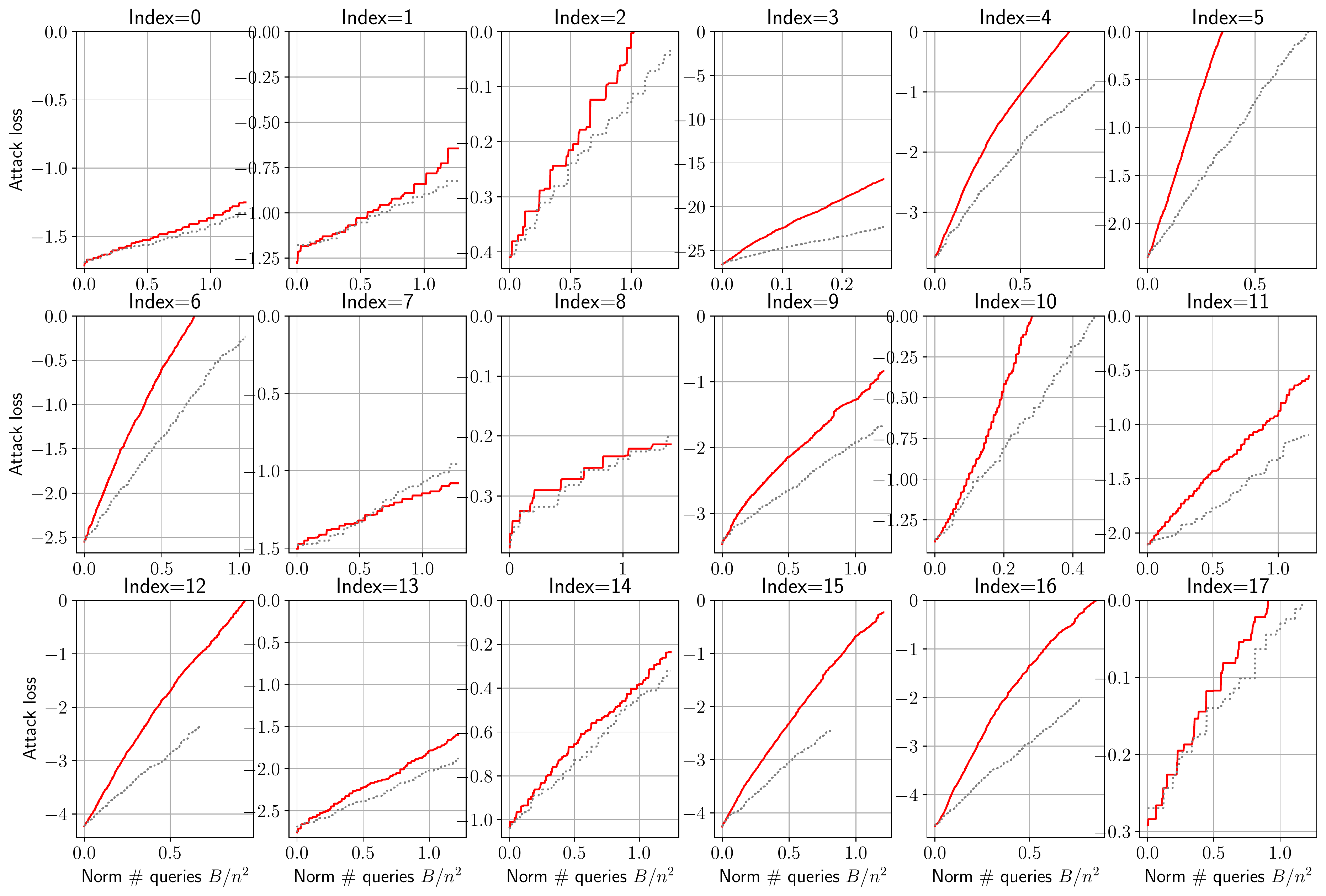}
    \caption{Graph U-Net}
    \end{subfigure}
    \quad
    \caption{Comparison of the attack loss as a function of the (normalised) number of queries to the \gls{GCN}/Graph U-net victim models of \textcolor{red}{\gls{GRABNEL}} and \textcolor{gray}{\texttt{SequentialRandom}} on the \gls{PROTEINS} dataset. 
    }    
    \label{fig:attack_losses}
  \vspace{-5mm}
\end{figure}

However, as discussed, looking at the
performance \textit{only when a relatively lenient budget has been exhausted} can be misleading. We thus test the algorithms while allowing fewer edits, hence making the task
more difficult. Taking the example on the \gls{PROTEINS} dataset where \texttt{SequentialRandom} performs the strongest, we show a set of results with reduced perturbation budgets in Table \ref{tab:comp_against_seqrandom_varying_budget} and the margin of \gls{GRABNEL} over \texttt{SequentialRandom} increases significantly as the task difficulty
increases. This is because \texttt{SequentialRandom} takes more queries to reach the similar level of \gls{ASR} of \gls{GRABNEL} and is less
efficient in increasing the attack loss: as concrete examples, we show the attack loss trajectory as a function of the number of queries to the victim model for both methods in randomly selected attack samples in Fig \ref{fig:attack_losses}: in successfully attacked samples (i.e. the attack loss reaches 0), it is clear to see that \gls{GRABNEL} typically requires fewer queries. Even for the samples that neither managed to successfully attack, \gls{GRABNEL} pushes the losses closer to 0. 

\begin{table}[h]
\centering
\caption{Test acc. of \gls{GCN} after attacks by \gls{GRABNEL} and \texttt{SequentialRandom} on the \gls{PROTEINS} dataset under varying perturbation budgets. Mean ($\pm$ standard deviation, if available) shown. }
\begin{tabular}{@{}lccc@{}}
\toprule
$\Delta$ & $0.03n^2$  & $0.003n^2$ & $0.001n^2$ \\
 \midrule
\gls{GRABNEL} & ${10.82}_{\pm 2.5}$ & $\mathbf{26.43}$ & $\textbf{52.59}$\\
SequentialRandom & $\mathbf{10.12}_{\pm 1.5}$ & ${32.09}$ & $60.51$ \\
\bottomrule
\label{tab:comp_against_seqrandom_varying_budget}
\vspace{-4mm}
\end{tabular}
\end{table}

\begin{wrapfigure}{r}{0.3\textwidth}
\vspace{-1.75cm}
  \begin{center}
    \includegraphics[width=0.3\textwidth]{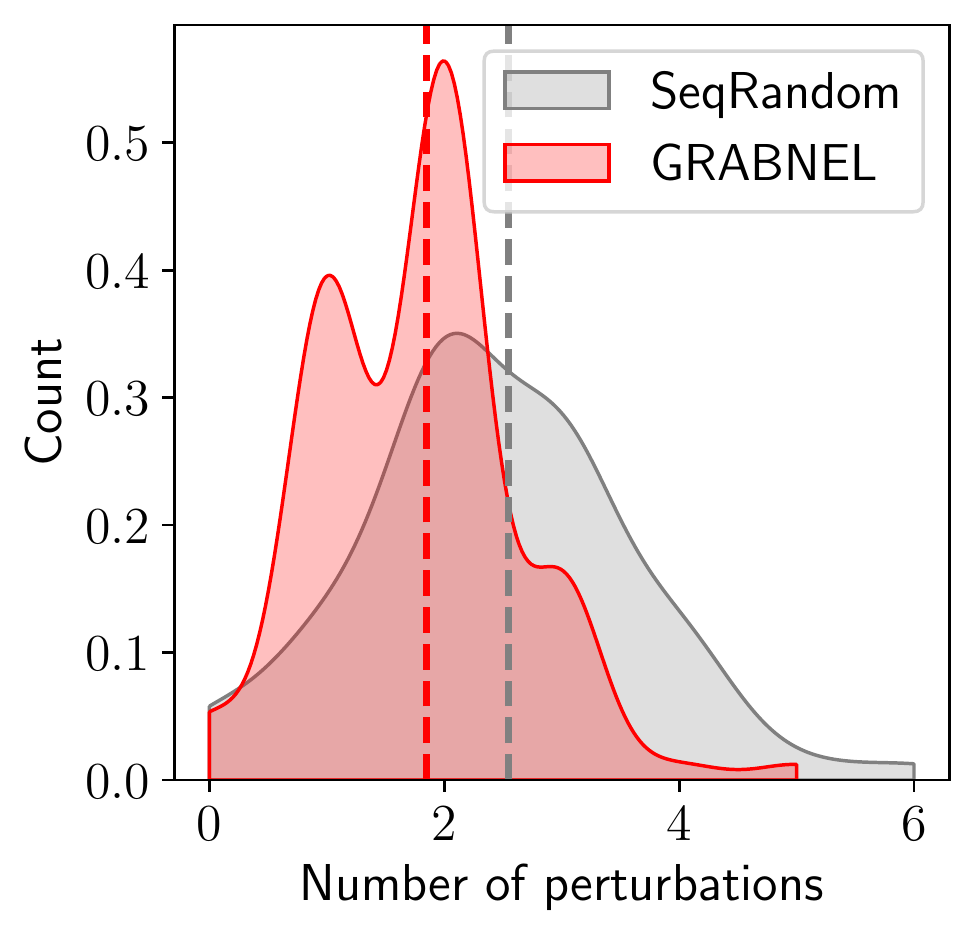}
  \end{center}
  \vspace{-5mm}
  \caption{Distribution of the number of edge rewiring/swapping required to successfully attack \gls{MNIST}-75sp samples in \gls{GRABNEL} and  \texttt{SequentialRandom}. The dotted lines denote the median numbers of edge operations required.}
    \vspace{-1cm}
  \label{fig:n_perturb_mnist}
\end{wrapfigure}

Another concrete example would be the attack on the \gls{MNIST}-75sp task: while both \gls{GRABNEL} and \texttt{SequentialRandom}
converge eventually to 100\% \gls{ASR}, \gls{GRABNEL} converges ~2 times faster: on average, in a
successfully attacked sample, \gls{GRABNEL} rewires $1.84\pm0.7$ edges but \texttt{SequentialRandom} rewires $2.54\pm1.0$ edges (Fig \ref{fig:n_perturb_mnist}),
which suggests that \texttt{SequentialRandom} needs to modify 38\% more edges on average to succeed. Imperceptibility in
graph attack is usually measured in terms the number of edge edits, and thus while in some cases the end
performance of \texttt{SequentialRandom} and \gls{GRABNEL} can be similar especially when a lenient perturbation budget is given,
\gls{GRABNEL} is both more sample-efficient and produces less perceptible attacks. Therefore, we conclude that while \texttt{SequentialRandom} could perform strongly for easier tasks (e.g. high perturbation budget and
smaller graphs) but otherwise \gls{GRABNEL} is significantly better.

\subsection{Runtime Analysis}
\label{app:runtime_analysis}
In this work, we particularly emphasise the desideratum of sample efficiency that we aim to find adversarial examples with the minimum number of queries to the victim model. We believe that this is a practical, and difficult, setup that accounts for the prohibitive monetary, logistic and/or opportunity costs of repeatedly querying a (possibly huge and complicated) real-life victim model. With a high query count, the attacker may also run a higher risk of getting detected. Given this objective, the cost of the algorithm should not only be considered from the viewpoint of computational runtime of the attack algorithm itself alone, and this is a primary reason why we use number of queries as the main cost criterion in our paper (this emphasis on the number of queries over runtime as the main cost metric is common in adversarial attacks in other data structures emphasising sample efficiency \cite{ru2019bayesopt, huang2020corrattack, zhang2021bo} and other related domains, like hyperparameter optimisation. The common assumption is that the cost of the \gls{BO} itself is secondary to the cost of querying the objective function (the cost here should not be interpreted as being the computing cost alone, but includes all the potential costs discussed above). Nevertheless, the runtime analysis is still a relevant metric, and we provide a more comprehensive analysis of the algorithm runtime below.

Each iteration in the main loop of our algorithm can be broadly separated into two parts:

\begin{enumerate}[leftmargin=0.1in, itemsep=0.05pt, topsep=0.05pt]
    \item Initialisation/updates of the surrogates: this step involves the \gls{WL} feature extraction, and initialisation/update of the \gls{BLR} surrogate (the complexity of this step was discussed in Sec \ref{sec:method} with the new data. Note that \gls{BLR} scales much better than \gls{GP}s used in related works \cite{ru2019bayesopt}.

    \item Acquisition function optimisation: this step involves using genetic algorithms to optimise the acquisition function. It is further broken into 2 sub-parts:
    
    \begin{enumerate}[leftmargin=0.1in, itemsep=0.05pt, topsep=0.05pt]
        \item \gls{GA} steps, which involve the selection of the population and mutation and crossover operations on the parents. For the manipulations here, we do not have to store the full graphs, but we instead only have to maintain a tuple of the edges that are flipped/rewired.
        \item Conversion of the tuple into full graph objects (we use \gls{DGL} \cite{wang2019deep} for implementation), and call the trained \gls{BLR} to obtain predicted mean/variance and compute the acquisition function value.
    \end{enumerate}

\end{enumerate}

\begin{table}[h]
\centering
\caption{Runtime analysis of \gls{GRABNEL} in terms of average second per iteration (standard deviation in brackets; slowest step in bold). $H$ denotes the number of \gls{WL} iterations performed. Benchmarked on an otherwise idle machine with AMD Ryzen 7 CPU and 32 GB RAM. We used \gls{GCN} victim model. Results may vary significantly depending on the hardware, system load and the hyperparameters used}
\resizebox{0.7\linewidth}{!}{
\begin{tabular}{@{}llllll@{}}
\toprule
\# nodes & \# edges & $H$ & Step 1 & Step 2a & Step 2b \\
 \midrule
17	& 106	& 1 & 0.022 (0.0027) & 0.0344 (0.0002)	& \textbf{0.482 (0.006)} \\
72	& 719	& 1 & 0.251 (0.003)	& 0.0371 (0.0006) &	\textbf{0.458 (0.009)} \\
1961 & 	5336 & 0 &	\textbf{1.76 (0.12)} & 	0.0555 (0.0007)	& 1.52 (0.016) \\
\bottomrule
\label{tab:runtime_analysis}
\end{tabular}
}
\end{table}

Note that even for a graph with almost 2k nodes and >5k edges (which is larger than most graphs in the \gls{TU} dataset for graph classification), the runtime is still manageable on a mainstream desktop-grade PC. In fact, the \gls{GA} itself is efficient, and the much slower step is Step 2b: instead of our algorithm itself being inefficient, this is because of the large overhead in graph representation conversion between the list of changed edges (e.g. $([1, 2], [3, 4])$, which denotes a perturbed graph with edges $e_{(1, 2)}$  and  $e_{(3, 4)}$ flipped from the original graph) and an actual \gls{DGL} graph object. For better efficiency, it should be possible to reduce the number of such conversions as the perturbed and original graphs differ only at the flipped edges which only make up a very small fraction of all edges. At the very least, since each conversion is independent for other graphs, we can parallelise it for candidates in the population of the genetic algorithm (the current code does this sequentially).

To give a better context, on a shared Intel Xeon Gold server where we conduct the majority of experiments (we unfortunately could not provide very reliable and accurate statistics due to the varying load by other users), each attack on a single graph on the \gls{IMDB-M} (average ~66 edges per graph) dataset usually takes < 1min. On \gls{COLLAB}, which on average is much larger, each attack on average takes ~1h (Note we set the maximum number of queries to be dependent on the sizes of the graph, so larger graphs are also proportionally allocated a higher query budget and hence each run could be much longer). Furthermore, the sizes of graphs within a dataset can vary a lot, and runtime also depends a lot on the difficulty of attack (if an adversarial perturbation is easily found the run is terminated early). A further comparison with \gls{RL-S2V} \cite{dai2018adversarial} is that to run the same task on ER-graphs attack with 15-20 nodes, our method takes 30min - 1h. \gls{RL-S2V}, which requires a separate validation set to train policy on, requires approx. 1.5h with \textsc{gpu} acceleration and approx. 12h without (we do not currently use any \textsc{gpu} acceleration for our method).

\end{document}

%% file: math.tex
\usepackage{amsmath,amsfonts,bm}

\def\eqref#1{equation~\ref{#1}}

\def\1{\bm{1}}

\def\eps{{\epsilon}}

\def\vphi{{\bm{\phi}}}

\def\mPhi{{\bm{\Phi}}}

\DeclareMathAlphabet{\mathsfit}{\encodingdefault}{\sfdefault}{m}{sl}
\SetMathAlphabet{\mathsfit}{bold}{\encodingdefault}{\sfdefault}{bx}{n}

\newcommand{\E}{\mathbb{E}}